\documentclass{article}
\usepackage[margin=2.8cm]{geometry}
\usepackage[]{inputenc} %
\usepackage[T1]{fontenc} %

\usepackage{graphicx}
\usepackage{picinpar}
\usepackage{subcaption}
\usepackage{colortbl}
\usepackage[table,xcdraw]{xcolor}

\usepackage{microtype} %

\usepackage{booktabs}
\usepackage{multirow}
\usepackage{soul}
\usepackage{enumerate}
\usepackage{color}
\usepackage{hyperref}
\usepackage{amsmath}
\usepackage{longtable}

\usepackage{algorithm}
\usepackage{algpseudocode}
\usepackage{kpfonts}

\usepackage{enumitem}
\usepackage{listings}
\usepackage[sort]{natbib}
\setcitestyle{round, elide, numbers}

\usepackage{cleveref}
\usepackage{twemojis}

\usepackage{siunitx}
\usepackage{appendix}
\usepackage{authblk}

\usepackage{fancyhdr}
\fancypagestyle{firstpage}{\fancyhead[L]{EMMA-500 Gen 2}\fancyhead[R]{}}

\newcommand{\furl}[1]{\footnote{\url{http://#1}}}
\usepackage{xspace}

 \def\website{\url{https://mala-lm.github.io/emma-500-v2}}

 \def\msummlong{\url{https://huggingface.co/datasets/MaLA-LM/MassiveSumm_long}}
 \def\msummshort{\url{https://huggingface.co/datasets/MaLA-LM/MassiveSumm_short}}
\def\LUMI{LUMI Supercomputer\xspace}
\def\MALA{\texttt{MaLA}\xspace}
\def\EMMA{\texttt{EMMA-500}\xspace}

\title{Massively Multilingual Adaptation of Large Language Models Using Bilingual Translation Data}

\author[$^{1}$]{Shaoxiong Ji\thanks{Corresponding author}}
\author[$^1$]{Zihao Li}
\author[$^1$]{Jaakko Paavola}
\author[$^1$]{Hengyu Luo}
\author[$^1$]{J\"org Tiedemann}

\affil[ ]{$^1$University of Helsinki}

\affil[ ]{\textit {\{shaoxiong.ji;~zihao.li;~jaakko.paavola;~hengyu.luo;~jorg.tiedemann\}@helsinki.fi
}
}

\date{}

\begin{document}
\maketitle

\thispagestyle{firstpage}

\begin{abstract}
This paper investigates a critical design decision in the practice of massively multilingual continual pre-training --- the inclusion of parallel data. Specifically, we study the impact of bilingual translation data for massively multilingual language adaptation of the Llama3 family of models to 500 languages. To this end, we construct the \MALA bilingual translation corpus, containing data from more than 2,500 language pairs. Subsequently, we develop the \EMMA Llama 3 suite of four massively multilingual models --- continually pre-trained from the Llama 3 family of base models extensively on diverse data mixes up to 671B tokens --- and explore the effect of continual pre-training with or without bilingual translation data. Comprehensive evaluation across 7 tasks and 12 benchmarks demonstrates that bilingual data tends to enhance language transfer and performance, particularly for low-resource languages. We open-source the \MALA corpus, \EMMA Llama 3 suite artefacts, code, and model generations.
\vspace{2ex}

\noindent \twemoji{globe with meridians} Website: \href{https://mala-lm.github.io/emma-500-gen2}{mala-lm.github.io/emma-500-gen2} \\
\noindent \twemoji{robot} Models: \href{https://huggingface.co/collections/MaLA-LM/emma-500-66eaa9acf1f512c8915b7166}{huggingface.co/collections/MaLA-LM $\rightarrow$~EMMA-500}\\
\noindent \twemoji{floppy disk} Data: \href{https://huggingface.co/datasets/MaLA-LM/mala-bilingual-translation-corpus}{huggingface.co/datasets/MaLA-LM/mala-bilingual-translation-corpus}\\
\noindent \twemoji{bar chart} Evaluation: \href{https://github.com/MaLA-LM/emma-500}{github.com/MaLA-LM/emma-500}

\end{abstract}

\tableofcontents

\section{Introduction}
\label{sec:introduction}

Large language models (LLMs) pre-trained on massive data have promoted multilingual natural language processing (NLP). 
However, multilingual models such as BLOOM \citep{scao2022bloom} and Llama \citep{touvron2023llama,touvron2023llama2} often struggle with low-resource languages and are still limited in their language coverage \citep{huang-etal-2023-languages,sindhujan2025llmsstrugglereferencelesstranslation,huang2025surveylargelanguagemodels}.
Recent works extend pre-trained LLMs into multiple languages via continual pre-training (CPT). 
For example, LLaMAX \citep{lu2024llamax} and xLLMs-100 \citep{lai-etal-2024-llms} adopt CPT and instruction fine-tuning to extend existing LLMs into 100 languages, and MaLA-500 \citep{lin2024mala} and EMMA-500 \citep{ji2024emma} perform continual pre-training (low-rank and full-parameter CPT using Llama 2) to adapt LLMs into 500 languages. 
Despite these efforts, challenges still remain in adapting LLMs to low-resource languages, especially in \textbf{a massively multilingual scenario with more than 500 languages}.

This paper studies CPT in a massively multilingual setting. 
Prior work like LlaMAX \citep{lu2024llamax} uses both monolingual and parallel texts for CPT in 100 languages, and EMMA-500 Llama 2 \citep{ji2024emma} uses only monolingual texts in 500 languages. 
Our novelty is to scale the number of languages up to more than 500 languages, compile a new corpus with bilingual translation data (\MALA), and study the comparative effects of continual pre-training with monolingual and bilingual translation data.\footnote{Monolingual data consists of texts written in a single language. Bilingual translation data, also called {\em parallel corpora}, comprises pairs of sentences in two different languages that express the same meaning. In this paper, we treat the terms bilingual translation corpora/texts/data, bilingual corpora/texts/data, parallel corpora/texts/data, and bitexts as equivalent.}  

\paragraph{Contributions} 
Our contributions are three-fold:

\begin{itemize}[nolistsep,noitemsep]
    \item \textbf{\textsc{Data:}} We compile a bilingual translation corpus for Massive Language Adaptation in more than 2,500 language pairs and 500 languages, namely the \textbf{\MALA} translation corpus.
    \item \textbf{\textsc{Models:}} We train and release 4 models, namely \EMMA  Llama 3/3.1 Mono/Bi for Enhancing Massively Multilingual Adaptation,\footnote{ %
    We use \MALA and \EMMA to name the corpus and models, following the naming convention of EMMA-500 Llama 2 \citep{ji2024emma} which is not an artifact of this paper. 
    \MALA and \EMMA are a collection of corpora and models. In this paper, we compile and release the MaLA bilingual translation corpus, i.e., \twemoji{smiling face with open hands} \href{https://huggingface.co/datasets/MaLA-LM/mala-bilingual-translation-corpus}{MaLA-LM/mala-bilingual-translation-corpus}.
    ``Mono'' and ``Bi'' in model names indicate CPT on monolingual (\cref{fig:mix_monolingual}) and bilingual (\cref{fig:mix_bilingual}) mixes, respectively.} by continually pre-training of Llama 3 \& 3.1 (8B) using both monolingual and bilingual MaLA corpus augmented with diverse data types, up to 671B tokens.
    \item \textbf{\textsc{Evaluation:}} We conduct a comprehensive evaluation across 7 tasks and 12 benchmarks. Our empirical evaluation ablates the impact of two diverse data mixes and analyzes gains in task generalization and multilingual robustness.
\end{itemize}

\paragraph{Evaluation Results and Findings}

\begin{itemize}[nolistsep,noitemsep]
	\item CPT using a data mix with bilingual translation data generally exhibits better multilingual performance than a monolingual mix\footnote{A monolingual mix (\cref{fig:mix_monolingual}) contains monolingual data in different languages but not in the aligned format as parallel data.}, particularly in low-resource languages and in machine translation tasks that directly benefit from parallel data.
	\item Heavily pre-trained models (e.g., Llama 3 and 3.1) that consume more training tokens are more resistant to further adaptation than English-centric models (e.g., Llama 2) when scaling to include many additional languages. 
	\item As for overall performance, our \EMMA models are the best at machine translation (Flores200) and competitive at text classification (Taxi1500 and SIB-200) and commonsense reasoning (XCOPA and XStoryCloze).
        \item \EMMA CPT models exhibit a lower average accuracy on the BELEBLE comprehension benchmark, but they outperform baselines across a greater number of languages.
\end{itemize}
While multilingual models can achieve broad coverage, perfect uniformity across all tasks and languages remains an unattainable goal.
However, we show that multilingual performance and language equality can be pushed forward with parallel training data.

\paragraph{Outline}
\Cref{sec:data_training} presents the data and model training with a newly compiled bilingual translation corpus introduced in \Cref{sec:mala_bilingual}, data mixing introduced in \Cref{sec:data_mixing}, and settings for model training introduced in \Cref{sec:model_training}. 
\Cref{sec:mala_bilingual_detailed,sec:data_mix_detailed} describe the details about the \MALA translation corpus and data mixes. 
We evaluate the resulting models and discuss the evaluation results in \Cref{sec:evaluation}. Detailed evaluation setup and per-benchmark and per-language results are presented in \Cref{sec:detailed_setup,sec:detailed_results} respectively.
We conclude the paper in \Cref{sec:conclusion}.
\Cref{sec:related,sec:ethics} introduce related work and ethics consideration.

\section{Data and Model Training}
\label{sec:data_training}

\subsection{\MALA Translation Corpus}
\label{sec:mala_bilingual}

Bilingual translation corpora are language datasets that contain text data in one language aligned with text data in another language. 
We extend the MaLA (Massive Language Adaptation) corpus \citep{ji2024emma} by incorporating parallel data in more than 500 languages and 2,500 language pairs. 
The resulting parallel dataset is named the \MALA translation corpus (\textbf{\MALA} for short), which is suitable for adapting language models in massively multilingual scenarios. 
The section describes the process of building the \MALA translation corpus in a way similar to the MaLA corpus with monolingual texts, but focuses on bilingual texts. 

We follow a similar data integration pipeline to the process of compiling the MaLA corpus \citep{ji2024emma}, including extraction, harmonization, language code normalization, and writing system recognition.  
The bilingual data comes from various sources, including OPUS \citep{opus_2012}, NLLB \citep{nllb2022}, and Tatoeba \citep{tatoeba}.
The datasets from OPUS are made by an existing compilation: Lego-MT \citep{yuan-etal-2023-lego}. 
\Cref{tab:bilingual_sources} in \Cref{sec:data_sources} shows the data sources for bilingual texts. 
The main difference in the script recognition, as well as language code conversion, with the bilingual corpora, is in the form of the label; we obtain a label in the form of a language pair, e.g., eng\_Latn-zho\_Hani. 

\paragraph{Language Code Normalization}
Language code normalization converts various language codes into a standardized format to ensure consistency and compatibility across different systems and applications.
With the bilingual corpora, we face similar issues as with the monolingual ones when converting language denotations given in OPUS\footnote{\url{https://github.com/Helsinki-NLP/OPUS}} to ISO 639-3 language codes. 
Moreover, with bilingual corpora, we want to specify dialects based on the ISO 3166-1 alpha-3 standard. The recognition and handling of language codes are based on the following procedure:
\begin{itemize}[nolistsep,noitemsep]
     \item If the language code of a dataset provided by OPUS matches a language code in ISO-639-3, then we consider it such.
     \item If the language code does not match one in ISO-639-3, we use the \texttt{langcodes} package\footnote{\url{https://pypi.org/project/langcodes/}} to convert it to ISO-639-3.
     \item If the above steps fail, we assign ``unknown'' as the language code.
\end{itemize}

\paragraph{Writing System Recognition}
We implement writing system identification following ISO 15924 standards using the GlotScript library \citep{kargaran2023glotlid}. For script detection, we analyze 100-line samples by default, reverting to first-line analysis when standard detection fails---a known limitation affecting both our bilingual and monolingual corpora \citep{ji2024emma}.
We do not classify a dataset into multiple scripts, even in cases of code-mixing, where multiple scripts are used.

\paragraph{Data Cleaning}
The bilingual corpus compilation faces significant quality variability, such as noisy sentence pairs. 
The Lego-MT dataset \citep{yuan-etal-2023-lego} applies some data cleaning, including deduplication, removing missing translations, and length mismatching. 
For other data sources, we add various procedures, including dataset-specific cleaning and deduplication, after integrating all data sets into our collection.
We eliminate lines that contain the exact word or character repeated more than five consecutive times.
This problem appears in particular in the Tatoeba parallel training data~\citep{tatoeba} and in the majority of these cases is erroneous.
We use OpusFilter~\citep{aulamo-etal-2020-opusfilter} for deduplication of data points.

\paragraph{Key Statistics}
After pre-processing and cleaning, we obtain the \MALA translation corpus in 2,507 language pairs.
\Cref{tab:key_stat} shows the total number of whitespace-separated tokens and the number of language pairs across different resource categories.
Compared with Lego-MT and NLLB, \MALA has a similar number of language pairs but more tokens. 
We categorize language pairs into 5 resource levels based on token counts: high-resource (>1B), medium-high (>500M), medium (>100M), medium-low (>10M), and low (>1M). 
Different from the monolingual MaLA, we add two categories of ``very high'' and ``very low'' resources in the resource level classification, i.e., very high-resource pairs (>10B) and very low-resource (<1M). %
In total, there are more than 426B tokens in the \MALA translation corpus.\footnote{Note that whitespace-based token counting is not accurate for a language where words are not typically separated by spaces. In these languages, the absence of whitespace makes it challenging to determine token boundaries, leading to inaccurate token counts when using whitespace as a delimiter. We use whitespace as the delimiter because of its efficiency in processing text.}
We further sample the \MALA translation corpus for continued training LLMs, considering a balanced corpus size and language coverage.

\begin{table}[ht]
\caption{Key statistics of the \MALA translation corpus and comparison with existing parallel corpora.}
\label{tab:key_stat}
\scriptsize
\centering
\setlength{\tabcolsep}{3pt}
\begin{tabular}{lllllll}
\toprule
\multirow{2}{*}{\textbf{Categories}} & \multicolumn{2}{c}{\textbf{Lego-MT}} & \multicolumn{2}{c}{\textbf{NLLB}} & \multicolumn{2}{c}{\textbf{\MALA}} \\ \cline{2-7} 
                                     & \textbf{Pairs}   & \textbf{Tokens}   & \textbf{Pairs}  & \textbf{Tokens} & \textbf{Pairs}   & \textbf{Tokens}  \\ \hline
very high                            & 4                & 5.1E+10           & 4               & 2.8E+10         & 4                & 8.5E+10          \\
high                                 & 51               & 1.4E+11           & 51              & 1.2E+11         & 83               & 2.1E+11          \\
medium-high                          & 22               & 1.5E+10           & 65              & 4.5E+10         & 67               & 4.7E+10          \\
medium                               & 75               & 1.8E+10           & 264             & 6.1E+10         & 281              & 6.4E+10          \\
medium-low                           & 113              & 4.1E+09           & 480             & 2.0E+10         & 508              & 2.0E+10          \\
low                                  & 350              & 1.4E+09           & 491             & 1.7E+09         & 655              & 2.5E+09          \\
very low                             & 1893             & 1.7E+08           & 1154            & 1.3E+08         & 909              & 1.8E+08          \\
\hline
sum                                  & 2508             & 2.2E+11           & 2509            & 2.8E+11         & 2507             & 4.3E+11     \\    
\bottomrule
\end{tabular}
\end{table}

\subsection{Data Mixing}
\label{sec:data_mixing}

We blend the data compiled in \MALA with multilingual non-parallel data obtained from the cleaned and deduplicated MaLA corpus \citep{ji2024emma} along with texts selectively sourced from factual and high-quality domains with a view to retaining the knowledge acquired during the \textit{annealing} phase of pre-training \citep{DBLP:journals/corr/abs-2404-06395}. We mirror existing practice and sample books and scientific papers \citep{DBLP:conf/acl/SoldainiKBSAABC24} along with instruction-like \citep{DBLP:conf/acl/MainiSBG0J24} data. Following \citet{ji2024emma}, we use scientific papers and books from CSL~\citep{li2022csl}, pes2o~\citep{peS2o}, and free e-books from the Gutenberg project\footnote{\url{https://www.gutenberg.org/}} compiled by \citet{project_gutenberg_HF_manu}. Multilingual instruction data is sourced from the training set of xp3x (Crosslingual Public Pool of Prompts eXtended)\footnote{\url{https://hf.co/datasets/CohereForAI/xP3x}} and the Aya collection\footnote{\url{https://hf.co/datasets/CohereForAI/aya_collection_language_split}}. Finally, we augment our mixes with code and code-adjacent procedural text due to its demonstrable benefits towards reasoning and entity-tracking \citep{DBLP:conf/iclr/RuisMBKGLKRGB25, DBLP:journals/corr/abs-2409-04556} (details in \Cref{sec:additional_code}).

\begin{figure}[ht!]
    \centering
    \begin{subfigure}[b]{0.3\textwidth}
        \centering
        \includegraphics[width=\textwidth]{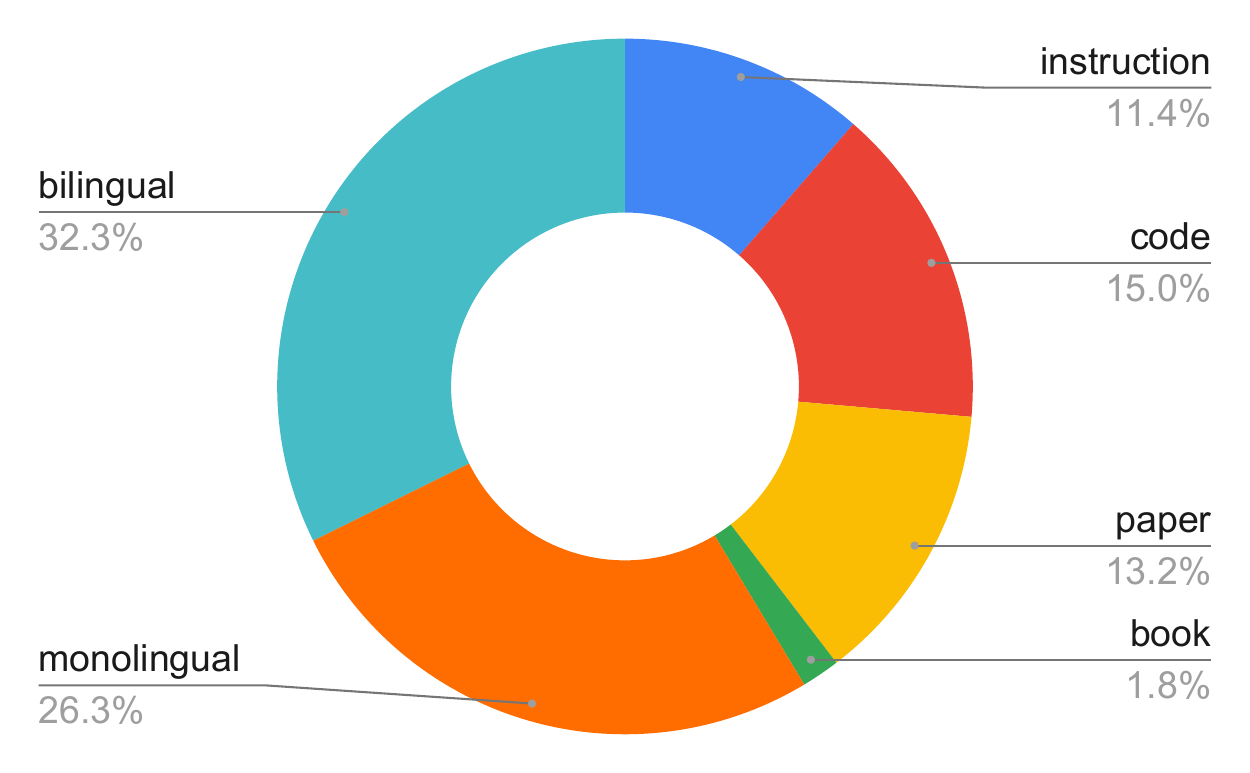}
        \caption{Data mix 1: bilingual}
        \label{fig:mix_bilingual}
    \end{subfigure}
    \begin{subfigure}[b]{0.3\textwidth}
        \centering
        \includegraphics[width=\textwidth]{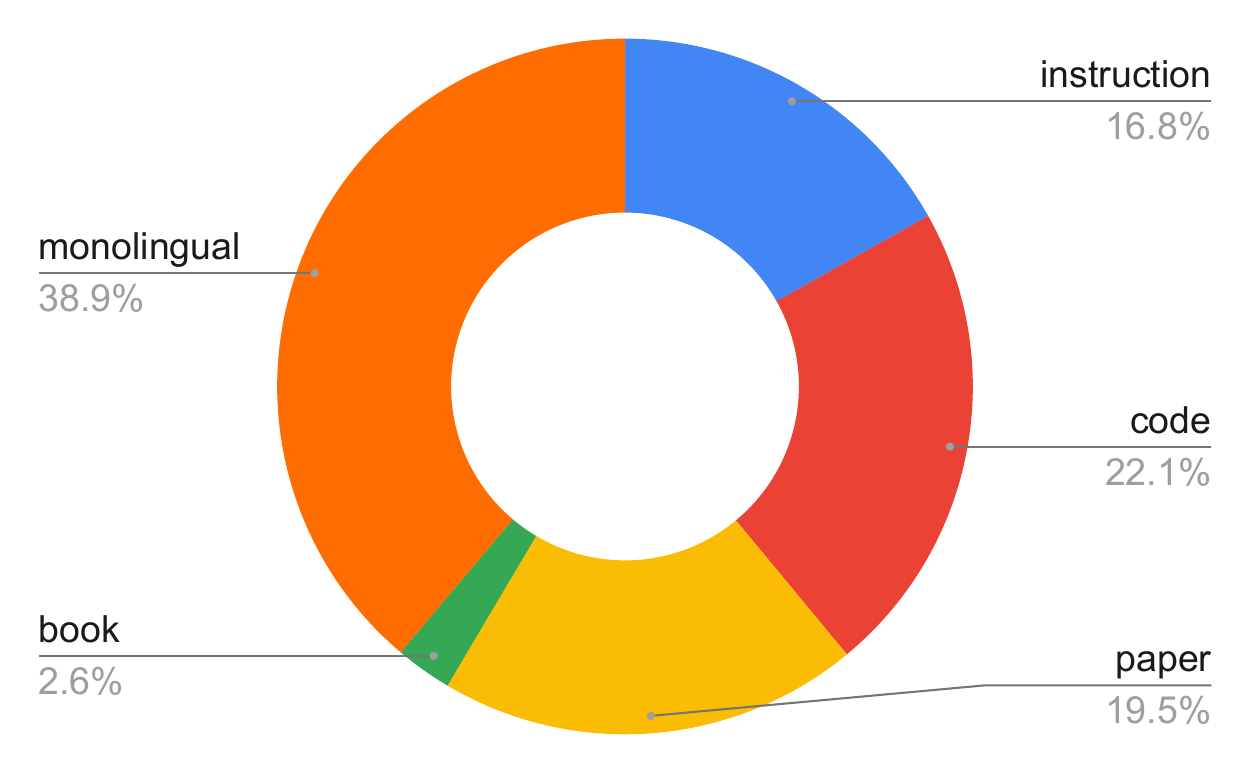}
        \caption{Data mix 2: monolingual}
        \label{fig:mix_monolingual}
    \end{subfigure}  
    \caption{Two date mixes and their composition. The bilingual mix includes all types of data. The monolingual mix consists of a subset of the bilingual mix that excludes bilingual data.}
    \label{fig:composition_mixes}
\end{figure}

We manually mix up different types of data to balance the language coverage across different levels of resources and types while ensuring that low- and medium-resource languages remain overrepresented. \Cref{fig:composition_mixes} shows the composition categorized by language resources according to the number of tokens. Notably, medium-resource language pairs contribute the majority of bilingual data, and medium-high-resource languages are the largest category for monolingual and instruction data.

\begin{figure}[ht!]
    \centering
    \begin{subfigure}[b]{0.3\textwidth}
        \centering
        \includegraphics[width=\textwidth]{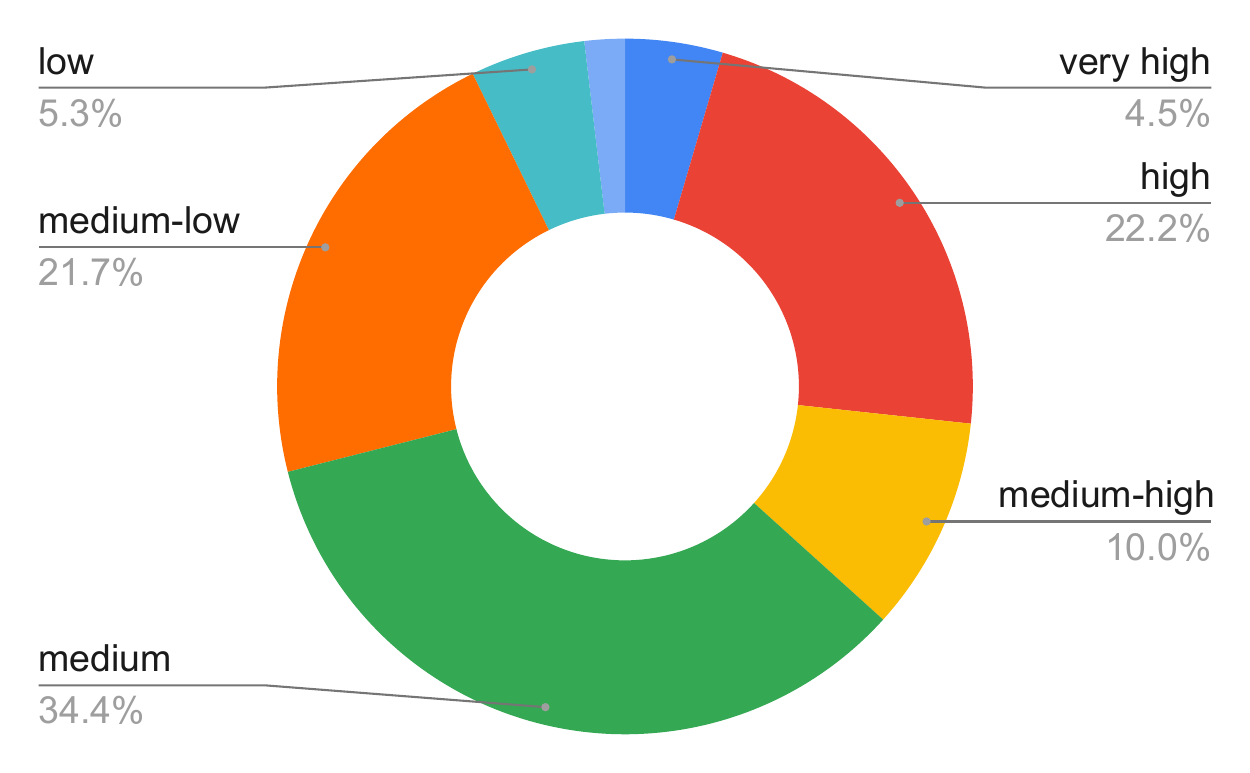}
        \caption{Bilingual translation data}
        \label{fig:type_bilingual}
    \end{subfigure}
    \begin{subfigure}[b]{0.3\textwidth}
        \centering
        \includegraphics[width=\textwidth]{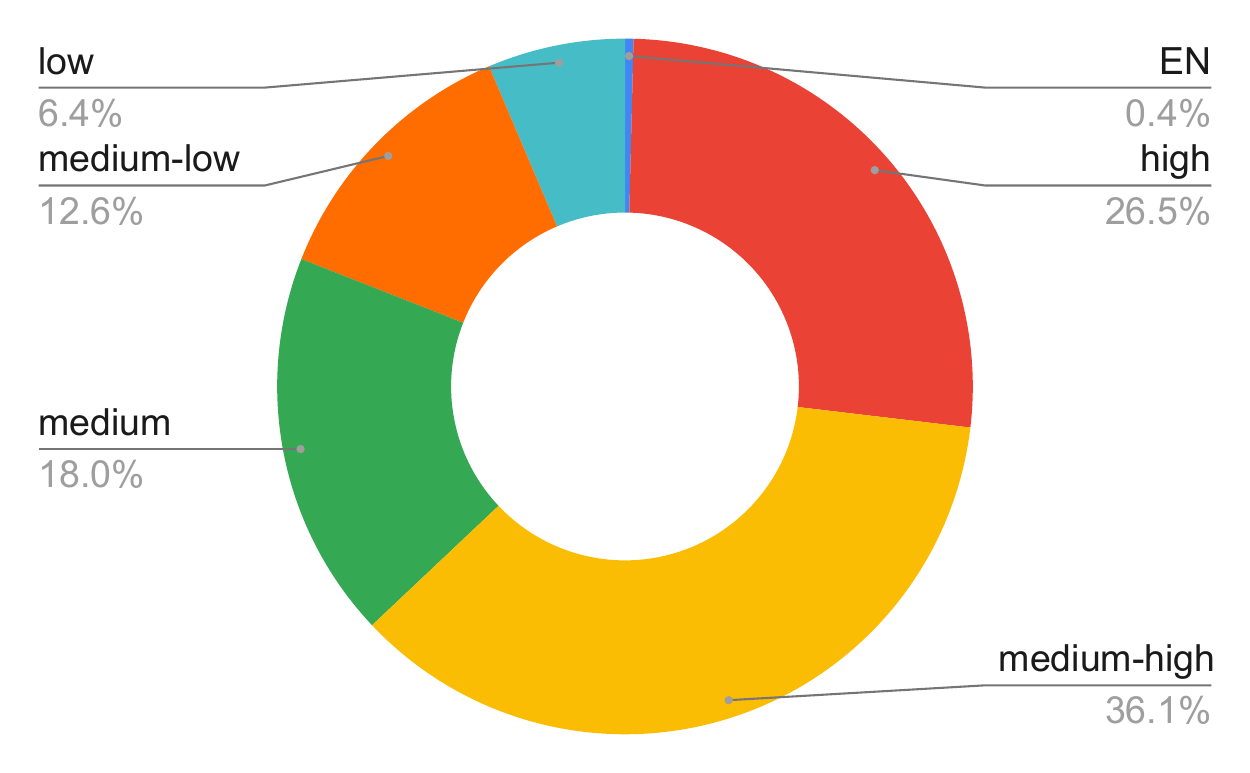}
        \caption{Monolingual data}
        \label{fig:type_monolingual}
    \end{subfigure}
    \begin{subfigure}[b]{0.3\textwidth}
        \centering
        \includegraphics[width=\textwidth]{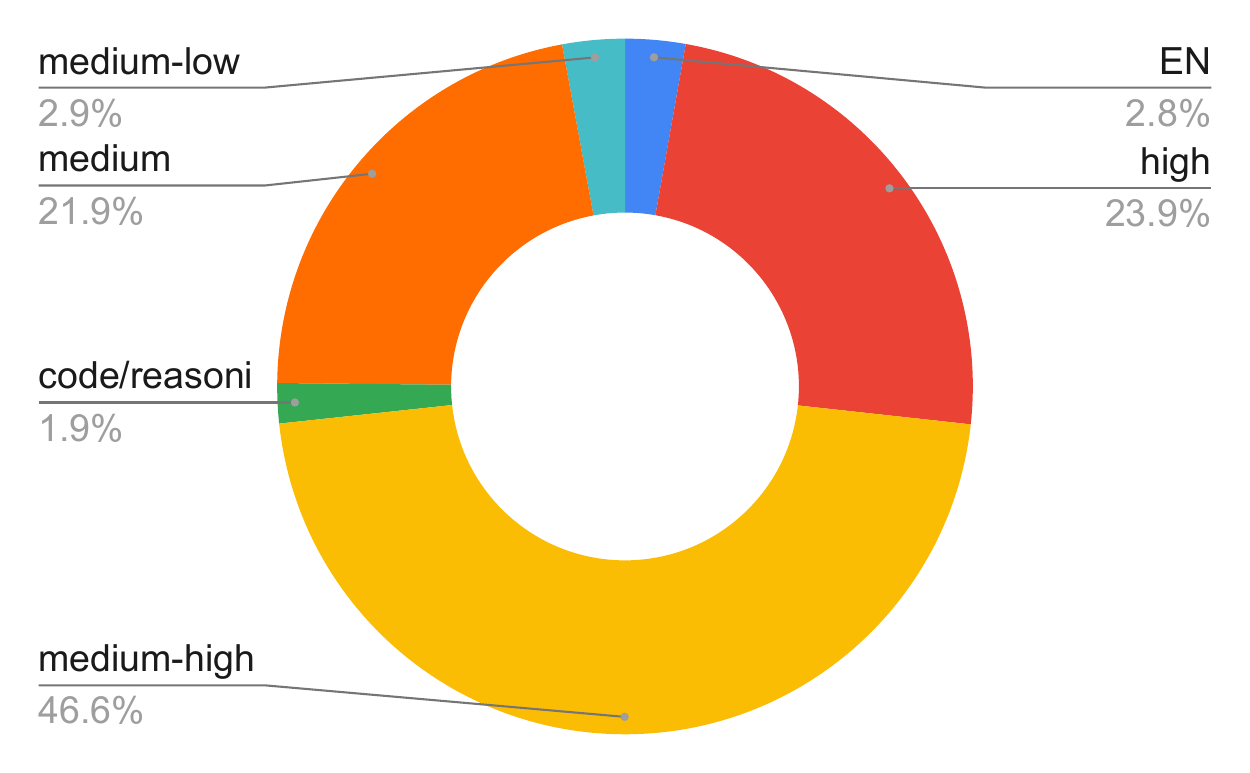}
        \caption{Instruction data}
        \label{fig:type_instruction}
    \end{subfigure}    
    \caption{Composition by language resource levels of different data types.}
    \label{fig:composition_types}
\end{figure}

\paragraph{Two Data Mixes}

We make two data mixes to ablate the effect of incorporating bilingual texts into continual pre-training.
The first data mix is a \textbf{bilingual mix} (\Cref{fig:mix_monolingual}), which incorporates various data types. 
The second is a \textbf{monolingual mix} (\Cref{fig:mix_monolingual}), which is derived from the bilingual mix but specifically omits any bilingual data, focusing solely on a subset with monolingual texts per document.
Detailed data statistics of these two mixes are presented in \Cref{tab:data_mix}.

Most translation data are aligned sentences, and there is not sufficient document-level aligned data publicly available to be collected in the \MALA corpus, especially in the massively multilingual scenario.
When using translation data, we concatenate the source and target language texts in a specific format to form a chunk of pairs in the same language pair.   
For every ten samples, i.e., sentence pairs, to make a document for training,\footnote{The choice of 10 is inspired by ``y\'i m\`u sh\'i h\'ang'', a Chinese idiom that literally translates to ``one glance ten lines'', means that someone reads very quickly and efficiently.} the format is structured as follows:
\begin{lstlisting}
[{src_lang_code}]: {src_text} [{tgt_lang_code}]: {tgt_text} \n
# 8 lines for 8 samples 
[{src_lang_code}]: {src_text} [{tgt_lang_code}]: {tgt_text}
\end{lstlisting}
In this format, the notation $\{\cdot\}$ denotes the variables for source and target language codes and texts.
This method allows us to work with pseudo-document-level data and clearly delineate between the source and target languages, facilitating better processing and understanding of the translation data without switching between multiple languages within the context window. 
By organizing the data this way, we facilitate the model to learn from the relationships between the two languages effectively.
This structured approach ensures clarity and consistency in how bilingual data is presented, making it easier to process and analyze.

\Cref{fig:composition_mixes} shows the composition of our two data mixes. In the monolingual mix, monolingual web-crawled text is the largest data type, as its name suggests. 
The bilingual mix incorporates a considerably bigger portion of bilingual texts, 6\% more than the monolingual texts.
Continual training on these two mixes facilitates the adaptation of LLMs to massively multilingual languages and analyzes the effect of scaling massively multilingual training using bilingual data.

\paragraph{Manual Curation of Data Mixes}
Our data mixes in \Cref{fig:composition_mixes} are crafted through experience and intuition.
We draw on our knowledge and understanding of the domain to create effective data combinations to balance the language resources and represent different text types. 
This allows for a reasonable selection that can capture the complexities of the data landscape and is aligned with the goals for massively multilingual adaptation. 
From an algorithmic perspective, grid search or other similar methods could be used to explore the space of possible data distributions.
However, algorithmic search involves systematically exploring a predefined set of hyperparameter values by evaluating all possible combinations through model training on the searched mixes, which can become computationally expensive and time-consuming, especially as the number of parameters increases. 
This makes it challenging to apply algorithms like grid search effectively in scenarios where the data mix needs to be optimized, which would require thousands of training runs and tens of millions of GPU hours or even more.
Searching for an optimal (or near-optimal) data mix is practically infeasible.
Our paper focuses on offering a useful resource for continual pre-training in a massively multilingual scenario. 
To do this, we select the corpus to maintain diversity and balance among various languages and text types that align with the model's intended goal. 

\subsection{Model Training}
\label{sec:model_training}

We continue training the decoder-only Llama 3 models (8B parameters) using the causal language modeling objective, exposing the pre-trained model to new data and languages to develop our \EMMA model.
To enhance efficiency, we use training strategies that optimize memory usage, precision handling, and distributed training. 
\EMMA Llama 3 series models are trained on the \LUMI, powered by 100\% renewable and carbon-neutral energy, utilizing 64 compute nodes with 256 AMD MI250x GPUs (512 AMD Graphics Compute Dies) with the GPT-NeoX framework~\citep{gpt-neox-library}.

We continue training the base models in a full-parameter manner without modifying the tokenizer.
The training setup includes a global batch size of 2048 and sequence lengths of 8192 tokens. 
Model- and data-specific settings are presented in \Cref{tab:models_settings}.
For training on the monolingual mix, the process spans 25,000 steps, accumulating a total of 419 billion Llama 3 tokens. 
For the bilingual mix that contains more tokens, we train for more steps, up to 40,000, leading to a total of 671 billion tokens. 
We employ the Adam optimizer \citep{kingma2015adam} with a learning rate of 0.0001, betas set to [0.9, 0.95], and an epsilon of 1e-8.
We experiment with different learning rates and evaluate early checkpoints trained up to 5,000 steps (84 billion tokens). 
We find that training with the original learning rate of 0.0003 used by the Llama 3 leads to instability, such as many fast spikes, resulting in poor performance. Thus, we opt for a smaller learning rate with a more stable training curve. 
All experiments consume more than 800k GPU hours on the \LUMI.
However, we could not perform a grid search on the learning rate due to the constraint of computing resources. 

A cosine learning rate scheduler, with a warm-up of 1,000 and 2,000 iterations for monolingual and bilingual mixes, respectively, is used to regulate learning dynamics. 
To optimize memory consumption, activation checkpointing is applied. Additionally, we leverage mixed-precision techniques, utilizing bfloat16 for computations while maintaining FP32 for gradient accumulation to balance efficiency and accuracy.

\begin{table}[ht]
\caption{Continual pre-trained models and settings.}
\label{tab:models_settings}
\setlength{\tabcolsep}{1pt}
\tiny
\centering
\begin{tabular}{lllllll}
\toprule
\textbf{Base Model}        & \textbf{Data Mix} & \textbf{Our Model Name}            & \textbf{Steps} & \textbf{Warmup} & \textbf{Tokens} \\
\midrule
\multirow{2}{*}{Llama 3}   & Monolingual (\cref{fig:mix_monolingual})      & \EMMA Llama 3 Mono   & 25,000         & 1,000            & 419B            \\
                                                & Bilingual (\cref{fig:mix_bilingual})         & \EMMA Llama 3 Bi     & 40,000         & 2,000            & 671B            \\
\multirow{2}{*}{Llama 3.1} & Monolingual (\cref{fig:mix_monolingual})       & \EMMA Llama 3.1 Mono & 25,000         & 1,000            & 419B            \\
                                                & Bilingual (\cref{fig:mix_bilingual})        & \EMMA Llama 3.1 Bi   & 40,000         & 2,000            & 671B           \\
\bottomrule
\end{tabular}
\end{table}

\section{Evaluation and Discussion}
\label{sec:evaluation}

This section is structured to systematically evaluate the overall performance of our models on multilingual and bilingual benchmarks, assessing improvements in both text understanding and generation.
We analyze the impact of bilingual continual pre-training on multilingual language models and conduct ablation studies to isolate the contributions of bilingual pre-training compared to monolingual training.
We also focus on low-resource language performance, demonstrating how continual pre-training enhances representation and generalization for underrepresented languages.
No single model can be universally the best among all baselines across the full spectrum of multilingual tasks, benchmarks, and languages.
We provide an analysis of language gains and failure cases, identifying which languages benefit the most and highlighting remaining challenges.

\paragraph{Tasks and Benchmarks}
We evaluate all models on 7 tasks and 12 benchmarks that cover from 10 to 1500 languages. 
\Cref{tab:downstream_tasks} shows the details of those tasks, benchmarks, evaluation metrics, the number of languages and samples per language, and the domain. 
We do not use LLMs-as-a-judge \citep{li2024llmsasjudgescomprehensivesurveyllmbased} for evaluation due to its well-known limitations, especially in multilingual scenarios, such as low agreement with human judges \citep{dewynter2024rtplxllmsevaluatetoxicity}.

\paragraph{Baselines}
We consider open-weight decoder-only models with 7-9B parameters as baselines. We primarily compare our CPT models with the original Llama 3/3.1 base models \citep{dubey2024llama3} and the LlaMAX models \citep{lu2024llamax} continually trained from Llama 3. 
We also compare with a wide set of baselines (\Cref{sec:baselines}), including (1) Llama 2 models and their CPT models; (2) multilingual models, including recent advances such as Aya 23 \citep{aryabumi2024aya23}, Gemma \citep{gemmateam2024gemma2improvingopen}, Qwen \citep{yang2024qwen2technicalreport}, and Marco-LLM \citep{ming2024marcollmbridginglanguagesmassive}.

\subsection{Impact of Mono- vs. Bilingual Training}
\label{sec:analysis_bilingual_vs_monolingual}

\begin{figure}[ht!]
    \centering
    \begin{subfigure}[b]{0.45\textwidth}
        \includegraphics[width=\textwidth]{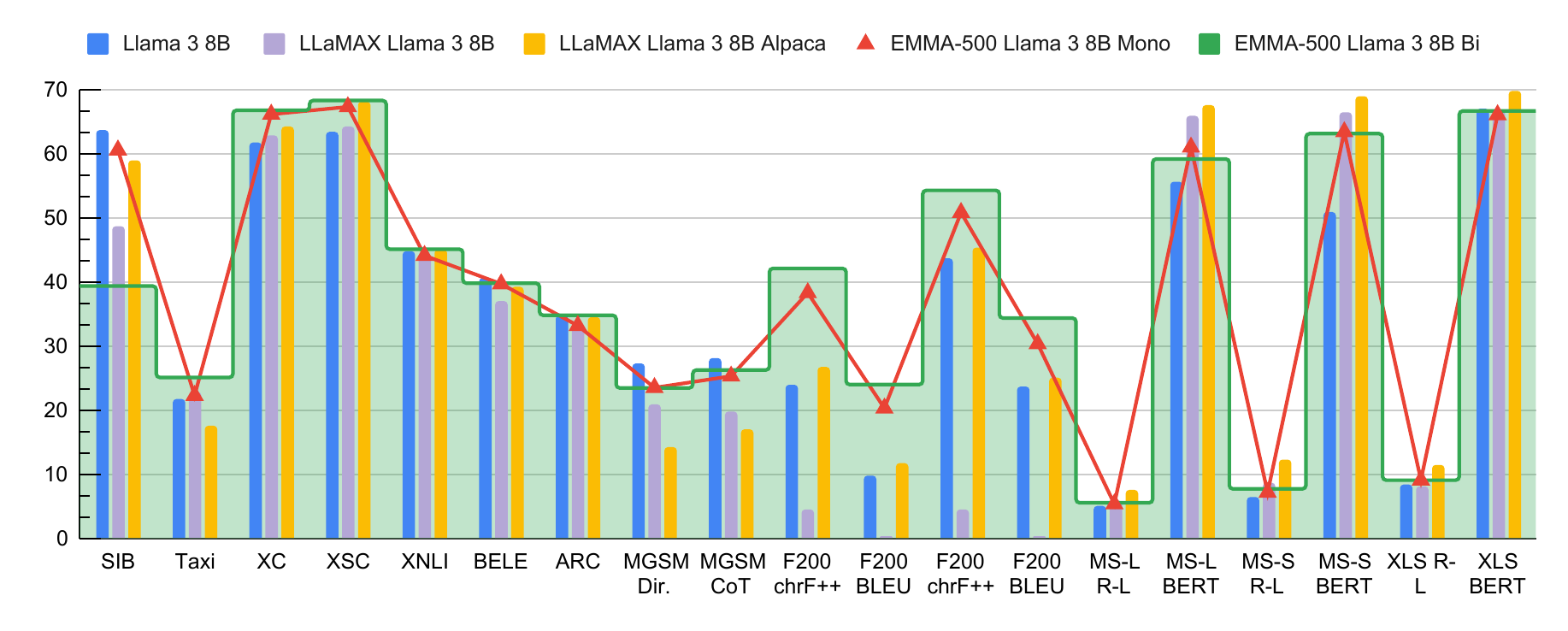}
        \caption{Llama 3}
        \label{fig:mono-vs-bi-l3}
    \end{subfigure}
    \hfill
    \begin{subfigure}[b]{0.45\textwidth}
        \includegraphics[width=\textwidth]{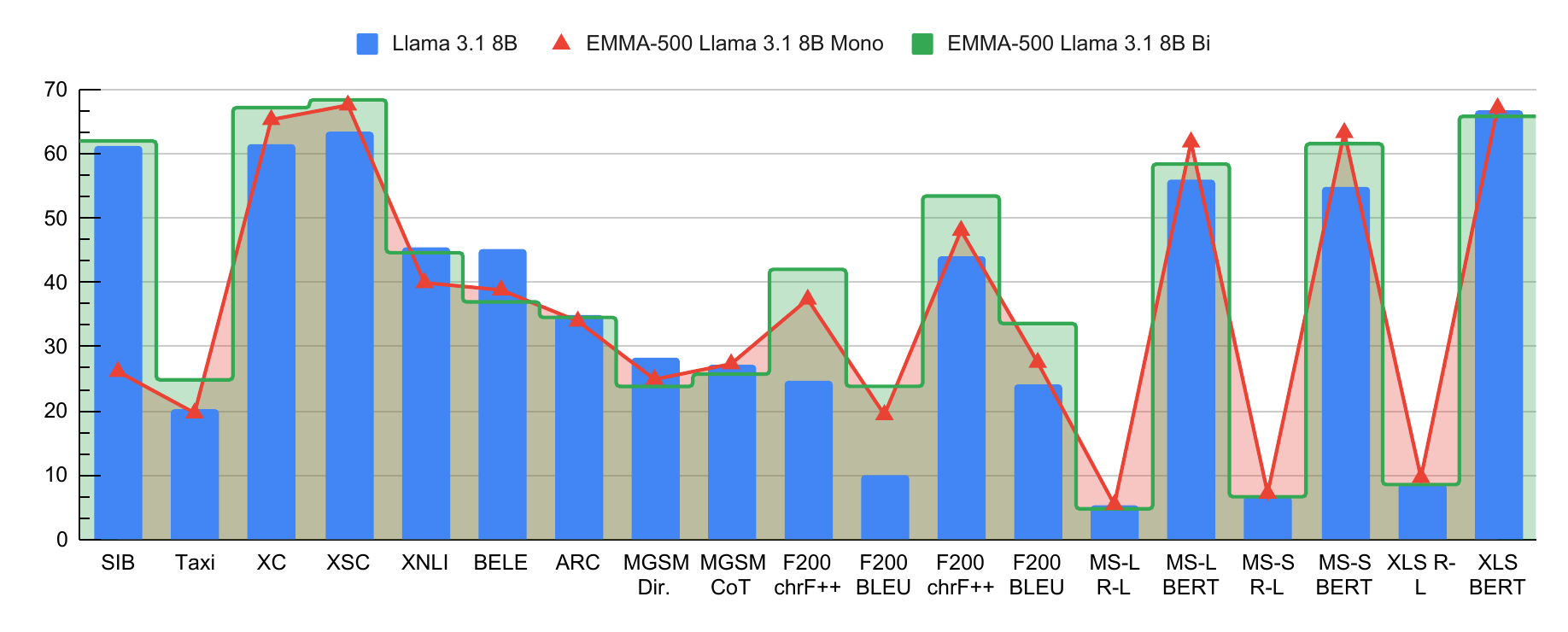}
        \caption{Llama 3.1}
        \label{fig:mono-vs-bi-l3.1}
    \end{subfigure}
    \caption{Comparison of monolingual and bilingual CPT. The scores are averaged across all evaluated languages of the corresponding benchmarks. The baseline model LlaMAX does not have a CPT variant trained on Llama 3.1. Our models show a tendency for bilingual CPT to be better than monolingual CPT in most benchmarks and a remarkable advance on the Flores200 translation benchmark.}
    \label{fig:mono-vs-bi}
\end{figure}

To understand the role of continual training with bilingual translation data, we conduct controlled ablation studies by comparing
monolingual continual pre-training (with the monolingual corpus only, i.e., data mix 2 in \Cref{fig:mix_monolingual}) vs. bilingual continual pre-training (with monolingual and bilingual extension, i.e., data mix 1 in \Cref{fig:mix_bilingual}).
\Cref{fig:mono-vs-bi} compares their average performance on each benchmark. 

For CPT with Llama 3 as shown in \Cref{fig:mono-vs-bi-l3}, continual training on data mix with bilingual translation data consistently improves commonsense reasoning, natural language inference, reading comprehension, machine translation, and sometimes improves text classification when evaluated on Taxi1500, and retains similar summarization and math performance to the original base models. 

For CPT with Llama 3.1 as shown in \Cref{fig:mono-vs-bi-l3.1}, continual training on data mix with bilingual translation data consistently improves text classification, commonsense reasoning, and machine translation. Except for summarization tasks, our CPT models trained on the monolingual mix are usually better than on the bilingual mix. 
More remarkably, CPT with both Llama 3 and 3.1 shows a large improvement on machine translation with an increase from 9\% to 140\% in terms of BLEU or chrF++ scores on translation directions from and to English on the Flores200 dataset. 

Our study provided insights into how bilingual texts contribute to language adaptation, transferability, and performance stability.
Overall, \textbf{continual training with bilingual translation data tends to improve the multilingual performance}, especially for machine translation, which benefits directly from the use of parallel texts in training. 

\subsection{Low-Resource Language Performance}

Low-resource languages often struggle with data scarcity and representation biases in multilingual models. 
We evaluate how baselines and our models perform on low-resource languages, and analyze whether continual pre-training on our data mixtures, especially the bilingual extension, can mitigate these limitations.

We focus on two benchmarks, SIB-200 and Flores200, with more than 200 languages, and one benchmark, Taxi1500, with more than 1500 languages, and examine the low-resource languages of these benchmarks according to the categorization in \Cref{tab:languages}.  
\Cref{tab:low-resource} shows the average performance on low-resource languages.  
Our \EMMA Llama 3 bilingual model obtained the best translation performance, followed by our Llama 3.1 bilingual one. 
They surpass the advanced Macro-LLM by a large margin on English to other translations and a small margin on other to English translations.
For text classification, our models experience different levels of drops in the low-resource languages of SIB-200. 
Notably, our bilingual models perform the best on Taxi1500, showing their outstanding performance on low-resource languages validated on this massively multilingual benchmark. 
The results on low-resource languages show that \textbf{CPT on massive parallel data enhances the performance on low-resource languages}, especially for massively multilingual classification and translation.

\begin{table}[ht!]
\caption{Performance on low-resource languages. The text classification task uses SIB-200 (SIB) and Taxi1500 (Taxi) datasets. \textbf{\underline{Underline and bold}} represents the absolute best, \underline{underline} means the second best, and \textbf{bold} signifies the best within a specific group. Our \EMMA models trained on bilingual data are the best two models in most cases.}
\label{tab:low-resource}
\setlength{\tabcolsep}{1pt}
\centering
\tiny
\begin{tabular}{l|rr|rrrr}
\toprule
\multirow{2}{*}{\textbf{Model}}            & \multicolumn{1}{c}{\textbf{SIB}} & \multicolumn{1}{c|}{\textbf{Taxi}} & \multicolumn{2}{c}{\textbf{Flores200 (Eng-X)}} & \multicolumn{2}{c}{\textbf{Flores200 (X-Eng)}}  \\
\cmidrule(lr){2-3}\cmidrule(lr){4-7}
                                  & ACC   & ACC  & chrF++& BLEU  & chrF++& BLEU       \\ \hline
Llama 2 7B                        & 22.74 & 18.04 & 16.23  & 4.93  & 31.26  & 13.69 \\
Llama 2 7B Chat                   & 26.02 & 16.01 & 18.12  & 5.30  & 32.61  & 13.05 \\
CodeLlama 2 7B                    & 23.84 & 17.15 & 15.98  & 4.57  & 29.47  & 11.52 \\
LLaMAX Llama 2 7B                 & 10.78 & 23.53 & 7.92   & 0.86  & 14.08  & 2.08  \\
LLaMAX Llama 2 7B Alpaca          & 28.66 & 15.57 & 30.45  & 13.36 & 44.08  & 23.85 \\
MaLA-500 Llama 2 10B v1           & 23.13 & \textbf{24.76} & 6.64   & 0.65  & 13.89  & 2.55  \\
MaLA-500 Llama 2 10B v2           & 19.04 & 22.72 & 6.93   & 0.55  & 15.76  & 3.17  \\
YaYi Llama 2 7B                   & 25.03 & 17.98 & 16.04  & 4.81  & 32.30  & 13.76 \\
TowerBase Llama 2 7B              & 19.53 & 18.06 & 17.15  & 5.10  & 32.44  & 14.50 \\
TowerInstruct Llama 2 7B          & 20.81 & 17.92 & 16.84  & 3.54  & 26.32  & 5.24  \\
EMMA-500 Llama 2 7B               & \textbf{31.97} & 21.73 & \textbf{35.88}  & \textbf{16.87} & \textbf{48.00}  & \textbf{27.34} \\
\hline
Occiglot Mistral 7B v0.1          & 33.27 & 22.62 & 17.23  & 4.69  & 32.13  & 14.00 \\
Occiglot Mistral 7B v0.1 Instruct & 34.87 & 19.59 & 16.92  & 4.31  & 32.61  & 12.31 \\
BLOOM 7B                          & 17.73 & 14.93 & 12.32  & 2.67  & 28.51  & 9.96  \\
BLOOMZ 7B                         & 29.99 & 17.00 & 16.83  & 7.48  & 35.63  & 21.06 \\
YaYi 7B                           & 36.40 & 16.19 & 14.20  & 4.23  & 21.25  & 4.72  \\
Aya 23 8B                         & 42.20 & 22.52 & 16.89  & 6.41  & 33.31  & 14.48 \\
Aya Expanse 8B                    & 58.06 & 19.08 & 25.13  & 6.78  & 37.85  & 13.61 \\
Gemma 7B                          & 59.97 & 16.55 & 24.75  & 9.56  & 45.65  & 25.47 \\
Gemma 2 9B                        & 47.19 & 21.48 & \textbf{28.50}  & \textbf{12.71} & 41.36  & 25.19 \\
Qwen 1.5 7B                       & 48.94 & 8.18  & 18.87  & 5.93  & 37.01  & 16.42 \\
Qwen 2 7B                         & 56.06 & \textbf{23.01} & 18.17  & 5.57  & 38.90  & 18.40 \\
Qwen 2.5 7B                       & 54.85 & 17.82 & 18.46  & 5.71  & 40.11  & 19.97 \\
Marco-LLM GLO 7B                  & \underline{65.07} & 21.97 & 24.79  & 9.51  & \textbf{45.89}  & \textbf{26.38} \\
\hline
Llama 3 8B                        & \textbf{\underline{65.53}} & \textbf{23.61} & 25.84  & 10.51 & 45.72  & 25.49 \\
Llama 3.1 8B                      & 63.17 & 22.02 & 26.46  & 10.66 & 46.13  & 25.91 \\
LLaMAX Llama 3 8B                 & 49.70 & 22.91 & 4.93   & 0.49  & 4.87   & 0.52  \\
LLaMAX Llama 3 8B Alpaca          & 60.91 & 20.48 & \textbf{28.94}  & \textbf{12.44} & \textbf{47.43}  & \textbf{26.91} \\
\hline
\EMMA Llama 3 8B Mono          & 62.34 & 22.16 & 41.29  & 22.03 & 53.16  & 32.55 \\
\EMMA Llama 3 8B Bi            & 39.65 & \textbf{\underline{26.72}} & \textbf{\underline{45.22}}  & \textbf{\underline{25.99}} & \textbf{\underline{56.69}}  & \textbf{\underline{36.72}} \\
\EMMA Llama 3.1 8B Mono        & 26.17 & 19.62 & 40.32  & 21.06 & 50.34  & 29.51 \\
\EMMA Llama 3.1 8B Bi          & \textbf{63.33} & \underline{25.42} & \underline{45.07}  & \underline{25.80} & \underline{55.84}  & \underline{35.91} \\
\bottomrule
\end{tabular}
\end{table}

\subsection{Model Adaptability}
\label{sec:analysis_model_adpatability}

It is unrealistic to expect a single model to perform the best across all tasks and benchmarks, given the inherent trade-offs in multilingual generalization, task specialization, and resource distribution.
In practice, we experience some performance drop of our models on certain benchmarks. 
In this section, we analyze model adaptability by examining how CPT impacts performance when applied to different base models across a wide range of languages. 
To quantify the effect, we compute the performance gain as the difference between the CPT model and its corresponding base model on each benchmark. 

\begin{figure}[ht!]
   \centering
       \includegraphics[width=0.9\columnwidth]{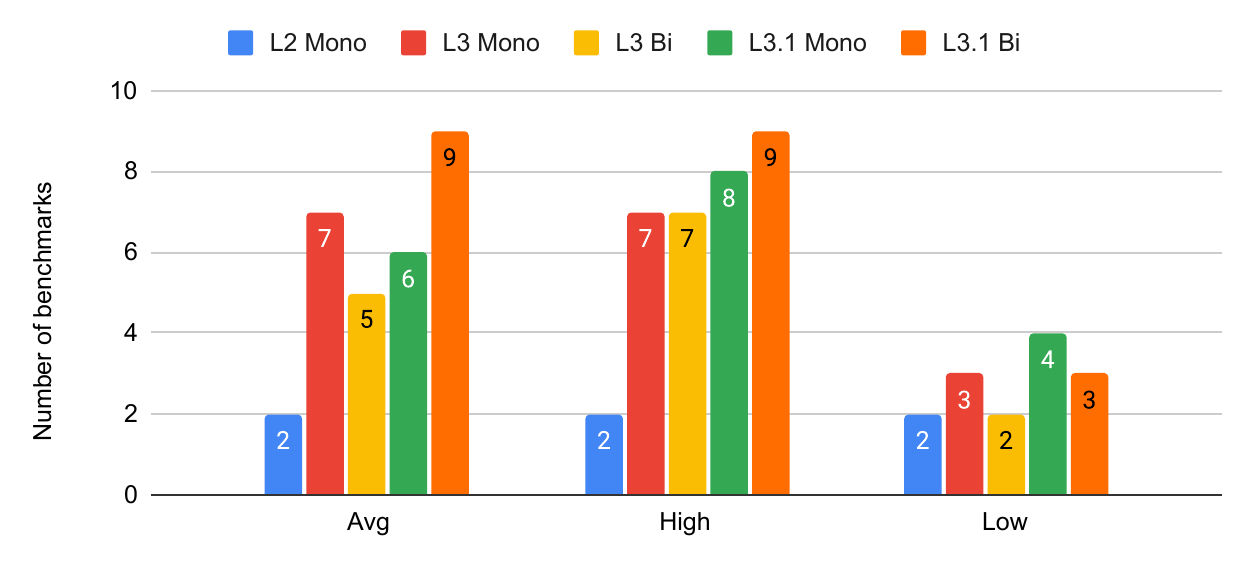}
   \caption{Model adaptability measured by the number of benchmarks on which CPT models are worse than the base model. CPT on LLaMA 2 (L2 Mono) shows a negligible BERTScore drop on MassiveSumm. More highly optimized models such as LLaMA 3 and 3.1 present greater challenges for effective continual pre-training compared to LLaMA 2, especially for high-resource languages, while the situation slightly eases for low-resource languages.}
   \label{fig:model_adaptability}
\end{figure}

We compare LLaMA 2, LLaMA 3, and LLaMA 3.1 as base models for continual pre-training, each offering progressively larger and more recent training corpora. 
LLaMA 2 was trained on 2T tokens, while LLaMA 3 significantly expands coverage with over 15T tokens. 
LLaMA 3.1 further updates the model with extensive data and long-context fine-tuning.
This progression allows us to assess the adaptability of CPT across different model generations and data scales.
\Cref{fig:model_adaptability} shows the performance difference and the number of benchmarks where CPT models experience degradation.

CPT on Llama 2 observes a very small drop of BERTScore on MassiveSumm, where the score of the long subset decreases from 63.89 to 63.80 and the short one from 65.35 to 65.14.
However, for Llama 3 and 3.1, both CPT on monolingual and bilingual mixes observe more cases of performance drops. 
For high-resource languages, Llama 3.1 degrades across more benchmarks than Llama 3. 
However, this trend diminishes notably when evaluating low-resource languages, where CPT remains effective to some extent.

This comparison provides insight into the effectiveness of continual pre-training with monolingual and bilingual data mixes in adapting different LLMs with various training to diverse linguistic settings.
Our results align with the findings from \citet{springer2025overtrained} that over-trained language models are harder to fine-tune. 
At a massively multilingual scale, we corroborate that \textbf{continuing pre-training on well-trained models}---particularly those already optimized for high-resource languages, e.g., Llama 3 and 3.1---\textbf{poses significant challenges when extending to hundreds of additional languages}.

\subsection{Per-Language Performance}
\label{sec:analysis_gains_limitations}

\begin{table*}[ht!]
\caption{Percentage of top-$k$ rankings across four multilingual benchmarks. The darker color indicates the better. \EMMA---especially with bilingual training---consistently outperforms baselines on Taxi1500 and Flores200. While Llama 3 and LlaMAX models yield higher accuracy on BELEBELE, \EMMA models cover more languages with improved relative performance.}
\label{tab:analysis_percentage}
\setlength{\tabcolsep}{3pt}
\centering
\scriptsize
\begin{tabular}{l|rrr|rrr|rrr|rrr}
\toprule
                           & \multicolumn{3}{c|}{\textbf{BELEBELE}}                                                             & \multicolumn{3}{c|}{\textbf{SIB-200}}                                                               & \multicolumn{3}{c|}{\textbf{Taxi1500}}                                                              & \multicolumn{3}{c}{\textbf{Flores200}}                                                               \\ \cline{2-13} 
\multirow{-2}{*}{Models}   & \textbf{Top 3}                 & \textbf{Top 5}                  & \textbf{Top 10}                 & \textbf{Top 3}                  & \textbf{Top 5}                  & \textbf{Top 10}                 & \textbf{Top 3}                  & \textbf{Top 5}                  & \textbf{Top 10}                 & \textbf{Top 3}                  & \textbf{Top 5}                  & \textbf{Top 10}                  \\
\midrule
Llama 3 8B                 & \cellcolor[HTML]{CBCEFB}0.82 & \cellcolor[HTML]{CBCEFB}0.82  & \cellcolor[HTML]{CBCEFB}25.41 & \cellcolor[HTML]{FCFF2F}43.41 & \cellcolor[HTML]{FCFF2F}65.37 & \cellcolor[HTML]{FCFF2F}95.61 & \cellcolor[HTML]{FFCE93}7.17  & \cellcolor[HTML]{FFCE93}13.80 & \cellcolor[HTML]{FFCE93}38.16 & \cellcolor[HTML]{9AFF99}0.00  & \cellcolor[HTML]{9AFF99}0.00  & \cellcolor[HTML]{9AFF99}25.12  \\
LLaMAX Llama 3 8B          & 0.00                         & 1.64                          & 11.48                         & 0.49                          & 7.32                          & 35.12                         & 6.10                          & 19.38                         & 58.73                         & 0.00                          & 0.00                          & 0.99                           \\
LLaMAX Llama 3 8B Alpaca   & 0.00                         & 1.64                          & 20.49                         & 5.37                          & 23.90                         & 77.07                         & 3.45                          & 6.10                          & 10.62                         & 0.00                          & 0.00                          & 44.83                          \\ \hline
\EMMA Llama 3 8B Mono   & \cellcolor[HTML]{9698ED}2.46 & \cellcolor[HTML]{9698ED}9.02  & \cellcolor[HTML]{9698ED}37.70 & \cellcolor[HTML]{FFFFC7}30.24 & \cellcolor[HTML]{FFFFC7}43.41 & \cellcolor[HTML]{FFFFC7}76.10 & \cellcolor[HTML]{FE996B}14.60 & \cellcolor[HTML]{FE996B}22.50 & \cellcolor[HTML]{FE996B}44.79 & \cellcolor[HTML]{32CB00}1.97  & \cellcolor[HTML]{32CB00}77.34 & \cellcolor[HTML]{32CB00}96.06  \\
\EMMA Llama 3 8B Bi     & \cellcolor[HTML]{9698ED}4.10 & \cellcolor[HTML]{9698ED}13.93 & \cellcolor[HTML]{9698ED}44.26 & \cellcolor[HTML]{FFFFC7}2.93  & \cellcolor[HTML]{FFFFC7}10.24 & \cellcolor[HTML]{FFFFC7}20.49 & \cellcolor[HTML]{FE996B}32.45 & \cellcolor[HTML]{FE996B}47.38 & \cellcolor[HTML]{FE996B}78.23 & \cellcolor[HTML]{32CB00}88.18 & \cellcolor[HTML]{32CB00}98.52 & \cellcolor[HTML]{32CB00}100.00 \\
\bottomrule                        
\end{tabular}
\end{table*}

Despite notable improvements, multilingual adaptation continues to present challenges. 
To scrutinize per-language performance, we evaluate each model by (1) counting the number of baseline models it outperforms across languages (from a total of 32 baselines; see \Cref{sec:baselines}), and (2) calculating the percentage of languages where the model ranks among the top $k$ performers. 
This provides a fine-grained view of cross-lingual competitiveness.

\Cref{tab:analysis_percentage} presents the top-$k$ performance percentage across four multilingual benchmarks—BELEBELE, SIB-200, Taxi1500, and FLORES-200.
The results demonstrate that \EMMA models, particularly those trained with bilingual data, achieve higher top-$k$ percentages on low-resource-heavy benchmarks like Taxi-1500 and FLORES-200. 
In contrast, base LLaMA 3 and LLaMAX variants exhibit more limited top-tier rankings. 
Notably, for BELEBELE, a machine comprehension benchmark, Llama 3 and 3.1 are not very strong models on this benchmark as evaluated in \Cref{sec:detailed_mrc}.
Existing CPT models, like LlaMAX Llama 3, and our CPT models obtained degraded performance. 
Despite having a decreased average accuracy, \textbf{our models have better performance for more languages} than the base model and the LlaMAX Llama 3 model. 

We further look at the evaluation results of selected languages on BELEBELE in \Cref{tab:analysis_belebele} for a mix of low-resource (e.g., Khalkha Mongolian, Northern Sotho, Plateau Malagasy) and high-resource (e.g., English, French, Italian) languages. 
\Cref{tab:analysis_belebele} reports absolute scores and the delta ($\Delta$) between each CPT model and the Llama 3 8B base model. 
Results show that while CPT variants generally improve scores for low-resource languages---especially in the \EMMA setups---they tend to suffer from notable performance degradation in high-resource languages. 
However, this trade-off is arguably acceptable as the main focus of this paper is to adapt LLMs into low-resource languages.  

These analyses provide a deeper understanding of the strengths and limitations of massively multilingual adaptation and guide future improvements in multilingual language model training and evaluation.
First, they show averaging scores across languages to compare model performance, while convenient, has several important limitations, e.g., obscuring disparities in performance across languages and resources, and ignoring per-language variance.
Second, this comparison of per-language performance highlights the benefits of our CPT models in promoting cross-lingual competitiveness.

\begin{table}[ht!]
\caption{Performance of selected languages in the BELEBELE benchmark. $\Delta$ denotes the difference between the CPT model's score and that of the base model. All models are based on Llama 3 8B. CPT on Llama 3 exhibits a significant drop in some high-resource languages, much bigger than the increase in some low-resource languages.
}
\label{tab:analysis_belebele}
\setlength{\tabcolsep}{1pt}
\centering
\tiny
\begin{tabular}{l|l|lll|lll}
\toprule
Model                        & Average                                               & khk\_Cyrl                                         & nso\_Latn                                        & plt\_Latn                                        & eng\_Latn                                          & fra\_Latn                                          & ita\_Latn                                          \\
\midrule
Llama 3 8B                   & 40.73                                             & 35.11                                             & 29.22                                            & 30.44                                            & 74.56                                              & 60.89                                              & 60.22                                              \\
\hline
LLaMAX           & 36.96                                             & 34.67                                             & 30.33                                            & 34.11                                            & 62.33                                              & 48.78                                              & 47.33                                              \\
\multicolumn{1}{r}{$\Delta$} & \multicolumn{1}{r}{\cellcolor[HTML]{EFEFEF}-3.77} & \multicolumn{1}{r}{\cellcolor[HTML]{EFEFEF}-0.44} & \multicolumn{1}{r}{\cellcolor[HTML]{EFEFEF}1.11} & \multicolumn{1}{r}{\cellcolor[HTML]{EFEFEF}3.67} & \multicolumn{1}{r}{\cellcolor[HTML]{EFEFEF}-12.23} & \multicolumn{1}{r}{\cellcolor[HTML]{EFEFEF}-12.11} & \multicolumn{1}{r}{\cellcolor[HTML]{EFEFEF}-12.89} \\
LLaMAX Alpaca     & 39.41                                             & 35.56                                             & 30.78                                            & 33.78                                            & 65.78                                              & 54.67                                              & 53.67                                              \\
\multicolumn{1}{r}{$\Delta$} & \multicolumn{1}{r}{\cellcolor[HTML]{EFEFEF}-1.32} & \multicolumn{1}{r}{\cellcolor[HTML]{EFEFEF}0.45}  & \multicolumn{1}{r}{\cellcolor[HTML]{EFEFEF}1.56} & \multicolumn{1}{r}{\cellcolor[HTML]{EFEFEF}3.34} & \multicolumn{1}{r}{\cellcolor[HTML]{EFEFEF}-8.78}  & \multicolumn{1}{r}{\cellcolor[HTML]{EFEFEF}-6.22}  & \multicolumn{1}{r}{\cellcolor[HTML]{EFEFEF}-6.55}  \\
\EMMA Mono     & 39.73                                             & 38.44                                             & 37.00                                            & 34.78                                            & 56.00                                              & 51.11                                              & 49.56                                              \\
\multicolumn{1}{r}{$\Delta$} & \multicolumn{1}{r}{\cellcolor[HTML]{EFEFEF}-1.00} & \multicolumn{1}{r}{\cellcolor[HTML]{EFEFEF}3.33}  & \multicolumn{1}{r}{\cellcolor[HTML]{EFEFEF}7.78} & \multicolumn{1}{r}{\cellcolor[HTML]{EFEFEF}4.34} & \multicolumn{1}{r}{\cellcolor[HTML]{EFEFEF}-18.56} & \multicolumn{1}{r}{\cellcolor[HTML]{EFEFEF}-9.78}  & \multicolumn{1}{r}{\cellcolor[HTML]{EFEFEF}-10.66} \\
\EMMA Bi       & 39.84                                             & 39.33                                             & 35.22                                            & 33.00                                            & 56.78                                              & 49.67                                              & 46.00                                              \\
\multicolumn{1}{r}{$\Delta$} & \multicolumn{1}{r}{\cellcolor[HTML]{EFEFEF}-0.89} & \multicolumn{1}{r}{\cellcolor[HTML]{EFEFEF}4.22}  & \multicolumn{1}{r}{\cellcolor[HTML]{EFEFEF}6.00} & \multicolumn{1}{r}{\cellcolor[HTML]{EFEFEF}2.56} & \multicolumn{1}{r}{\cellcolor[HTML]{EFEFEF}-17.78} & \multicolumn{1}{r}{\cellcolor[HTML]{EFEFEF}-11.22} & \multicolumn{1}{r}{\cellcolor[HTML]{EFEFEF}-14.22} \\
\bottomrule
\end{tabular}
\end{table}

\subsection{Overall Results}
\label{sec:overall_results}

We evaluate the model on a comprehensive set of multilingual and bilingual benchmarks, assessing both language understanding and generation tasks.
\Cref{tab:overall_deterministic,tab:overall_generation} 
present the results of deterministic and generation tasks.
Our \EMMA models are the best at machine translation (Flores200) and competitive at text classification (Taxi1500 and SIB-200) and commonsense reasoning (XCOPA and XStoryCloze).
While our models are competitive in many cases, we also observe some performance degradation, such asin  math tasks 
Performance inevitably varies across tasks and languages, reflecting trade-offs in model capacity and data representation.
\Cref{sec:detailed_results} presents a deeper dive into the performance details across all tasks and languages evaluated, providing insights into model behavior across multilingual datasets and benchmarks.
We also provide detailed per-language results, available at \website

\begin{table*}[ht!]
\caption{Overall results of deterministic tasks including text classification, commonsense reasoning, natural language inference, reading comprehension, and math reasoning. XC: XCOPA; XSC: XStoryCloze; BELE: BELEBELE; ARC: the multilingual AI2 Reasoning Challenge; Dir.: MGSM by direct prompting; CoT: MGSM by CoT prompting. \textbf{\underline{Underline and bold}} represents the absolute best, \underline{underline} means the second best, and \textbf{bold} signifies the best within a specific group. Our \EMMA models are among the top 2 models on text classification and commonsense reasoning.}
\label{tab:overall_deterministic}
\setlength{\tabcolsep}{7pt}
\centering
\scriptsize
\begin{tabular}{l|rr|rr|r|rr|rr}
\toprule
\multirow{2}{*}{\textbf{Model}}            & \multicolumn{2}{c|}{\textbf{Classification}} & \multicolumn{2}{c|}{\textbf{Commonsense}} & \multicolumn{1}{c|}{\textbf{NLI}} & \multicolumn{2}{c|}{\textbf{Comprehension}} & \multicolumn{2}{c}{\textbf{Math}} \\ 
\cmidrule(lr){2-3}\cmidrule(lr){4-5}\cmidrule(lr){6-6}\cmidrule(lr){7-8}\cmidrule(lr){9-10}
                             &   SIB &  Taxi &    XC &   XSC  &  XNLI &  BELE &   ARC &  Dir. &   CoT \\
\midrule
Llama 2 7B                        & 22.41                      & 17.54                     & 56.67                    & 57.55                    & 40.19                                       & 26.27                     & 27.56                     & 6.69                 & 6.36                 \\
Llama 2 7B Chat                   & 25.58                      & 15.44                     & 55.85                    & 58.41                    & 38.58                                       & \textbf{29.05}                     & 28.02                     & 10.22                & 10.91                \\
CodeLlama 2 7B                    & 23.35                      & 17.03                     & 54.69                    & 55.68                    & 40.19                                       & 27.38                     & 25.23                     & 5.93                 & 6.64                 \\
LLaMAX Llama 2 7B                 & 10.61                      & 23.52                     & 54.38                    & 60.36                    & 44.27                                       & 23.09                     & 26.09                     & 3.35                 & 3.62                 \\
LLaMAX Llama 2 7B Alpaca          & 27.89                      & 15.09                     & 56.60                    & 63.83                    & 45.09                                       & 24.48                     & 31.06                     & 5.05                 & 6.35                 \\
MaLA-500 Llama 2 10B v1           & 23.25                      & \underline{\textbf{25.27}}                     & 53.09                    & 53.07                    & 38.11                                       & 22.96                     & 21.16                     & 0.91                 & 0.73                 \\
MaLA-500 Llama 2 10B v2           & 19.30                      & 23.39                     & 53.09                    & 53.07                    & 38.11                                       & 22.96                     & 21.16                     & 0.91                 & 0.73                 \\
YaYi Llama 2 7B                   & 24.57                      & 17.73                     & 56.71                    & 58.42                    & 41.28                                       & 28.32                     & 28.40                     & 7.09                 & 7.22                 \\
TowerBase Llama 2 7B              & 19.34                      & 17.73                     & 56.33                    & 57.78                    & 39.84                                       & 26.36                     & 27.94                     & 6.15                 & 6.16                 \\
TowerInstruct Llama 2 7B          & 20.53                      & 17.29                     & 57.05                    & 59.24                    & 40.36                                       & 27.93                     & \textbf{30.10}                     & 7.24                 & 8.24                 \\
EMMA-500 Llama 2 7B               & \textbf{31.27}                      & 19.82                     & \textbf{63.11}                    & \textbf{66.38}                    & \textbf{45.14}                                       & 26.75                     & 29.53                     & \textbf{17.02}                & \textbf{18.09}                \\
\hline
Occiglot Mistral 7B v0.1          & 32.69                      & 22.26                     & 56.67                    & 58.10                    & 42.35                                       & 30.16                     & 29.77                     & 13.31                & 14.07                \\
Occiglot Mistral 7B v0.1 Instruct & 34.31                      & 18.76                     & 56.55                    & 59.39                    & 40.81                                       & 32.05                     & 30.88                     & 22.76                & 22.16                \\
BLOOM 7B                          & 17.82                      & 14.76                     & 56.89                    & 59.30                    & 41.60                                       & 24.11                     & 23.65                     & 2.87                 & 2.29                 \\
BLOOMZ 7B                         & 29.73                      & 16.96                     & 54.87                    & 57.12                    & 37.13                                       & 39.32                     & 23.95                     & 2.55                 & 2.15                 \\
YaYi 7B                           & 35.76                      & 16.12                     & 56.64                    & 60.67                    & 39.87                                       & 37.97                     & 24.44                     & 2.76                 & 3.02                 \\
Aya 23 8B                         & 41.50                      & \textbf{22.64}                     & 55.13                    & 60.93                    & 43.12                                       & 40.08                     & 31.08                     & 22.29                & 24.71                \\
Aya Expanse 8B                    & 57.01                      & 18.73                     & 56.38                    & 64.80                    & \underline{\textbf{45.56}}                                       & 46.98                     & 36.56                     & 43.02                & 41.45                \\
Gemma 7B                          & 58.21                      & 13.83                     & 63.64                    & 65.01                    & 42.58                                       & 43.37                     & 38.68                     & 38.22                & 35.78                \\
Gemma 2 9B                        & 46.25                      & 18.05                     & 66.33                    & \textbf{67.67}                    & 46.74                                       & \underline{\textbf{54.49}}                     & \underline{\textbf{44.15}}                     & 32.95                & 44.69                \\
Qwen 1.5 7B                       & 47.95                      & 7.29                      & 59.44                    & 59.85                    & 39.47                                       & 41.83                     & 28.93                     & 31.56                & 30.36                \\
Qwen 2 7B                         & 54.95                      & 21.87                     & 60.31                    & 61.46                    & 42.77                                       & 49.31                     & 33.82                     & 48.95                & 51.47                \\
Qwen 2.5 7B                       & 53.89                      & 17.87                     & 61.71                    & 62.06                    & 43.31                                       & \underline{54.11}                     & 35.30                     & \underline{\textbf{53.78}}                & \underline{\textbf{55.60}}                \\
Marco-LLM GLO 7B                  & \underline{\textbf{64.15}}                      & 21.99                     & \textbf{62.45}                    & 63.87                    & 43.99                                       & 53.95                     & \underline{36.34}                     & \underline{51.85}                & \underline{52.02}                \\
\hline
Llama 3 8B                        & \textbf{63.70}                      & \textbf{21.73}                     & 61.71                    & 63.41                    & 44.97                                       & 40.73                     & 34.80                     & 27.45                & \textbf{28.13}                \\
Llama 3.1 8B                      & 61.42                      & 20.20                     & 61.71                    & 63.58                    & \underline{45.62}                                       & \textbf{45.19}                     & \textbf{34.93}                     & \textbf{28.36}                & 27.31                \\
LLaMAX Llama 3 8B                 & 48.60                      & 23.01                     & 63.04                    & 64.31                    & 44.13                                       & 36.96                     & 33.54                     & 20.80                & 19.96                \\
LLaMAX Llama 3 8B Alpaca          & 58.97                      & 17.71                     & \textbf{64.36}                    & \textbf{68.27}                    & 45.08                                       & 39.41                     & 34.53                     & 14.18                & 17.16                \\
\hline
\EMMA Llama 3 8B Mono          & 60.62                      & 22.32                     & 66.20                    & 67.36                    & 44.15                                       & 39.73                     & 33.22                     & 23.53                & 25.33                \\
\EMMA Llama 3 8B Bi            & 39.40                      & \underline{25.13}                     & \underline{66.82}                    & \underline{68.35}                    & \textbf{45.15}                                       & \textbf{39.84}                     & \textbf{34.84}                     & 23.49                & 26.29                \\
\EMMA Llama 3.1 8B Mono        & 26.16                      & 19.71                     & 65.38                    & 67.64                    & 39.98                                       & 38.86                     & 34.00                     & \textbf{24.95}                & \textbf{27.35}                \\
\EMMA Llama 3.1 8B Bi          & \underline{62.07}                      & 24.87                     & \underline{\textbf{67.25}}                    & \underline{\textbf{68.47}}                    & 44.67                                       & 37.00                     & 34.59                     & 23.85                & 25.76                \\
\bottomrule
\end{tabular}
\end{table*}

\begin{table*}[ht!]
\caption{Overall results of generation tasks including text summarization and machine translation. BERT: BERTScore; R-L: ROUGE-L. \textbf{\underline{Underline and bold}} represents the absolute best, \underline{underline} means the second best, and \textbf{bold} signifies the best within a specific group. Our \EMMA models trained with bilingual mix are the best-performing models on machine translation.}
\label{tab:overall_generation}
\setlength{\tabcolsep}{5pt}
\centering
\scriptsize
\begin{tabular}{l|rrrr|rrrr|rr}
\toprule
\multirow{2}{*}{\textbf{Model}}            & \multicolumn{4}{c|}{\textbf{Flores200}} & \multicolumn{2}{c}{\textbf{MassiveSumm-L}} & \multicolumn{2}{c|}{\textbf{MassiveSumm-S}} & \multicolumn{2}{c}{\textbf{XL-Sum}} \\
\cmidrule(lr){2-5}\cmidrule(lr){6-7}\cmidrule(lr){8-9}\cmidrule(lr){10-11}
                                  & chrF++& BLEU  & chrF++& BLEU  & R-L  & BERT  & R-L   & BERT  & R-L   & BERT        \\ \hline
            
                       Llama 2 7B & 15.13 &  4.62 & 30.32 & 12.93 & 4.74 & 63.89 &  7.85 & 65.35 &  7.11 & 66.52 \\
                  Llama 2 7B Chat & 16.95 &  4.95 & 31.72 & 12.28 & 4.73 & 63.52 &  9.76 & 67.01 &  8.84 & 68.44 \\
                   CodeLlama 2 7B & 14.94 &  4.27 & 28.57 & 10.82 & 5.63 & 64.51 &  7.59 & 64.83 &  7.15 & 65.74 \\
                LLaMAX Llama 2 7B &  7.42 &  0.80 & 13.66 &  1.99 & 4.56 & 62.69 &  5.22 & 63.06 &  5.29 & 64.59 \\
         LLaMAX Llama 2 7B Alpaca & 28.35 & 12.51 & 42.27 & 22.29 & 4.61 & 62.76 & \underline{10.71} & \textbf{67.92} & \textbf{10.11} & \textbf{69.24} \\
          MaLA-500 Llama 2 10B v1 &  6.08 &  0.60 & 13.60 &  2.29 & 4.39 & 64.50 &  4.97 & 63.51 &  5.45 & 63.96 \\
          MaLA-500 Llama 2 10B v2 &  6.38 &  0.54 & 15.44 &  2.87 & 4.37 & \textbf{64.66} &  5.02 & 63.75 &  5.44 & 64.28 \\
                  YaYi Llama 2 7B & 14.87 &  4.41 & 31.38 & 12.98 & \textbf{4.98} & 64.17 &  7.80 & 65.24 &  7.98 & 67.21 \\
             TowerBase Llama 2 7B & 16.03 &  4.83 & 31.47 & 13.74 & 4.81 & 64.51 &  8.11 & 65.53 &  7.65 & 67.09 \\
         TowerInstruct Llama 2 7B & 15.64 &  3.23 & 25.43 &  4.81 & 4.82 & 64.61 & 10.14 & 67.76 &  8.89 & 68.46 \\
              EMMA-500 Llama 2 7B & \textbf{33.25} & \textbf{15.58} & \textbf{45.78} & \textbf{25.37} & 4.79 & 63.80 &  8.32 & 65.14 &  8.58 & 67.20 \\
\hline
         Occiglot Mistral 7B v0.1 & 16.10 &  4.32 & 31.13 & 13.12 & 5.14 & 63.95 &  8.16 & 63.65 &  7.33 & 66.20 \\
Occiglot Mistral 7B v0.1 Instruct & 15.80 &  3.99 & 31.65 & 11.61 & 5.16 & 63.50 &  7.82 & 63.79 &  8.31 & 66.96 \\
                         BLOOM 7B & 11.80 &  2.81 & 27.84 &  9.57 & 4.88 & 64.36 &  6.79 & 62.30 &  6.99 & 64.78 \\
                        BLOOMZ 7B & 16.10 &  7.44 & 34.74 & 20.22 & 2.91 & 57.20 &  3.28 & 29.75 & 11.15 & 69.82 \\
                          YaYi 7B & 13.50 &  4.37 & 21.36 &  4.82 & 4.95 & 64.24 &  8.28 & 65.44 & \underline{\textbf{12.06}} & 69.74 \\
                        Aya 23 8B & 16.15 &  6.46 & 32.36 & 13.87 & 6.33 & 65.94 &  8.43 & 65.85 &  8.68 & 66.79 \\
                   Aya Expanse 8B & 23.89 &  6.88 & 36.86 & 13.12 & \underline{7.44} & \underline{\textbf{67.66}} &  \textbf{9.24} & \underline{67.68} & 10.51 & 68.73 \\
                         Gemma 7B & 23.05 &  9.05 & 43.68 & 23.79 & 6.18 & 62.14 &  8.35 & 62.25 &  6.70 & 64.52 \\
                       Gemma 2 9B & \textbf{26.48} & \textbf{12.09} & 38.87 & 23.15 & 5.86 & 59.70 &  7.86 & 58.11 &  7.38 & 65.45 \\
                      Qwen 1.5 7B & 17.77 &  5.87 & 35.87 & 15.58 & 6.09 & 59.19 &  8.49 & 62.70 &  9.58 & 69.13 \\
                        Qwen 2 7B & 17.17 &  5.56 & 37.61 & 17.39 & 6.65 & 56.31 &  8.63 & 56.14 & 10.18 & 69.35 \\
                      Qwen 2.5 7B & 17.49 &  5.72 & 38.89 & 18.95 & \underline{7.44} & 61.62 &  9.04 & 58.91 & 10.41 & 69.69 \\
                 Marco-LLM GLO 7B & 23.34 &  9.27 & \textbf{44.57} & \textbf{25.17} & 6.57 & 57.15 &  8.10 & 57.45 & \underline{11.46} & \underline{\textbf{70.41}} \\
\hline
                       Llama 3 8B & 24.08 &  9.93 & 43.72 & 23.78 & 5.10 & 55.58 &  6.44 & 50.98 &  8.47 & 67.08 \\
                     Llama 3.1 8B & 24.69 & 10.11 & 44.10 & 24.19 & 5.41 & 56.09 &  6.77 & 54.96 &  8.57 & 66.97 \\
                LLaMAX Llama 3 8B &  4.65 &  0.45 &  4.66 &  0.48 & 6.00 & 65.90 &  8.77 & 66.46 &  8.28 & 66.72 \\
         LLaMAX Llama 3 8B Alpaca & \textbf{26.86} & \textbf{11.64} & \textbf{45.45} & \textbf{25.10} & \underline{\textbf{7.62}} & \underline{67.64} & \underline{\textbf{12.44}} & \underline{\textbf{68.95}} & \textbf{11.39} & \underline{69.99} \\
\hline
         \EMMA Llama 3 8B Mono & 38.34 & 20.33 & 50.86 & 30.38 & 5.38 & 61.06 &  7.18 & \textbf{63.46} &  9.11 & 66.12 \\
           \EMMA Llama 3 8B Bi & \textbf{\underline{42.15}} & \textbf{\underline{24.02}} & \textbf{\underline{54.33}} & \textbf{\underline{34.40}} & \textbf{5.57} & 59.23 &  \textbf{7.74} & 63.18 &  9.10 & 66.72 \\
       \EMMA Llama 3.1 8B Mono & 37.41 & 19.44 & 48.12 & 27.57 & 5.44 & \textbf{61.89} &  7.21 & 63.36 &  \textbf{9.66} & \textbf{67.21} \\
         \EMMA Llama 3.1 8B Bi & \underline{42.07} & \underline{23.86} & \underline{53.49} & \underline{33.64} & 4.78 & 58.47 &  6.67 & 61.64 &  8.56 & 65.90 \\
\bottomrule
\end{tabular}
\end{table*}

\section{Related Work}
\label{sec:related}

\paragraph{Multilingual Continual pre-training}
Multilingual language models have made significant progress in extending language understanding across diverse linguistic landscapes.
Models such as mT5 \citep{xue2021mt5} and BLOOM \citep{scao2022bloom} have demonstrated strong multilingual capabilities by pre-training on extensive multilingual corpora. However, its performance on low-resource languages remains limited.
Recent developments such as Qwen 2 \citep{yang2024qwen2technicalreport} and Llama 3 \citep{dubey2024llama3} have further improved the efficiency and scalability of LLMs and shown stronger multilingual capabilities.
Continual pre-training has emerged as an effective strategy for expanding language models’ linguistic and domain coverage without requiring full retraining. Prior works such as MaLA-500 \citep{lin2024mala}, LlaMAX \citep{lu2024llamax}, and EMMA-500 \citep{ji2024emma} have demonstrated the feasibility of continual pre-training on diverse multilingual datasets.
These approaches highlight the importance of incremental learning to incorporate new languages and improve cross-lingual transfer.

Despite these advancements, prior work has primarily focused on monolingual text pre-training, with limited emphasis on bilingual adaptation for improved cross-lingual transfer. Our work extends this line of research by introducing a bilingual translation corpus for continual pre-training and adapting Llama 3 and 3.1 (8B) to enhance bilingual and multilingual generalization, particularly for underrepresented languages.

\paragraph{Training on Bilingual Translation Data}
Parallel texts have been used for pre-training LLMs.
PolyLM \citep{wei2023polylmopensourcepolyglot} trained on 1 billion parallel multilingual data (0.16\% of pre-training corpora). 
Poro \citep{luukkonen2024poro34bblessingmultilinguality} trained on 8 billion English-Finnish cross-lingual texts (slightly under 1\% of pre-training corpora). 
\citet{li2024comparison} trained small language models, i.e., BERT \citep{devlin2019bert}, GPT-2 \citep{radford2019language}, and BART \citep{lewis2019bart}, in a controlled setting with multilingual and bilingual texts using different learning objectives. 
In continual pre-training or fine-tuning, bilingual texts are also widely used.
\citet{ji2024can} showed continual pre-training of multilingual BART \citep{tang2020multilingual} with machine translation failed enhance cross-lingual representation learning.
\citet{xu2024paradigmshiftmachinetranslation} fine-tuned LLMs with monolingual data and parallel data for machine translation. 
\citet{kondo2024enhancingtranslationaccuracylarge} showed that continual pre-training and supervised fine-tuning on parallel data can enhance the accuracy in English-Japanese translation. 
Similarly, \citet{fujii2024continualpretrainingcrosslingualllm} investigated cross-lingual continual pre-training on English and Japanese and showed that the use of parallel corpora enhanced translation performance.
\citet{li2025rethinkingmultilingualcontinualpretraining} studied the data mixing strategies for continual pre-training using monolingual and parallel texts and code data across 30+ languages.
While previous studies have explored bilingual and multilingual pre-training with parallel texts, most efforts have been limited to a small subset of languages or specific translation tasks. 
Unlike PolyLM and Poro, which incorporate a modest fraction of bilingual data into predominantly monolingual training, our work systematically integrates bilingual texts across 500+ languages.

\section{Conclusion}
\label{sec:conclusion}

This work advances massively multilingual adaptation of LLMs using bilingual translation data across 2,500+ language pairs (500+ languages). 
The four released \EMMA models using Llama 3 and 3.1 trained on both monolingual and bilingual data mixes, including the newly compiled \MALA translation corpus, establish new benchmarks in multilingual coverage while maintaining competitive performance across 7 diverse tasks. 
Notably, they achieve state-of-the-art results on machine translation while showing robust generalization to text classification and reasoning tasks.
Despite advancements, achieving consistently high performance across diverse benchmarks is constrained by linguistic and task variability, which highlights intrinsic tensions between scale and specialization in multilingual NLP.

\section*{Limitations}

\paragraph{Multilingual Benchmark}
Multilingual language models are designed to accommodate users across various linguistic and cultural backgrounds. 
However, many widely used multilingual evaluation benchmarks, including some in this research, are developed through human or machine translation. 
As a result, they often emphasize knowledge and subject matter primarily from English-speaking sources, with potential translation-related distortions that can undermine the accuracy of model assessments \citep{chen2024good}. 
This discrepancy highlights the need for more comprehensive, native-language test sets that better reflect the full range of linguistic diversity. 
We encourage collaborative efforts to create large-scale benchmarks that offer a more reliable evaluation of these models across different languages and cultures.

\paragraph{Human Evaluation}
Although human assessment is a valuable tool, it presents challenges such as variability, subjectivity, and high costs---especially when evaluating models across numerous languages. 
Recruiting expert annotators proficient in less common languages is particularly difficult, making large-scale human evaluation unfeasible. Even assessing a limited selection of languages requires significant resources. 
While this study acknowledges these constraints, we recognize that human evaluations play a crucial role in supplementing automated assessment methods. 
Given these limitations, we rely on automatic evaluation tools to ensure scalability and consistency, despite their imperfections.

\paragraph{Model Performance} Despite its strengths in multilingual processing, our models face challenges in areas such as mathematical reasoning and machine reading comprehension. 
Its performance on machine-translated mathematical benchmarks remains limited, likely due to the inherent difficulty of numerical reasoning and translation artifacts that may obscure problem clarity. 
Similarly, while our model achieves improvements in various NLP tasks, it struggles with reading comprehension, which demands deep contextual understanding and logical inference. 
Addressing these weaknesses will require further refinements, such as incorporating domain-specific training data or exploring alternative model architectures optimized for these challenges.
Besides, averaging scores across languages can mask important disparities, such as strong performance on high-resource languages and poor results on low-resource ones. 
It also ignores linguistic diversity and per-language difficulty, potentially leading to misleading conclusions about overall model performance.

\paragraph{Real-world Usage}
This research primarily focuses on enhancing continual training with bilingual texts and improving language model performance through continual pre-training. 
However, the model is not yet suitable for deployment in real-world applications. 
It has not undergone thorough human alignment processes or adversarial robustness testing (red-teaming) to ensure safety and reliability. 
While our work contributes to advancements in multilingual NLP, additional refinements---such as aligning the model with human preferences and conducting rigorous safety evaluations—are necessary before it can be practically implemented.

\paragraph{Data Mix}
We examine two manually constructed data mixes to assess the impact of CPT with bilingual translation data. 
While numerous alternative configurations are possible, systematically validating the effectiveness of different data mixes or searching for the optimal data mixes requires a lot of computing resources, which are not affordable for a small research team like us. 
Nonetheless, the two data mixes carefully decided by us already demonstrate measurable improvements in the downstream evaluation, which highlight the utility of our design choices. 

\paragraph{Model Training}
Model training involves a few hyperparameters, and a wide range of base models can be selected for continual pre-training.
When preparing the bilingual translation data for training, the number of lines is intuitively chosen. 
However, an exhaustive search over all possible configuration combinations—including data volume, model choice, and hyperparameter settings—would incur prohibitive computational costs. Consequently, we focus on a limited set of configurations that are feasible under available resources.

\paragraph{Community Collaboration}
This study was conducted without direct community collaboration. Nonetheless, we recognize the value of community collaboration and are open to future partnerships with researchers and practitioners in this area. 

\section*{Acknowledgment}
The work has received funding from the European Union's Horizon Europe research and innovation programme under grant agreement No 101070350 and from UK Research and Innovation (UKRI) under the UK government's Horizon Europe funding guarantee [grant number 10052546], and the Digital Europe Programme under grant agreement No 101195233.
The authors wish to acknowledge CSC – IT Center for Science, Finland, for computational resources.
We thank Indraneil Paul for his contributions to preparing the code data.

\bibliography{emma-500-bilingual}

\begin{thebibliography}{106}
\providecommand{\natexlab}[1]{#1}
\providecommand{\url}[1]{\texttt{#1}}
\expandafter\ifx\csname urlstyle\endcsname\relax
  \providecommand{\doi}[1]{doi: #1}\else
  \providecommand{\doi}{doi: \begingroup \urlstyle{rm}\Url}\fi

\bibitem[Aakanksha et~al.(2024)Aakanksha, Ahmadian, Ermis, Goldfarb-Tarrant,
  Kreutzer, Fadaee, and Hooker]{aakanksha2024aya_redteaming}
Aakanksha, A.~Ahmadian, B.~Ermis, S.~Goldfarb-Tarrant, J.~Kreutzer, M.~Fadaee,
  and S.~Hooker.
\newblock The multilingual alignment prism: Aligning global and local
  preferences to reduce harm, 2024.
\newblock URL \url{https://arxiv.org/abs/2406.18682}.

\bibitem[Adelani et~al.(2022)Adelani, Alabi, Fan, Kreutzer, Shen, Reid, Ruiter,
  Klakow, Nabende, Chang, et~al.]{lafand-mt}
D.~I. Adelani, J.~O. Alabi, A.~Fan, J.~Kreutzer, X.~Shen, M.~Reid, D.~Ruiter,
  D.~Klakow, P.~Nabende, E.~Chang, et~al.
\newblock A few thousand translations go a long way! leveraging pre-trained
  models for african news translation.
\newblock \emph{arXiv preprint arXiv:2205.02022}, 2022.
\newblock URL \url{https://aclanthology.org/2022.naacl-main.223/}.

\bibitem[Adelani et~al.(2023)Adelani, Liu, Shen, Vassilyev, Alabi, Mao, Gao,
  and Lee]{sib-200}
D.~I. Adelani, H.~Liu, X.~Shen, N.~Vassilyev, J.~O. Alabi, Y.~Mao, H.~Gao, and
  E.~A. Lee.
\newblock {SIB-200:} {A} simple, inclusive, and big evaluation dataset for
  topic classification in 200+ languages and dialects.
\newblock \emph{CoRR}, abs/2309.07445, 2023.
\newblock \doi{10.48550/arXiv.2309.07445}.
\newblock URL \url{https://doi.org/10.48550/arXiv.2309.07445}.

\bibitem[Alves et~al.(2024)Alves, Pombal, Guerreiro, Martins, Alves, Farajian,
  Peters, Rei, Fernandes, Agrawal, et~al.]{alves2024tower}
D.~M. Alves, J.~Pombal, N.~M. Guerreiro, P.~H. Martins, J.~Alves, A.~Farajian,
  B.~Peters, R.~Rei, P.~Fernandes, S.~Agrawal, et~al.
\newblock Tower: An open multilingual large language model for
  translation-related tasks.
\newblock \emph{arXiv preprint arXiv:2402.17733}, 2024.
\newblock URL \url{https://arxiv.org/abs/2402.17733}.

\bibitem[Amrhein et~al.(2022)Amrhein, Moghe, and Guillou]{aces}
C.~Amrhein, N.~Moghe, and L.~Guillou.
\newblock {ACES}: Translation accuracy challenge sets for evaluating machine
  translation metrics.
\newblock In \emph{Proceedings of the Seventh Conference on Machine Translation
  (WMT)}, pages 479--513, Abu Dhabi, United Arab Emirates (Hybrid), Dec. 2022.
  Association for Computational Linguistics.
\newblock URL \url{https://aclanthology.org/2022.wmt-1.44}.

\bibitem[Andonian et~al.(2023)Andonian, Anthony, Biderman, Black, Gali, Gao,
  Hallahan, Levy-Kramer, Leahy, Nestler, Parker, Pieler, Phang, Purohit,
  Schoelkopf, Stander, Songz, Tigges, Th{\'e}rien, Wang, and
  Weinbach]{gpt-neox-library}
A.~Andonian, Q.~Anthony, S.~Biderman, S.~Black, P.~Gali, L.~Gao, E.~Hallahan,
  J.~Levy-Kramer, C.~Leahy, L.~Nestler, K.~Parker, M.~Pieler, J.~Phang,
  S.~Purohit, H.~Schoelkopf, D.~Stander, T.~Songz, C.~Tigges, B.~Th{\'e}rien,
  P.~Wang, and S.~Weinbach.
\newblock {GPT-NeoX: Large Scale Autoregressive Language Modeling in PyTorch},
  9 2023.
\newblock URL \url{https://www.github.com/eleutherai/gpt-neox}.

\bibitem[Aryabumi et~al.(2024)Aryabumi, Dang, Talupuru, Dash, Cairuz, Lin,
  Venkitesh, Smith, Campos, Tan, et~al.]{aryabumi2024aya23}
V.~Aryabumi, J.~Dang, D.~Talupuru, S.~Dash, D.~Cairuz, H.~Lin, B.~Venkitesh,
  M.~Smith, J.~A. Campos, Y.~C. Tan, et~al.
\newblock Aya 23: Open weight releases to further multilingual progress.
\newblock \emph{arXiv preprint arXiv:2405.15032}, 2024.

\bibitem[Aulamo et~al.(2020)Aulamo, Virpioja, and
  Tiedemann]{aulamo-etal-2020-opusfilter}
M.~Aulamo, S.~Virpioja, and J.~Tiedemann.
\newblock {O}pus{F}ilter: A configurable parallel corpus filtering toolbox.
\newblock In \emph{Proceedings of the 58th Annual Meeting of the Association
  for Computational Linguistics: System Demonstrations}, pages 150--156.
  Association for Computational Linguistics, July 2020.
\newblock \doi{10.18653/v1/2020.acl-demos.20}.
\newblock URL \url{https://www.aclweb.org/anthology/2020.acl-demos.20}.

\bibitem[Bai et~al.(2023)Bai, Bai, Chu, Cui, Dang, Deng, Fan, Ge, Han, Huang,
  et~al.]{bai2023qwen}
J.~Bai, S.~Bai, Y.~Chu, Z.~Cui, K.~Dang, X.~Deng, Y.~Fan, W.~Ge, Y.~Han,
  F.~Huang, et~al.
\newblock Qwen technical report.
\newblock \emph{arXiv preprint arXiv:2309.16609}, 2023.

\bibitem[Bandarkar et~al.(2023)Bandarkar, Liang, Muller, Artetxe, Shukla, Husa,
  Goyal, Krishnan, Zettlemoyer, and Khabsa]{bandarkar2023belebele}
L.~Bandarkar, D.~Liang, B.~Muller, M.~Artetxe, S.~N. Shukla, D.~Husa, N.~Goyal,
  A.~Krishnan, L.~Zettlemoyer, and M.~Khabsa.
\newblock The {BELEBELE} benchmark: a parallel reading comprehension dataset in
  122 language variants.
\newblock \emph{arXiv preprint arXiv:2308.16884}, 2023.
\newblock URL \url{https://aclanthology.org/2024.acl-long.44/}.

\bibitem[Chen et~al.(2024)Chen, Yu, Guo, and Haddow]{chen2024good}
P.~Chen, S.~Yu, Z.~Guo, and B.~Haddow.
\newblock Is it good data for multilingual instruction tuning or just bad
  multilingual evaluation for large language models?
\newblock \emph{arXiv preprint arXiv:2406.12822}, 2024.
\newblock URL \url{https://arxiv.org/abs/2406.12822}.

\bibitem[Chiruzzo et~al.(2020)Chiruzzo, Amarilla, R{\'\i}os, and
  Gim{\'e}nez~Lugo]{americasnlp2023_source3}
L.~Chiruzzo, P.~Amarilla, A.~R{\'\i}os, and G.~Gim{\'e}nez~Lugo.
\newblock Development of a {G}uarani - {S}panish parallel corpus.
\newblock In N.~Calzolari, F.~B{\'e}chet, P.~Blache, K.~Choukri, C.~Cieri,
  T.~Declerck, S.~Goggi, H.~Isahara, B.~Maegaard, J.~Mariani, H.~Mazo,
  A.~Moreno, J.~Odijk, and S.~Piperidis, editors, \emph{Proceedings of the
  Twelfth Language Resources and Evaluation Conference}, pages 2629--2633,
  Marseille, France, May 2020. European Language Resources Association.
\newblock ISBN 979-10-95546-34-4.
\newblock URL \url{<https://aclanthology.org/2020.lrec-1.320>}.

\bibitem[Clark et~al.(2018)Clark, Cowhey, Etzioni, Khot, Sabharwal, Schoenick,
  and Tafjord]{clark2018think}
P.~Clark, I.~Cowhey, O.~Etzioni, T.~Khot, A.~Sabharwal, C.~Schoenick, and
  O.~Tafjord.
\newblock Think you have solved question answering? try arc, the ai2 reasoning
  challenge.
\newblock \emph{ArXiv}, abs/1803.05457, 2018.

\bibitem[Cobbe et~al.(2021)Cobbe, Kosaraju, Bavarian, Hilton, Nakano, Hesse,
  and Schulman]{cobbe2021training}
K.~Cobbe, V.~Kosaraju, M.~Bavarian, J.~Hilton, R.~Nakano, C.~Hesse, and
  J.~Schulman.
\newblock Training verifiers to solve math word problems, 2021.

\bibitem[Conneau et~al.(2018)Conneau, Rinott, Lample, Williams, Bowman,
  Schwenk, and Stoyanov]{conneau2018xnli}
A.~Conneau, R.~Rinott, G.~Lample, A.~Williams, S.~R. Bowman, H.~Schwenk, and
  V.~Stoyanov.
\newblock Xnli: Evaluating cross-lingual sentence representations.
\newblock In \emph{Proceedings of the 2018 Conference on Empirical Methods in
  Natural Language Processing}. Association for Computational Linguistics,
  2018.

\bibitem[Costa-juss{\`a} et~al.(2022)Costa-juss{\`a}, Cross, {\c{C}}elebi,
  Elbayad, Heafield, Heffernan, Kalbassi, Lam, Licht, Maillard,
  et~al.]{costa2022no}
M.~R. Costa-juss{\`a}, J.~Cross, O.~{\c{C}}elebi, M.~Elbayad, K.~Heafield,
  K.~Heffernan, E.~Kalbassi, J.~Lam, D.~Licht, J.~Maillard, et~al.
\newblock No language left behind: Scaling human-centered machine translation.
\newblock \emph{arXiv preprint arXiv:2207.04672}, 2022.

\bibitem[Dang et~al.(2024)Dang, Singh, D'souza, Ahmadian, Salamanca, Smith,
  Peppin, Hong, Govindassamy, Zhao, et~al.]{dang2024aya}
J.~Dang, S.~Singh, D.~D'souza, A.~Ahmadian, A.~Salamanca, M.~Smith, A.~Peppin,
  S.~Hong, M.~Govindassamy, T.~Zhao, et~al.
\newblock Aya expanse: Combining research breakthroughs for a new multilingual
  frontier.
\newblock \emph{arXiv preprint arXiv:2412.04261}, 2024.

\bibitem[de~Wynter et~al.(2024)de~Wynter, Watts, Wongsangaroonsri, Zhang,
  Farra, Altıntoprak, Baur, Claudet, Gajdusek, Gören, Gu, Kaminska, Kaminski,
  Kuo, Kyuba, Lee, Mathur, Merok, Milovanović, Paananen, Paananen, Pavlenko,
  Vidal, Strika, Tsao, Turcato, Vakhno, Velcsov, Vickers, Visser, Widarmanto,
  Zaikin, and Chen]{dewynter2024rtplxllmsevaluatetoxicity}
A.~de~Wynter, I.~Watts, T.~Wongsangaroonsri, M.~Zhang, N.~Farra, N.~E.
  Altıntoprak, L.~Baur, S.~Claudet, P.~Gajdusek, C.~Gören, Q.~Gu,
  A.~Kaminska, T.~Kaminski, R.~Kuo, A.~Kyuba, J.~Lee, K.~Mathur, P.~Merok,
  I.~Milovanović, N.~Paananen, V.-M. Paananen, A.~Pavlenko, B.~P. Vidal,
  L.~Strika, Y.~Tsao, D.~Turcato, O.~Vakhno, J.~Velcsov, A.~Vickers, S.~Visser,
  H.~Widarmanto, A.~Zaikin, and S.-Q. Chen.
\newblock {RTP-LX}: Can {LLMs} evaluate toxicity in multilingual scenarios?
\newblock \emph{arXiv preprint 2404.14397}, 2024.
\newblock URL \url{https://arxiv.org/abs/2404.14397}.

\bibitem[Devlin et~al.(2019)Devlin, Chang, Lee, and Toutanova]{devlin2019bert}
J.~Devlin, M.-W. Chang, K.~Lee, and K.~Toutanova.
\newblock {BERT}: Pre-training of deep bidirectional transformers for language
  understanding.
\newblock In \emph{Proceedings of the 2019 Conference of the North American
  Chapter of the Association for Computational Linguistics: Human Language
  Technologies}, 2019.

\bibitem[Dubey et~al.(2024)Dubey, Jauhri, Pandey, Kadian, Al-Dahle, Letman,
  Mathur, Schelten, Yang, Fan, et~al.]{dubey2024llama3}
A.~Dubey, A.~Jauhri, A.~Pandey, A.~Kadian, A.~Al-Dahle, A.~Letman, A.~Mathur,
  A.~Schelten, A.~Yang, A.~Fan, et~al.
\newblock The llama 3 herd of models.
\newblock \emph{arXiv preprint arXiv:2407.21783}, 2024.

\bibitem[{EdTeKLA Research Group}(2022)]{IndigenousLanguages_Corpora}
{EdTeKLA Research Group}.
\newblock Indigenous languages corpora.
\newblock \url{https://github.com/EdTeKLA/IndigenousLanguages_Corpora}, 2022.
\newblock Accessed: 2024-05-30.

\bibitem[Faysse(2023)]{project_gutenberg_HF_manu}
M.~Faysse.
\newblock Dataset card for "project gutenberg", 2023.
\newblock URL \url{https://huggingface.co/datasets/manu/project_gutenberg}.

\bibitem[Fujii et~al.(2024)Fujii, Nakamura, Loem, Iida, Ohi, Hattori, Shota,
  Mizuki, Yokota, and Okazaki]{fujii2024continualpretrainingcrosslingualllm}
K.~Fujii, T.~Nakamura, M.~Loem, H.~Iida, M.~Ohi, K.~Hattori, H.~Shota,
  S.~Mizuki, R.~Yokota, and N.~Okazaki.
\newblock Continual pre-training for cross-lingual llm adaptation: Enhancing
  japanese language capabilities.
\newblock \emph{arXiv preprint 2404.17790}, 2024.
\newblock URL \url{https://arxiv.org/abs/2404.17790}.

\bibitem[Fujii et~al.(2025)Fujii, Tajima, Mizuki, Shimada, Shiotani, Saito,
  Ohi, Kawamura, Nakamura, Okamoto, Ishida, Hattori, Ma, Takamura, Yokota, and
  Okazaki]{fujii2025rewritingpretrainingdataboosts}
K.~Fujii, Y.~Tajima, S.~Mizuki, H.~Shimada, T.~Shiotani, K.~Saito, M.~Ohi,
  M.~Kawamura, T.~Nakamura, T.~Okamoto, S.~Ishida, K.~Hattori, Y.~Ma,
  H.~Takamura, R.~Yokota, and N.~Okazaki.
\newblock Rewriting pre-training data boosts llm performance in math and code,
  2025.
\newblock URL \url{https://arxiv.org/abs/2505.02881}.

\bibitem[Galarreta et~al.(2017)Galarreta, Melgar, and
  Oncevay]{americasnlp2021_source3}
A.-P. Galarreta, A.~Melgar, and A.~Oncevay.
\newblock Corpus creation and initial {SMT} experiments between {S}panish and
  {S}hipibo-konibo.
\newblock In R.~Mitkov and G.~Angelova, editors, \emph{Proceedings of the
  International Conference Recent Advances in Natural Language Processing,
  {RANLP} 2017}, pages 238--244, Varna, Bulgaria, Sept. 2017. INCOMA Ltd.
\newblock \doi{10.26615/978-954-452-049-6_033}.
\newblock URL \url{https://doi.org/10.26615/978-954-452-049-6_033}.

\bibitem[Gao et~al.(2023)Gao, Tow, Biderman, Black, DiPofi, Foster, Golding,
  Hsu, McDonell, Muennighoff, et~al.]{eval-harness}
L.~Gao, J.~Tow, S.~Biderman, S.~Black, A.~DiPofi, C.~Foster, L.~Golding,
  J.~Hsu, K.~McDonell, N.~Muennighoff, et~al.
\newblock A framework for few-shot language model evaluation.
\newblock Zenodo, 2023.

\bibitem[Gbedevi(2019)]{gbedevi2024}
Y.~Gbedevi.
\newblock Ewe language corpus, 2019.
\newblock URL
  \url{https://www.kaggle.com/datasets/yvicherita/ewe-language-corpus}.
\newblock Accessed: 2024-08-27.

\bibitem[Gutierrez-Vasques et~al.(2016)Gutierrez-Vasques, Sierra, and
  Pompa]{americasnlp2023_source2}
X.~Gutierrez-Vasques, G.~Sierra, and I.~H. Pompa.
\newblock {A}xolotl: a web accessible parallel corpus for {S}panish-{N}ahuatl.
\newblock In N.~Calzolari, K.~Choukri, T.~Declerck, S.~Goggi, M.~Grobelnik,
  B.~Maegaard, J.~Mariani, H.~Mazo, A.~Moreno, J.~Odijk, and S.~Piperidis,
  editors, \emph{Proceedings of the Tenth International Conference on Language
  Resources and Evaluation ({LREC}'16)}, pages 4210--4214, Portoro{\v{z}},
  Slovenia, May 2016. European Language Resources Association (ELRA).
\newblock URL \url{https://aclanthology.org/L16-1666}.

\bibitem[Hasan et~al.(2021)Hasan, Bhattacharjee, Islam, Mubasshir, Li, Kang,
  Rahman, and Shahriyar]{hasan-etal-2021-xl}
T.~Hasan, A.~Bhattacharjee, M.~S. Islam, K.~Mubasshir, Y.-F. Li, Y.-B. Kang,
  M.~S. Rahman, and R.~Shahriyar.
\newblock {XL}-sum: Large-scale multilingual abstractive summarization for 44
  languages.
\newblock In C.~Zong, F.~Xia, W.~Li, and R.~Navigli, editors, \emph{Findings of
  the Association for Computational Linguistics: ACL-IJCNLP 2021}, pages
  4693--4703, Online, Aug. 2021. Association for Computational Linguistics.
\newblock \doi{10.18653/v1/2021.findings-acl.413}.
\newblock URL \url{https://aclanthology.org/2021.findings-acl.413}.

\bibitem[Hu et~al.(2022)Hu, Shen, Wallis, Allen-Zhu, Li, Wang, Wang, and
  Chen]{hu2022lora}
E.~J. Hu, Y.~Shen, P.~Wallis, Z.~Allen-Zhu, Y.~Li, S.~Wang, L.~Wang, and
  W.~Chen.
\newblock {LoRA}: Low-rank adaptation of large language models.
\newblock In \emph{International Conference on Learning Representations}, 2022.
\newblock URL \url{https://openreview.net/forum?id=nZeVKeeFYf9}.

\bibitem[Hu et~al.(2024)Hu, Tu, Han, He, Cui, Long, Zheng, Fang, Huang, Zhao,
  Zhang, Thai, Zhang, Wang, Yao, Zhao, Zhou, Cai, Zhai, Ding, Jia, Zeng, Li,
  Liu, and Sun]{DBLP:journals/corr/abs-2404-06395}
S.~Hu, Y.~Tu, X.~Han, C.~He, G.~Cui, X.~Long, Z.~Zheng, Y.~Fang, Y.~Huang,
  W.~Zhao, X.~Zhang, Z.~L. Thai, K.~Zhang, C.~Wang, Y.~Yao, C.~Zhao, J.~Zhou,
  J.~Cai, Z.~Zhai, N.~Ding, C.~Jia, G.~Zeng, D.~Li, Z.~Liu, and M.~Sun.
\newblock Minicpm: Unveiling the potential of small language models with
  scalable training strategies.
\newblock \emph{CoRR}, abs/2404.06395, 2024.
\newblock \doi{10.48550/ARXIV.2404.06395}.
\newblock URL \url{https://doi.org/10.48550/arXiv.2404.06395}.

\bibitem[Huang et~al.(2023)Huang, Tang, Zhang, Zhao, Song, Xia, and
  Wei]{huang-etal-2023-languages}
H.~Huang, T.~Tang, D.~Zhang, X.~Zhao, T.~Song, Y.~Xia, and F.~Wei.
\newblock Not all languages are created equal in {LLM}s: Improving multilingual
  capability by cross-lingual-thought prompting.
\newblock In H.~Bouamor, J.~Pino, and K.~Bali, editors, \emph{Findings of the
  Association for Computational Linguistics: EMNLP 2023}, pages 12365--12394,
  Singapore, Dec. 2023. Association for Computational Linguistics.
\newblock \doi{10.18653/v1/2023.findings-emnlp.826}.
\newblock URL \url{https://aclanthology.org/2023.findings-emnlp.826/}.

\bibitem[Huang et~al.(2025)Huang, Mo, Zhang, Li, Li, Zhang, Yi, Mao, Liu, Xu,
  Xu, Nie, and Liu]{huang2025surveylargelanguagemodels}
K.~Huang, F.~Mo, X.~Zhang, H.~Li, Y.~Li, Y.~Zhang, W.~Yi, Y.~Mao, J.~Liu,
  Y.~Xu, J.~Xu, J.-Y. Nie, and Y.~Liu.
\newblock A survey on large language models with multilingualism: Recent
  advances and new frontiers, 2025.
\newblock URL \url{https://arxiv.org/abs/2405.10936}.

\bibitem[Ji et~al.(2024{\natexlab{a}})Ji, Li, Paul, Paavola, Lin, Chen,
  O'Brien, Luo, Sch{\"u}tze, Tiedemann, et~al.]{ji2024emma}
S.~Ji, Z.~Li, I.~Paul, J.~Paavola, P.~Lin, P.~Chen, D.~O'Brien, H.~Luo,
  H.~Sch{\"u}tze, J.~Tiedemann, et~al.
\newblock Emma-500: Enhancing massively multilingual adaptation of large
  language models.
\newblock \emph{arXiv preprint arXiv:2409.17892}, 2024{\natexlab{a}}.

\bibitem[Ji et~al.(2024{\natexlab{b}})Ji, Mickus, Segonne, and
  Tiedemann]{ji2024can}
S.~Ji, T.~Mickus, V.~Segonne, and J.~Tiedemann.
\newblock Can machine translation bridge multilingual pretraining and
  cross-lingual transfer learning?
\newblock In \emph{Proceedings of LREC-COLING}, 2024{\natexlab{b}}.

\bibitem[Jiang et~al.(2023)Jiang, Sablayrolles, Mensch, Bamford, Chaplot,
  de~las Casas, Bressand, Lengyel, Lample, Saulnier, Lavaud, Lachaux, Stock,
  Scao, Lavril, Wang, Lacroix, and Sayed]{jiang2023mistral7b}
A.~Q. Jiang, A.~Sablayrolles, A.~Mensch, C.~Bamford, D.~S. Chaplot, D.~de~las
  Casas, F.~Bressand, G.~Lengyel, G.~Lample, L.~Saulnier, L.~R. Lavaud, M.-A.
  Lachaux, P.~Stock, T.~L. Scao, T.~Lavril, T.~Wang, T.~Lacroix, and W.~E.
  Sayed.
\newblock Mistral 7b, 2023.
\newblock URL \url{https://arxiv.org/abs/2310.06825}.

\bibitem[Joanis et~al.(2020)Joanis, Knowles, Kuhn, Larkin, Littell, Lo,
  Stewart, and Micher]{nunavut-hansard}
E.~Joanis, R.~Knowles, R.~Kuhn, S.~Larkin, P.~Littell, C.-k. Lo, D.~Stewart,
  and J.~Micher.
\newblock The {N}unavut {H}ansard {I}nuktitut{--}{E}nglish parallel corpus 3.0
  with preliminary machine translation results.
\newblock In N.~Calzolari, F.~B{\'e}chet, P.~Blache, K.~Choukri, C.~Cieri,
  T.~Declerck, S.~Goggi, H.~Isahara, B.~Maegaard, J.~Mariani, H.~Mazo,
  A.~Moreno, J.~Odijk, and S.~Piperidis, editors, \emph{Proceedings of the
  Twelfth Language Resources and Evaluation Conference}, pages 2562--2572,
  Marseille, France, May 2020. European Language Resources Association.
\newblock ISBN 979-10-95546-34-4.
\newblock URL \url{https://aclanthology.org/2020.lrec-1.312}.

\bibitem[Kargaran et~al.(2023)Kargaran, Imani, Yvon, and
  Sch{\"u}tze]{kargaran2023glotlid}
A.~H. Kargaran, A.~Imani, F.~Yvon, and H.~Sch{\"u}tze.
\newblock {GlotLID}: Language identification for low-resource languages.
\newblock In \emph{The 2023 Conference on Empirical Methods in Natural Language
  Processing}, 2023.
\newblock URL \url{https://openreview.net/forum?id=dl4e3EBz5j}.

\bibitem[Kingma and Ba(2015)]{kingma2015adam}
D.~P. Kingma and J.~Ba.
\newblock Adam: A method for stochastic optimization.
\newblock In \emph{International Conference for Learning Representations},
  2015.

\bibitem[Kocetkov et~al.(2023)Kocetkov, Li, Allal, Li, Mou, Jernite, Mitchell,
  Ferrandis, Hughes, Wolf, Bahdanau, von Werra, and
  de~Vries]{kocetkov2023stack}
D.~Kocetkov, R.~Li, L.~B. Allal, J.~Li, C.~Mou, Y.~Jernite, M.~Mitchell, C.~M.
  Ferrandis, S.~Hughes, T.~Wolf, D.~Bahdanau, L.~von Werra, and H.~de~Vries.
\newblock The stack: 3 {TB} of permissively licensed source code.
\newblock \emph{Trans. Mach. Learn. Res.}, 2023, 2023.
\newblock URL \url{https://openreview.net/forum?id=pxpbTdUEpD}.

\bibitem[Kondo et~al.(2024)Kondo, Utsuro, and
  Nagata]{kondo2024enhancingtranslationaccuracylarge}
M.~Kondo, T.~Utsuro, and M.~Nagata.
\newblock Enhancing translation accuracy of large language models through
  continual pre-training on parallel data.
\newblock \emph{arXiv preprint 2407.03145}, 2024.
\newblock URL \url{https://arxiv.org/abs/2407.03145}.

\bibitem[Kwon et~al.(2023)Kwon, Li, Zhuang, Sheng, Zheng, Yu, Gonzalez, Zhang,
  and Stoica]{kwon2023efficient}
W.~Kwon, Z.~Li, S.~Zhuang, Y.~Sheng, L.~Zheng, C.~H. Yu, J.~E. Gonzalez,
  H.~Zhang, and I.~Stoica.
\newblock Efficient memory management for large language model serving with
  pagedattention.
\newblock In \emph{Proceedings of the ACM SIGOPS 29th Symposium on Operating
  Systems Principles}, 2023.

\bibitem[Lai et~al.(2023)Lai, Nguyen, Ngo, Nguyen, Dernoncourt, Rossi, and
  Nguyen]{lai2023okapi}
V.~D. Lai, C.~V. Nguyen, N.~T. Ngo, T.~Nguyen, F.~Dernoncourt, R.~A. Rossi, and
  T.~H. Nguyen.
\newblock Okapi: Instruction-tuned large language models in multiple languages
  with reinforcement learning from human feedback.
\newblock In \emph{Proceedings of the 2023 Conference on Empirical Methods in
  Natural Language Processing: System Demonstrations}, pages 318--327, 2023.

\bibitem[Lai et~al.(2024)Lai, Mesgar, and Fraser]{lai-etal-2024-llms}
W.~Lai, M.~Mesgar, and A.~Fraser.
\newblock {LLM}s beyond {E}nglish: Scaling the multilingual capability of
  {LLM}s with cross-lingual feedback.
\newblock In L.-W. Ku, A.~Martins, and V.~Srikumar, editors, \emph{Findings of
  the Association for Computational Linguistics ACL 2024}, pages 8186--8213,
  Bangkok, Thailand and virtual meeting, Aug. 2024. Association for
  Computational Linguistics.
\newblock URL \url{https://aclanthology.org/2024.findings-acl.488}.

\bibitem[Lewis(2019)]{lewis2019bart}
M.~Lewis.
\newblock {BART}: Denoising sequence-to-sequence pre-training for natural
  language generation, translation, and comprehension.
\newblock \emph{arXiv preprint arXiv:1910.13461}, 2019.

\bibitem[Li et~al.(2024{\natexlab{a}})Li, Dong, Chen, Su, Zhou, Ai, Ye, and
  Liu]{li2024llmsasjudgescomprehensivesurveyllmbased}
H.~Li, Q.~Dong, J.~Chen, H.~Su, Y.~Zhou, Q.~Ai, Z.~Ye, and Y.~Liu.
\newblock Llms-as-judges: A comprehensive survey on llm-based evaluation
  methods, 2024{\natexlab{a}}.
\newblock URL \url{https://arxiv.org/abs/2412.05579}.

\bibitem[Li et~al.(2022)Li, Zhang, Zhao, Shen, Liu, Mao, and Zhang]{li2022csl}
Y.~Li, Y.~Zhang, Z.~Zhao, L.~Shen, W.~Liu, W.~Mao, and H.~Zhang.
\newblock {CSL}: A large-scale chinese scientific literature dataset.
\newblock In \emph{Proceedings of the 29th International Conference on
  Computational Linguistics}, pages 3917--3923, 2022.

\bibitem[Li et~al.(2024{\natexlab{b}})Li, Ji, Mickus, Segonne, and
  Tiedemann]{li2024comparison}
Z.~Li, S.~Ji, T.~Mickus, V.~Segonne, and J.~Tiedemann.
\newblock A comparison of language modeling and translation as multilingual
  pretraining objectives.
\newblock In \emph{Proceedings of EMNLP}, 2024{\natexlab{b}}.

\bibitem[Li et~al.(2025)Li, Ji, Luo, and
  Tiedemann]{li2025rethinkingmultilingualcontinualpretraining}
Z.~Li, S.~Ji, H.~Luo, and J.~Tiedemann.
\newblock Rethinking multilingual continual pretraining: Data mixing for
  adapting llms across languages and resources.
\newblock \emph{arXiv preprint 2504.04152}, 2025.
\newblock URL \url{https://arxiv.org/abs/2504.04152}.

\bibitem[Lin(2004)]{lin-2004-rouge}
C.-Y. Lin.
\newblock {ROUGE}: A package for automatic evaluation of summaries.
\newblock In \emph{Text Summarization Branches Out}, pages 74--81, Barcelona,
  Spain, July 2004. Association for Computational Linguistics.
\newblock URL \url{https://aclanthology.org/W04-1013}.

\bibitem[Lin et~al.(2024)Lin, Ji, Tiedemann, Martins, and
  Sch{\"u}tze]{lin2024mala}
P.~Lin, S.~Ji, J.~Tiedemann, A.~F. Martins, and H.~Sch{\"u}tze.
\newblock {MaLA}-500: Massive language adaptation of large language models.
\newblock \emph{arXiv preprint arXiv:2401.13303}, 2024.

\bibitem[Lin et~al.(2022)Lin, Mihaylov, Artetxe, Wang, Chen, Simig, Ott, Goyal,
  Bhosale, Du, et~al.]{lin2022few}
X.~V. Lin, T.~Mihaylov, M.~Artetxe, T.~Wang, S.~Chen, D.~Simig, M.~Ott,
  N.~Goyal, S.~Bhosale, J.~Du, et~al.
\newblock Few-shot learning with multilingual generative language models.
\newblock In \emph{Proceedings of the 2022 Conference on Empirical Methods in
  Natural Language Processing}, pages 9019--9052, 2022.

\bibitem[Lozhkov et~al.(2024)Lozhkov, Li, Allal, Cassano, Lamy-Poirier, Tazi,
  Tang, Pykhtar, Liu, Wei, et~al.]{lozhkov2024starcoder}
A.~Lozhkov, R.~Li, L.~B. Allal, F.~Cassano, J.~Lamy-Poirier, N.~Tazi, A.~Tang,
  D.~Pykhtar, J.~Liu, Y.~Wei, et~al.
\newblock Starcoder 2 and the stack v2: The next generation.
\newblock \emph{arXiv preprint arXiv:2402.19173}, 2024.

\bibitem[Lu et~al.(2024)Lu, Zhu, Li, Qiao, and Yuan]{lu2024llamax}
Y.~Lu, W.~Zhu, L.~Li, Y.~Qiao, and F.~Yuan.
\newblock {LLaMAX}: Scaling linguistic horizons of {LLM} by enhancing
  translation capabilities beyond 100 languages.
\newblock \emph{arXiv preprint arXiv:2407.05975}, 2024.

\bibitem[Luukkonen et~al.(2024)Luukkonen, Burdge, Zosa, Talman, Komulainen,
  Hatanpää, Sarlin, and Pyysalo]{luukkonen2024poro34bblessingmultilinguality}
R.~Luukkonen, J.~Burdge, E.~Zosa, A.~Talman, V.~Komulainen, V.~Hatanpää,
  P.~Sarlin, and S.~Pyysalo.
\newblock Poro 34b and the blessing of multilinguality.
\newblock \emph{arXiv preprint 2404.01856}, 2024.
\newblock URL \url{https://arxiv.org/abs/2404.01856}.

\bibitem[Ma et~al.(2023)Ma, ImaniGooghari, Ye, Asgari, and
  Sch{\"u}tze]{ma2023taxi1500}
C.~Ma, A.~ImaniGooghari, H.~Ye, E.~Asgari, and H.~Sch{\"u}tze.
\newblock Taxi1500: A multilingual dataset for text classification in 1500
  languages, 2023.

\bibitem[Mager et~al.(2018)Mager, Carrillo, and Meza]{americasnlp2023_source1}
M.~Mager, D.~Carrillo, and I.~Meza.
\newblock Probabilistic finite-state morphological segmenter for wixarika
  (huichol) language.
\newblock \emph{Journal of Intelligent \& Fuzzy Systems}, 34\penalty0
  (5):\penalty0 3081--3087, 2018.

\bibitem[Maini et~al.(2024)Maini, Seto, Bai, Grangier, Zhang, and
  Jaitly]{DBLP:conf/acl/MainiSBG0J24}
P.~Maini, S.~Seto, R.~H. Bai, D.~Grangier, Y.~Zhang, and N.~Jaitly.
\newblock Rephrasing the web: {A} recipe for compute and data-efficient
  language modeling.
\newblock In L.~Ku, A.~Martins, and V.~Srikumar, editors, \emph{Proceedings of
  the 62nd Annual Meeting of the Association for Computational Linguistics
  (Volume 1: Long Papers), {ACL} 2024, Bangkok, Thailand, August 11-16, 2024},
  pages 14044--14072. Association for Computational Linguistics, 2024.
\newblock \doi{10.18653/V1/2024.ACL-LONG.757}.
\newblock URL \url{https://doi.org/10.18653/v1/2024.acl-long.757}.

\bibitem[Masakhane(2022)]{lacuna_pos_ner}
Masakhane.
\newblock lacuna\_pos\_ner: {POS} and {NER} for african languages.
\newblock \url{https://github.com/masakhane-io/lacuna_pos_ner}, 2022.
\newblock Accessed: 2024-05-30.

\bibitem[Mayer and Cysouw(2014)]{pbc}
T.~Mayer and M.~Cysouw.
\newblock Creating a massively parallel bible corpus.
\newblock In N.~Calzolari, K.~Choukri, T.~Declerck, H.~Loftsson, B.~Maegaard,
  J.~Mariani, A.~Moreno, J.~Odijk, and S.~Piperidis, editors, \emph{Proceedings
  of the Ninth International Conference on Language Resources and Evaluation,
  {LREC} 2014, Reykjavik, Iceland, May 26-31, 2014}, pages 3158--3163. European
  Language Resources Association {(ELRA)}, 2014.
\newblock URL
  \url{http://www.lrec-conf.org/proceedings/lrec2014/summaries/220.html}.

\bibitem[Ming et~al.(2024)Ming, Zeng, Lyu, Shi, Zhao, Yang, Liu, Wang, Xu, Liu,
  Zhao, Wang, Liu, Zhou, Yin, Shang, Li, Wang, Luo, and
  Zhang]{ming2024marcollmbridginglanguagesmassive}
L.~Ming, B.~Zeng, C.~Lyu, T.~Shi, Y.~Zhao, X.~Yang, Y.~Liu, Y.~Wang, L.~Xu,
  Y.~Liu, X.~Zhao, H.~Wang, H.~Liu, H.~Zhou, H.~Yin, Z.~Shang, H.~Li, L.~Wang,
  W.~Luo, and K.~Zhang.
\newblock Marco-llm: Bridging languages via massive multilingual training for
  cross-lingual enhancement, 2024.
\newblock URL \url{https://arxiv.org/abs/2412.04003}.

\bibitem[Muennighoff et~al.(2022)Muennighoff, Wang, Sutawika, Roberts,
  Biderman, Scao, Bari, Shen, Yong, Schoelkopf,
  et~al.]{muennighoff2022crosslingual}
N.~Muennighoff, T.~Wang, L.~Sutawika, A.~Roberts, S.~Biderman, T.~L. Scao,
  M.~S. Bari, S.~Shen, Z.-X. Yong, H.~Schoelkopf, et~al.
\newblock Crosslingual generalization through multitask finetuning.
\newblock \emph{arXiv preprint arXiv:2211.01786}, 2022.

\bibitem[Mukiibi et~al.(2021)Mukiibi, Claire, and
  Joyce]{english-luganda_parallel_corpus}
J.~Mukiibi, B.~Claire, and N.-N. Joyce.
\newblock An english-luganda parallel corpus, May 2021.
\newblock URL \url{https://doi.org/10.5281/zenodo.4764039}.

\bibitem[{NLLB Team} et~al.(2022){NLLB Team}, R, Cross, {\c{C}}elebi, Elbayad,
  Heafield, Heffernan, Kalbassi, Lam, Licht, Maillard, et~al.]{nllb2022}
{NLLB Team}, M.~R, J.~Cross, O.~{\c{C}}elebi, M.~Elbayad, K.~Heafield,
  K.~Heffernan, E.~Kalbassi, J.~Lam, D.~Licht, J.~Maillard, et~al.
\newblock No language left behind: Scaling human-centered machine translation.
\newblock \emph{arXiv preprint arXiv:2207.04672}, 2022.

\bibitem[Ortega et~al.(2020)Ortega, Castro-Mamani, and
  Montoya~Samame]{americasnlp2023_source9}
J.~E. Ortega, R.~A. Castro-Mamani, and J.~R. Montoya~Samame.
\newblock Overcoming resistance: The normalization of an {A}mazonian tribal
  language.
\newblock In A.~Karakanta, A.~K. Ojha, C.-H. Liu, J.~Abbott, J.~Ortega,
  J.~Washington, N.~Oco, S.~M. Lakew, T.~A. Pirinen, V.~Malykh, V.~Logacheva,
  and X.~Zhao, editors, \emph{Proceedings of the 3rd Workshop on Technologies
  for MT of Low Resource Languages}, pages 1--13, Suzhou, China, Dec. 2020.
  Association for Computational Linguistics.
\newblock URL \url{https://aclanthology.org/2020.loresmt-1.1}.

\bibitem[Papineni et~al.(2002)Papineni, Roukos, Ward, and
  Zhu]{papineni-etal-2002-bleu}
K.~Papineni, S.~Roukos, T.~Ward, and W.-J. Zhu.
\newblock {B}leu: a method for automatic evaluation of machine translation.
\newblock In P.~Isabelle, E.~Charniak, and D.~Lin, editors, \emph{Proceedings
  of the 40th Annual Meeting of the Association for Computational Linguistics},
  pages 311--318, Philadelphia, Pennsylvania, USA, July 2002. Association for
  Computational Linguistics.
\newblock \doi{10.3115/1073083.1073135}.
\newblock URL \url{https://aclanthology.org/P02-1040}.

\bibitem[Paul et~al.(2024)Paul, Glavas, and Gurevych]{paul2024ircoder}
I.~Paul, G.~Glavas, and I.~Gurevych.
\newblock Ircoder: Intermediate representations make language models robust
  multilingual code generators.
\newblock In L.~Ku, A.~Martins, and V.~Srikumar, editors, \emph{Proceedings of
  the 62nd Annual Meeting of the Association for Computational Linguistics
  (Volume 1: Long Papers), {ACL} 2024, Bangkok, Thailand, August 11-16, 2024},
  pages 15023--15041. Association for Computational Linguistics, 2024.
\newblock URL \url{https://aclanthology.org/2024.acl-long.802}.

\bibitem[Paul et~al.(2025)Paul, Yang, Glavas, Kersting, and
  Gurevych]{DBLP:conf/iclr/PaulYGKG25}
I.~Paul, H.~Yang, G.~Glavas, K.~Kersting, and I.~Gurevych.
\newblock Obscuracoder: Powering efficient code {LM} pre-training via
  obfuscation grounding.
\newblock In \emph{The Thirteenth International Conference on Learning
  Representations, {ICLR} 2025, Singapore, April 24-28, 2025}. OpenReview.net,
  2025.
\newblock URL \url{https://openreview.net/forum?id=VYvxrD7aS0}.

\bibitem[Petty et~al.(2024)Petty, van Steenkiste, and
  Linzen]{DBLP:journals/corr/abs-2409-04556}
J.~Petty, S.~van Steenkiste, and T.~Linzen.
\newblock How does code pretraining affect language model task performance?
\newblock \emph{CoRR}, abs/2409.04556, 2024.
\newblock \doi{10.48550/ARXIV.2409.04556}.
\newblock URL \url{https://doi.org/10.48550/arXiv.2409.04556}.

\bibitem[Ponti et~al.(2020)Ponti, Glava{\v{s}}, Majewska, Liu, Vuli{\'c}, and
  Korhonen]{ponti-etal-2020-xcopa}
E.~M. Ponti, G.~Glava{\v{s}}, O.~Majewska, Q.~Liu, I.~Vuli{\'c}, and
  A.~Korhonen.
\newblock {XCOPA}: A multilingual dataset for causal commonsense reasoning.
\newblock In \emph{Proceedings of the 2020 Conference on Empirical Methods in
  Natural Language Processing (EMNLP)}, 2020.

\bibitem[Popovi{\'c}(2015)]{popovic-2015-chrf}
M.~Popovi{\'c}.
\newblock chr{F}: character n-gram {F}-score for automatic {MT} evaluation.
\newblock In O.~Bojar, R.~Chatterjee, C.~Federmann, B.~Haddow, C.~Hokamp,
  M.~Huck, V.~Logacheva, and P.~Pecina, editors, \emph{Proceedings of the Tenth
  Workshop on Statistical Machine Translation}, pages 392--395, Lisbon,
  Portugal, Sept. 2015. Association for Computational Linguistics.
\newblock \doi{10.18653/v1/W15-3049}.
\newblock URL \url{https://aclanthology.org/W15-3049}.

\bibitem[Post(2018)]{post-2018-call}
M.~Post.
\newblock A call for clarity in reporting {BLEU} scores.
\newblock In \emph{Proceedings of the Third Conference on Machine Translation:
  Research Papers}, pages 186--191, Belgium, Brussels, Oct. 2018. Association
  for Computational Linguistics.
\newblock URL \url{https://www.aclweb.org/anthology/W18-6319}.

\bibitem[Puri et~al.(2021)Puri, Kung, Janssen, Zhang, Domeniconi, Zolotov,
  Dolby, Chen, Choudhury, Decker, Thost, Buratti, Pujar, Ramji, Finkler,
  Malaika, and Reiss]{DBLP:conf/nips/Puri0JZDZD0CDTB21}
R.~Puri, D.~S. Kung, G.~Janssen, W.~Zhang, G.~Domeniconi, V.~Zolotov, J.~Dolby,
  J.~Chen, M.~R. Choudhury, L.~Decker, V.~Thost, L.~Buratti, S.~Pujar,
  S.~Ramji, U.~Finkler, S.~Malaika, and F.~Reiss.
\newblock Codenet: {A} large-scale {AI} for code dataset for learning a
  diversity of coding tasks.
\newblock In J.~Vanschoren and S.~Yeung, editors, \emph{Proceedings of the
  Neural Information Processing Systems Track on Datasets and Benchmarks 1,
  NeurIPS Datasets and Benchmarks 2021, December 2021, virtual}, 2021.
\newblock URL
  \url{https://datasets-benchmarks-proceedings.neurips.cc/paper/2021/hash/a5bfc9e07964f8dddeb95fc584cd965d-Abstract-round2.html}.

\bibitem[Qwen et~al.(2025)Qwen, :, Yang, Yang, Zhang, Hui, Zheng, Yu, Li, Liu,
  Huang, Wei, Lin, Yang, Tu, Zhang, Yang, Yang, Zhou, Lin, Dang, Lu, Bao, Yang,
  Yu, Li, Xue, Zhang, Zhu, Men, Lin, Li, Tang, Xia, Ren, Ren, Fan, Su, Zhang,
  Wan, Liu, Cui, Zhang, and Qiu]{qwen2025qwen25technicalreport}
Qwen, :, A.~Yang, B.~Yang, B.~Zhang, B.~Hui, B.~Zheng, B.~Yu, C.~Li, D.~Liu,
  F.~Huang, H.~Wei, H.~Lin, J.~Yang, J.~Tu, J.~Zhang, J.~Yang, J.~Yang,
  J.~Zhou, J.~Lin, K.~Dang, K.~Lu, K.~Bao, K.~Yang, L.~Yu, M.~Li, M.~Xue,
  P.~Zhang, Q.~Zhu, R.~Men, R.~Lin, T.~Li, T.~Tang, T.~Xia, X.~Ren, X.~Ren,
  Y.~Fan, Y.~Su, Y.~Zhang, Y.~Wan, Y.~Liu, Z.~Cui, Z.~Zhang, and Z.~Qiu.
\newblock Qwen2.5 technical report, 2025.
\newblock URL \url{https://arxiv.org/abs/2412.15115}.

\bibitem[Radford et~al.(2019)Radford, Wu, Child, Luan, Amodei, Sutskever,
  et~al.]{radford2019language}
A.~Radford, J.~Wu, R.~Child, D.~Luan, D.~Amodei, I.~Sutskever, et~al.
\newblock Language models are unsupervised multitask learners.
\newblock \emph{OpenAI blog}, 1\penalty0 (8):\penalty0 9, 2019.

\bibitem[Roziere et~al.(2023)Roziere, Gehring, Gloeckle, Sootla, Gat, Tan, Adi,
  Liu, Remez, Rapin, et~al.]{roziere2023code}
B.~Roziere, J.~Gehring, F.~Gloeckle, S.~Sootla, I.~Gat, X.~E. Tan, Y.~Adi,
  J.~Liu, T.~Remez, J.~Rapin, et~al.
\newblock Code llama: Open foundation models for code.
\newblock \emph{arXiv preprint arXiv:2308.12950}, 2023.

\bibitem[Ruis et~al.(2025)Ruis, Mozes, Bae, Kamalakara, Gnaneshwar, Locatelli,
  Kirk, Rockt{\"{a}}schel, Grefenstette, and
  Bartolo]{DBLP:conf/iclr/RuisMBKGLKRGB25}
L.~Ruis, M.~Mozes, J.~Bae, S.~R. Kamalakara, D.~Gnaneshwar, A.~Locatelli,
  R.~Kirk, T.~Rockt{\"{a}}schel, E.~Grefenstette, and M.~Bartolo.
\newblock Procedural knowledge in pretraining drives reasoning in large
  language models.
\newblock In \emph{The Thirteenth International Conference on Learning
  Representations, {ICLR} 2025, Singapore, April 24-28, 2025}. OpenReview.net,
  2025.
\newblock URL \url{https://openreview.net/forum?id=1hQKHHUsMx}.

\bibitem[Scao et~al.(2022)Scao, Fan, Akiki, Pavlick, Ili{\'c}, Hesslow,
  Castagn{\'e}, Luccioni, Yvon, Gall{\'e}, et~al.]{scao2022bloom}
T.~L. Scao, A.~Fan, C.~Akiki, E.~Pavlick, S.~Ili{\'c}, D.~Hesslow,
  R.~Castagn{\'e}, A.~S. Luccioni, F.~Yvon, M.~Gall{\'e}, et~al.
\newblock {BLOOM}: A {176B}-parameter open-access multilingual language model.
\newblock \emph{arXiv preprint}, 2022.

\bibitem[Shi et~al.(2022)Shi, Suzgun, Freitag, Wang, Srivats, Vosoughi, Chung,
  Tay, Ruder, Zhou, Das, and Wei]{shi2022language}
F.~Shi, M.~Suzgun, M.~Freitag, X.~Wang, S.~Srivats, S.~Vosoughi, H.~W. Chung,
  Y.~Tay, S.~Ruder, D.~Zhou, D.~Das, and J.~Wei.
\newblock Language models are multilingual chain-of-thought reasoners, 2022.

\bibitem[Sindhujan et~al.(2025)Sindhujan, Kanojia, Orasan, and
  Qian]{sindhujan2025llmsstrugglereferencelesstranslation}
A.~Sindhujan, D.~Kanojia, C.~Orasan, and S.~Qian.
\newblock When llms struggle: Reference-less translation evaluation for
  low-resource languages, 2025.
\newblock URL \url{https://arxiv.org/abs/2501.04473}.

\bibitem[Soldaini and Lo(2023)]{peS2o}
L.~Soldaini and K.~Lo.
\newblock {peS2o (Pretraining Efficiently on S2ORC) Dataset}.
\newblock Technical report, {Allen Institute for AI}, 2023.
\newblock ODC-By, \url{https://github.com/allenai/pes2o}.

\bibitem[Soldaini et~al.(2024)Soldaini, Kinney, Bhagia, Schwenk, Atkinson,
  Authur, Bogin, Chandu, Dumas, Elazar, Hofmann, Jha, Kumar, Lucy, Lyu,
  Lambert, Magnusson, Morrison, Muennighoff, Naik, Nam, Peters, Ravichander,
  Richardson, Shen, Strubell, Subramani, Tafjord, Walsh, Zettlemoyer, Smith,
  Hajishirzi, Beltagy, Groeneveld, Dodge, and
  Lo]{DBLP:conf/acl/SoldainiKBSAABC24}
L.~Soldaini, R.~Kinney, A.~Bhagia, D.~Schwenk, D.~Atkinson, R.~Authur,
  B.~Bogin, K.~R. Chandu, J.~Dumas, Y.~Elazar, V.~Hofmann, A.~H. Jha, S.~Kumar,
  L.~Lucy, X.~Lyu, N.~Lambert, I.~Magnusson, J.~Morrison, N.~Muennighoff,
  A.~Naik, C.~Nam, M.~E. Peters, A.~Ravichander, K.~Richardson, Z.~Shen,
  E.~Strubell, N.~Subramani, O.~Tafjord, P.~Walsh, L.~Zettlemoyer, N.~A. Smith,
  H.~Hajishirzi, I.~Beltagy, D.~Groeneveld, J.~Dodge, and K.~Lo.
\newblock Dolma: an open corpus of three trillion tokens for language model
  pretraining research.
\newblock In L.~Ku, A.~Martins, and V.~Srikumar, editors, \emph{Proceedings of
  the 62nd Annual Meeting of the Association for Computational Linguistics
  (Volume 1: Long Papers), {ACL} 2024, Bangkok, Thailand, August 11-16, 2024},
  pages 15725--15788. Association for Computational Linguistics, 2024.
\newblock \doi{10.18653/V1/2024.ACL-LONG.840}.
\newblock URL \url{https://doi.org/10.18653/v1/2024.acl-long.840}.

\bibitem[Springer et~al.(2025)Springer, Goyal, Wen, Kumar, Yue, Malladi,
  Neubig, and Raghunathan]{springer2025overtrained}
J.~M. Springer, S.~Goyal, K.~Wen, T.~Kumar, X.~Yue, S.~Malladi, G.~Neubig, and
  A.~Raghunathan.
\newblock Overtrained language models are harder to fine-tune.
\newblock \emph{arXiv preprint arXiv:2503.19206}, 2025.

\bibitem[Su et~al.(2024)Su, Kong, Lin, Jennings, Norick, Kliegl, Patwary,
  Shoeybi, and Catanzaro]{DBLP:journals/corr/abs-2412-02595}
D.~Su, K.~Kong, Y.~Lin, J.~Jennings, B.~Norick, M.~Kliegl, M.~Patwary,
  M.~Shoeybi, and B.~Catanzaro.
\newblock Nemotron-cc: Transforming common crawl into a refined long-horizon
  pretraining dataset.
\newblock \emph{CoRR}, abs/2412.02595, 2024.
\newblock \doi{10.48550/ARXIV.2412.02595}.
\newblock URL \url{https://doi.org/10.48550/arXiv.2412.02595}.

\bibitem[Szafraniec et~al.(2023)Szafraniec, Rozi{\`{e}}re, Leather, Labatut,
  Charton, and Synnaeve]{szafraniec2022translation}
M.~Szafraniec, B.~Rozi{\`{e}}re, H.~Leather, P.~Labatut, F.~Charton, and
  G.~Synnaeve.
\newblock Code translation with compiler representations.
\newblock In \emph{The Eleventh International Conference on Learning
  Representations, {ICLR} 2023, Kigali, Rwanda, May 1-5, 2023}. OpenReview.net,
  2023.
\newblock URL \url{https://openreview.net/forum?id=XomEU3eNeSQ}.

\bibitem[Tang et~al.(2020)Tang, Tran, Li, Chen, Goyal, Chaudhary, Gu, and
  Fan]{tang2020multilingual}
Y.~Tang, C.~Tran, X.~Li, P.-J. Chen, N.~Goyal, V.~Chaudhary, J.~Gu, and A.~Fan.
\newblock Multilingual translation with extensible multilingual pretraining and
  finetuning.
\newblock \emph{arXiv preprint arXiv:2008.00401}, 2020.

\bibitem[Taori et~al.(2023)Taori, Gulrajani, Zhang, Dubois, Li, Guestrin,
  Liang, and Hashimoto]{alpaca}
R.~Taori, I.~Gulrajani, T.~Zhang, Y.~Dubois, X.~Li, C.~Guestrin, P.~Liang, and
  T.~B. Hashimoto.
\newblock Stanford {Alpaca}: An instruction-following {LLaMA} model.
\newblock \url{https://github.com/tatsu-lab/stanford_alpaca}, 2023.

\bibitem[Team et~al.(2024)Team, Riviere, Pathak, Sessa, Hardin, Bhupatiraju,
  Hussenot, Mesnard, Shahriari, Ram{\'e},
  et~al.]{gemmateam2024gemma2improvingopen}
G.~Team, M.~Riviere, S.~Pathak, P.~G. Sessa, C.~Hardin, S.~Bhupatiraju,
  L.~Hussenot, T.~Mesnard, B.~Shahriari, A.~Ram{\'e}, et~al.
\newblock Gemma 2: Improving open language models at a practical size.
\newblock \emph{arXiv preprint arXiv:2408.00118}, 2024.

\bibitem[Tiedemann(2012)]{opus_2012}
J.~Tiedemann.
\newblock Parallel data, tools and interfaces in {OPUS}.
\newblock In N.~Calzolari, K.~Choukri, T.~Declerck, M.~U. Do{\u{g}}an,
  B.~Maegaard, J.~Mariani, A.~Moreno, J.~Odijk, and S.~Piperidis, editors,
  \emph{Proceedings of the Eighth International Conference on Language
  Resources and Evaluation ({LREC}'12)}, pages 2214--2218, Istanbul, Turkey,
  May 2012. European Language Resources Association (ELRA).
\newblock URL
  \url{http://www.lrec-conf.org/proceedings/lrec2012/pdf/463_Paper.pdf}.

\bibitem[Tiedemann(2020)]{tatoeba}
J.~Tiedemann.
\newblock The {T}atoeba {T}ranslation {C}hallenge {--} {R}ealistic data sets
  for low resource and multilingual {MT}.
\newblock In \emph{Proceedings of the Fifth Conference on Machine Translation},
  pages 1174--1182, Online, Nov. 2020. Association for Computational
  Linguistics.
\newblock URL \url{https://www.aclweb.org/anthology/2020.wmt-1.139}.

\bibitem[Touvron et~al.(2023{\natexlab{a}})Touvron, Lavril, Izacard, Martinet,
  Lachaux, Lacroix, Rozi{\`{e}}re, Goyal, Hambro, Azhar, Rodriguez, Joulin,
  Grave, and Lample]{touvron2023llama}
H.~Touvron, T.~Lavril, G.~Izacard, X.~Martinet, M.~Lachaux, T.~Lacroix,
  B.~Rozi{\`{e}}re, N.~Goyal, E.~Hambro, F.~Azhar, A.~Rodriguez, A.~Joulin,
  E.~Grave, and G.~Lample.
\newblock {LLaMA}: Open and efficient foundation language models.
\newblock \emph{CoRR}, abs/2302.13971, 2023{\natexlab{a}}.
\newblock \doi{10.48550/arXiv.2302.13971}.
\newblock URL \url{https://doi.org/10.48550/arXiv.2302.13971}.

\bibitem[Touvron et~al.(2023{\natexlab{b}})Touvron, Martin, Stone, Albert,
  Almahairi, Babaei, Bashlykov, Batra, Bhargava, Bhosale,
  et~al.]{touvron2023llama2}
H.~Touvron, L.~Martin, K.~Stone, P.~Albert, A.~Almahairi, Y.~Babaei,
  N.~Bashlykov, S.~Batra, P.~Bhargava, S.~Bhosale, et~al.
\newblock Llama 2: Open foundation and fine-tuned chat models.
\newblock \emph{arXiv preprint}, 2023{\natexlab{b}}.

\bibitem[University(2010)]{haitian_creole}
C.~M. University.
\newblock Haitian creole data, 2010.
\newblock URL \url{http://www.speech.cs.cmu.edu/haitian/}.
\newblock Accessed: 2024-05-30.

\bibitem[Varab and Schluter(2021)]{varab-schluter-2021-massivesumm}
D.~Varab and N.~Schluter.
\newblock {M}assive{S}umm: a very large-scale, very multilingual, news
  summarisation dataset.
\newblock In M.-F. Moens, X.~Huang, L.~Specia, and S.~W.-t. Yih, editors,
  \emph{Proceedings of the 2021 Conference on Empirical Methods in Natural
  Language Processing}, pages 10150--10161, Online and Punta Cana, Dominican
  Republic, Nov. 2021. Association for Computational Linguistics.
\newblock \doi{10.18653/v1/2021.emnlp-main.797}.
\newblock URL \url{https://aclanthology.org/2021.emnlp-main.797/}.

\bibitem[Vel{\'a}zquez(2021)]{americasnlp2023_source11}
J.~M. L.~C. Vel{\'a}zquez.
\newblock Spanish-hn{\"a}hn{\"u} corpus, 2021.
\newblock URL \url{https://tsunkua.elotl.mx/about/}.

\bibitem[Wang et~al.(2023)Wang, Tu, Chen, Yuan, Huang, Jiao, and
  Lyu]{wang2023all}
W.~Wang, Z.~Tu, C.~Chen, Y.~Yuan, J.-t. Huang, W.~Jiao, and M.~R. Lyu.
\newblock All languages matter: On the multilingual safety of large language
  models.
\newblock \emph{arXiv preprint arXiv:2310.00905}, 2023.

\bibitem[Wei et~al.(2022)Wei, Wang, Schuurmans, Bosma, Xia, Chi, Le, Zhou,
  et~al.]{wei2022chain}
J.~Wei, X.~Wang, D.~Schuurmans, M.~Bosma, F.~Xia, E.~H. Chi, Q.~V. Le, D.~Zhou,
  et~al.
\newblock Chain-of-thought prompting elicits reasoning in large language
  models.
\newblock In \emph{Advances in Neural Information Processing Systems}, 2022.
\newblock URL \url{https://openreview.net/forum?id=_VjQlMeSB_J}.

\bibitem[Wei et~al.(2023)Wei, Wei, Lin, Li, Zhang, Ren, Li, Wan, Cao, Xie, Hu,
  Li, Hui, Yu, Liu, Yang, Huang, and Xie]{wei2023polylmopensourcepolyglot}
X.~Wei, H.~Wei, H.~Lin, T.~Li, P.~Zhang, X.~Ren, M.~Li, Y.~Wan, Z.~Cao, B.~Xie,
  T.~Hu, S.~Li, B.~Hui, B.~Yu, D.~Liu, B.~Yang, F.~Huang, and J.~Xie.
\newblock Polylm: An open source polyglot large language model.
\newblock \emph{arXiv preprint 2307.06018}, 2023.
\newblock URL \url{https://arxiv.org/abs/2307.06018}.

\bibitem[Winata et~al.(2023)Winata, Aji, Cahyawijaya, Mahendra, Koto,
  Romadhony, Kurniawan, Moeljadi, Prasojo, Fung, Baldwin, Lau, Sennrich, and
  Ruder]{nusax}
G.~I. Winata, A.~F. Aji, S.~Cahyawijaya, R.~Mahendra, F.~Koto, A.~Romadhony,
  K.~Kurniawan, D.~Moeljadi, R.~E. Prasojo, P.~Fung, T.~Baldwin, J.~H. Lau,
  R.~Sennrich, and S.~Ruder.
\newblock Nusax: Multilingual parallel sentiment dataset for 10 indonesian
  local languages.
\newblock In \emph{Proceedings of the 17th Conference of the European Chapter
  of the Association for Computational Linguistics}, pages 815--834, Dubrovnik,
  Croatia, May 2023. Association for Computational Linguistics.
\newblock URL \url{https://aclanthology.org/2023.eacl-main.57}.

\bibitem[Wolf et~al.(2019)Wolf, Debut, Sanh, Chaumond, Delangue, Moi, Cistac,
  Rault, Louf, Funtowicz, et~al.]{wolf2019huggingface}
T.~Wolf, L.~Debut, V.~Sanh, J.~Chaumond, C.~Delangue, A.~Moi, P.~Cistac,
  T.~Rault, R.~Louf, M.~Funtowicz, et~al.
\newblock Huggingface's transformers: State-of-the-art natural language
  processing.
\newblock \emph{arXiv preprint}, 2019.

\bibitem[Xu et~al.(2024)Xu, Kim, Sharaf, and
  Awadalla]{xu2024paradigmshiftmachinetranslation}
H.~Xu, Y.~J. Kim, A.~Sharaf, and H.~H. Awadalla.
\newblock A paradigm shift in machine translation: Boosting translation
  performance of large language models.
\newblock \emph{arXiv preprint 2309.11674}, 2024.
\newblock URL \url{https://arxiv.org/abs/2309.11674}.

\bibitem[Xue et~al.(2021)Xue, Constant, Roberts, Kale, Al-Rfou, Siddhant,
  Barua, and Raffel]{xue2021mt5}
L.~Xue, N.~Constant, A.~Roberts, M.~Kale, R.~Al-Rfou, A.~Siddhant, A.~Barua,
  and C.~Raffel.
\newblock {mT5}: A massively multilingual pre-trained text-to-text transformer.
\newblock In \emph{Proceedings of the 2021 Conference of the North American
  Chapter of the Association for Computational Linguistics: Human Language
  Technologies}, pages 483--498, 2021.

\bibitem[Yang et~al.(2024)Yang, Yang, Hui, Zheng, Yu, Zhou, Li, Li, Liu, Huang,
  et~al.]{yang2024qwen2technicalreport}
A.~Yang, B.~Yang, B.~Hui, B.~Zheng, B.~Yu, C.~Zhou, C.~Li, C.~Li, D.~Liu,
  F.~Huang, et~al.
\newblock Qwen2 technical report.
\newblock \emph{arXiv preprint arXiv:2407.10671}, 2024.

\bibitem[Yuan et~al.(2023)Yuan, Lu, Zhu, Kong, Li, Qiao, and
  Xu]{yuan-etal-2023-lego}
F.~Yuan, Y.~Lu, W.~Zhu, L.~Kong, L.~Li, Y.~Qiao, and J.~Xu.
\newblock {L}ego-{MT}: Learning detachable models for massively multilingual
  machine translation.
\newblock In A.~Rogers, J.~Boyd-Graber, and N.~Okazaki, editors, \emph{Findings
  of the Association for Computational Linguistics: ACL 2023}, pages
  11518--11533, Toronto, Canada, July 2023. Association for Computational
  Linguistics.
\newblock \doi{10.18653/v1/2023.findings-acl.731}.
\newblock URL \url{https://aclanthology.org/2023.findings-acl.731}.

\bibitem[Zhang et~al.(2020)Zhang, Frey, and Bansal]{chren}
S.~Zhang, B.~Frey, and M.~Bansal.
\newblock Chren: Cherokee-english machine translation for endangered language
  revitalization.
\newblock In \emph{EMNLP2020}, 2020.

\bibitem[Zhang et~al.(2019)Zhang, Kishore, Wu, Weinberger, and
  Artzi]{zhang2019bertscore}
T.~Zhang, V.~Kishore, F.~Wu, K.~Q. Weinberger, and Y.~Artzi.
\newblock Bertscore: Evaluating text generation with bert.
\newblock \emph{arXiv preprint arXiv:1904.09675}, 2019.

\end{thebibliography}

\clearpage

\begin{appendices}

\section{Frequently Asked Questions}
\label{sec:faq}

\subsection*{Goals and Achievements}

\noindent \textbf{Q:} What is the main goal of this paper?\\
The main goal of this paper, as suggested by its title, is the perform massively multilingual adaptation of large language models using bilingual translation data. Specifically, we compile a massive parallel corpus, compose two data mixes, perform continual pre-training of large language models and comprehensive evaluation, and examine interesting research questions such as the effect of using bilingual translation data.\\

\noindent \textbf{Q:} What are the main achievements and the scientific novelty?\\
\noindent \textbf{A:} Our main achievements are summarized as the contributions as written in the \Cref{sec:introduction} Introduction. For continual pre-training, we do not modify the model architecture---we do not claim this novelty. However, our scientific novelty is more on the resource construction side, including the creation of \MALA translation corpus and continually pre-trained \EMMA models based on Llama 3 and 3.1 on two diverse data mixes. Besides, our empirical evaluation reveals some valuable findings that can potentially promote future development. 

\subsection*{Data}

\noindent \textbf{Q:} How was the data mix tuned?  \\
Data mix optimization is a distinct research area. Our focus was on creating a robust bilingual translation corpus through careful curation for language and content diversity. While we didn't perform specific data mix ablations, we ensured balanced representation across languages and domains and compared two data mixes with or without bilingual translation data to examine the effect of including parallel data for continual pre-training. Searching for the optimal data composition requires a lot of computing resources and remains an interesting direction for future work. \\

\noindent \textbf{Q:} Does whitespace cleaning only apply to whitespace-tokenized languages, or do other languages require different cleaning methods?  \\
\noindent \textbf{A:} Whitespace cleaning is mainly relevant for whitespace-tokenized languages (like English) to standardize text for processing. However, non-whitespace-tokenized languages (e.g., Chinese, Japanese, Korean) may not need this step but often require other pre-processing, such as character normalization or segmentation.  
\\

\noindent \textbf{Q:} How were the existing datasets and their languages selected for inclusion?  \\
\noindent \textbf{A:} The datasets and languages were chosen based on \citet{ji2024emma} adopting the the following criteria:  
\begin{itemize}[nolistsep,noitemsep]  
    \item \textbf{Language Coverage:} Emphasis on broad multilingual representation, including both high- and low-resource languages.  
    \item \textbf{Data Quality:} Preference for high-quality, diverse content (books, papers, code) to enhance model generalization.  
    \item \textbf{Task Relevance:} Alignment with goals for continual pre-training and multilingual adaptation (e.g., translation, classification, reasoning).  
    \item \textbf{Availability:} Prioritization of publicly accessible datasets to ensure transparency and reproducibility.  
\end{itemize}

\subsection*{Evaluation}

\noindent \textbf{Q:}  How do we determine the number of shots for in-context learning?\\
\noindent \textbf{A:} 
We use the same number of shots as \citet{ji2024emma}, where the number of shots used in evaluation follows the setup from Llama 2, which demonstrates strong performance on high-resource languages with this configuration. Rather than performing an exhaustive grid search for optimal shot count, we maintain this established approach. We provide the model weights, enabling researchers to experiment with different numbers of shots according to their specific evaluation needs.\\

\noindent \textbf{Q:} Why no human evaluation?\\
\noindent \textbf{A:} Human evaluation faces scalability challenges in multilingual settings due to subjectivity, cultural biases, and high costs---especially for low-resource languages requiring specialized annotators. While valuable, these limitations make automated metrics more practical for our large-scale study. We acknowledge human evaluation's importance, but emphasize its unsustainability for massively multilingual work.\\

\noindent \textbf{Q:} How reliable is machine-generated benchmark?\\
\noindent \textbf{A:} Machine-generated benchmarks provide scalable, consistent, and objective evaluation across many tasks and languages, enabling efficient model comparison. However, they often lack nuance in assessing creativity, contextual appropriateness, and cultural sensitivity. Potential biases in generation or evaluation metrics may also affect reliability. The study on the reliability of machine-generated benchmarks is out of the scope of this paper. \\

\noindent \textbf{Q:} Why no safety-related evaluation?\\
\noindent \textbf{A:} Safety evaluation is a distinct research area. Our work focuses on CPT, releasing a multilingual \textbf{base} model (not instruction-tuned) and corpus. As this isn't an end-user product, safety wasn't a primary focus. Existing benchmarks like \citet{wang2023all} and \citet{aakanksha2024aya_redteaming} cover a few languages and aren't ideal for massively multilingual settings.\\

\subsection*{Comparison with Baselines}
\noindent \textbf{Q:} Is comparing models on unsupported languages unfair?\\
\noindent \textbf{A:} We evaluate all models across diverse languages, including unsupported ones, to assess generalization capabilities. There's no established standard for ``fair'' baselines in such cross-lingual comparisons. Evaluating on unsupported languages stresses the model's zero-shot transferability.\\

\noindent \textbf{Q:} What is the relation to the EMMA-500 Llama 2 model \citep{ji2024emma}?\\
\noindent \textbf{A:} This work is a follow-up of the EMMA-500 Llama 2 model. Our models follow most of the settings of the EMMA-500 Llama 2 model, but use additional bilingual translation data, continue pre-pretraining on Llama 3 and 3.1 models, and study on two big data mixes. Moreover, our analyses reveals valuable findings. \\

\noindent \textbf{Q:} Why not comparing with model XXX?\\
\noindent \textbf{A:} Our model selection prioritizes representative architectures that have diverse capabilities. Particularly, we focus on open-weight decoder-only models from 7-9B parameters. Our current exclusion reflects either the overlap with existing benchmarks or technical constraints. Also, some newer models would come up. We actively welcome feedback to expand coverage of baselines.

\subsection*{Others}

\noindent \textbf{Q:} Will the corpus, model weights, benchmarks, and scripts be publicly available?\\
\noindent \textbf{A:} Yes. We release the \MALA translation corpus, all model weights, and code. Benchmarks and processing scripts are open-sourced by their original authors. \\

\noindent \textbf{Q:} How do we categorize languages into resource groups?\label{q:language_categorization} \\
\noindent \textbf{A:} We follow the resource groups used by \citet{ji2024emma}, where languages are grouped into five categories (high to low resource) based on token counts in the MaLA monolingual corpus. This data-driven approach reflects actual text availability. We include ISO codes and writing systems in the \MALA parallel corpus for standardization across NLP applications. \\

\noindent \textbf{Q:} Why don't we separate metrics for officially supported vs. unseen languages?\\
\noindent \textbf{A:} While possible for models declaring supported languages, this can't be done for LLaMA or GEMMA which lack such specifications. We include per-language results in the Appendix and will share all model generations on our project website (\website), enabling community analysis with custom metrics while keeping the paper concise.

\section{Details of \MALA Translation Corpus}
\label{sec:mala_bilingual_detailed}

\subsection{Data Sources}
\label{sec:data_sources}

\Cref{tab:bilingual_sources} lists the corpora and collections we use as bilingual data sources in this work. A data source being bilingual means that it contains corresponding, or parallel, text records in two languages. This usually means one side has been translated, either by a machine or a human. Pedantically, massively parallel bilingual corpora can be considered a special case of multilingual corpora containing sentences in different languages that share the same meaning. In this work, some data sources have parallel text data in more than two languages, but in such cases, we always iteratively determine one as the source and one as the target language, i.e., create bilingual corpora out of them. In some source datasets, the source language is not explicitly stated, and as such we pick one language to regard as the source language without actual knowledge of the direction of translation. In such cases, we tend to pick a high-resource language, such as English, as the source language.

 \paragraph{Metadata}

 In the case of bilingual corpora, we define each record in the \texttt{JSONL} file to consist of the fields \texttt{src\_lang}, \texttt{src\_txt}, \texttt{tgt\_lang}, \texttt{tgt\_txt}, \texttt{url}, \texttt{collection}, \texttt{source}, \texttt{original\_src\_lang}, and \texttt{original\_tgt\_lang}. The contents of these fields are as follows. The fields \texttt{src\_txt} and \texttt{tgt\_txt} contain the parallel text data in the source and target language, respectively, and \texttt{src\_lang} and \texttt{tgt\_lang} are the ISO 639-3 language codes of those languages. Identically with the monolingual corpora, \texttt{url} contains a URL indicating the web address from which the text data has been extracted, if available, \texttt{collection} the name of the collection, and \texttt{source} the name of a more specific part of the collection which the text data was extracted from. Lastly, \texttt{original\_src\_lang} and \texttt{original\_tgt\_lang} contain the language denotations of the source and target language, respectively, as they are given in the source data.

 \begin{table*}[ht!]
 \scriptsize
 \centering
 \caption{Source datasets used for compiling the \MALA bilingual translation data.}
 \label{tab:bilingual_sources}
 \begin{tabular}{p{5cm} p{2cm} p{8cm}} 
 \toprule
\textbf{Dataset} & \textbf{Languages}         & \textbf{URL} \\
 \midrule
 CMU Haitian Creole~\citep{haitian_creole}	&	Haitian Creole	&	\url{http://www.speech.cs.cmu.edu/haitian/}	\\
 English-Luganda Parallel Corpus~\citep{english-luganda_parallel_corpus}	&	Luganda-English	&	\url{https://zenodo.org/records/4764039}	\\
 Ewe language corpus~\citep{gbedevi2024}	&	Ewe—English	&	\url{https://www.kaggle.com/datasets/yvicherita/ewe-language-corpus}	\\
 Nunavut Hansard Inuktitut–English Parallel Corpus 3.0~\citep{nunavut-hansard}	&	Inuktitut–English	&	\url{https://nrc-digital-repository.canada.ca/eng/view/object/?id=c7e34fa7-7629-43c2-bd6d-19b32bf64f60}	\\
 AmericasNLP 2021~\citep{americasnlp2023_source2,americasnlp2023_source11,americasnlp2021_source3}	&	10	&	\url{https://turing.iimas.unam.mx/americasnlp/americasnlp_2021.html;https://github.com/AmericasNLP/americasnlp2021}	\\
 AmericasNLP 2023~\citep{americasnlp2023_source1,americasnlp2023_source2,americasnlp2023_source3,americasnlp2023_source9,americasnlp2023_source11}	&	11	&	\url{https://turing.iimas.unam.mx/americasnlp/2023_st.html}	\\
 AmericasNLP 2022	&	4	&	\url{https://github.com/AmericasNLP/americasnlp2022}	\\
 Indigenous Languages Corpora~\citep{IndigenousLanguages_Corpora}	&	Cree	&	\url{https://github.com/EdTeKLA/IndigenousLanguages_Corpora}	\\
 ACES~\citep{aces}	&	146	&	\url{https://huggingface.co/datasets/nikitam/ACES}	\\
 ChrEn~\citep{chren}	&	2	&	\url{https://huggingface.co/datasets/chr_en}	\\
 NusaX-MT~\citep{nusax}	&	12	&	\url{https://huggingface.co/datasets/indonlp/NusaX-MT}	\\
 Lego-MT~\citep{yuan-etal-2023-lego}	&	433	&	\url{https://github.com/CONE-MT/Lego-MT}	\\
 Tatoeba~\citep{tatoeba}	&	487	&	\url{https://github.com/Helsinki-NLP/Tatoeba-Challenge/tree/master/data}	\\
 NLLB~\citep{nllb2022}	&	202	&	\url{https://huggingface.co/datasets/allenai/nllb}	\\
 Lacuna Project~\citep{lacuna_pos_ner}	&	1	&	\url{https://github.com/masakhane-io/lacuna_pos_ner}	\\
 lafand-mt~\citep{lafand-mt}	&	21	&	\url{https://github.com/masakhane-io/lafand-mt/tree/main}	\\                                                 
 \bottomrule
 \end{tabular}
 \end{table*}

\subsection{Supported Languages}
\label{sec:supported_languages}
\Cref{tab:languages} shows the language codes of the MaLA corpus \citep{ji2024emma}, where ``unseen'' means the languages are not used for training \EMMA.
The classification system for token counts categorizes language resources based on their size into five distinct tiers: ``high'' for resources exceeding 1 billion tokens, indicating a vast amount of data; ``medium-high'' for those with more than 100 million tokens, reflecting a substantial dataset; ``medium'' for resources that contain over 10 million tokens, representing a moderate size; ``medium-low'' for datasets with over ``1 million tokens'', indicating a smaller yet significant amount of data; and finally, ``low'' for resources containing less than 1 million tokens, which suggests a minimal data presence. This hierarchy helps in understanding the scale and potential utility of the language resources available.

\begin{table*}[t!]
\caption{Languages by resource groups categorized by counting the number of tokens in the MaLA monolingual corpus \citep{ji2024emma}. ``Unseen'' means those languages are not used for continual pre-training in this paper.}
\label{tab:languages}
\centering
\tiny
\begin{tabular}{lrp{12cm}}
\toprule
\textbf{Category} & \textbf{Languages} & \textbf{Language Codes} \\
\midrule
high & 27 & fra\_Latn, mon\_Cyrl, kat\_Geor, tgk\_Cyrl, kaz\_Cyrl, glg\_Latn, hbs\_Latn, kan\_Knda, mal\_Mlym, rus\_Cyrl, cat\_Latn, hye\_Armn, guj\_Gujr, slv\_Latn, fil\_Latn, bel\_Cyrl, isl\_Latn, nep\_Deva, mlt\_Latn, pan\_Guru, afr\_Latn, urd\_Arab, mkd\_Cyrl, aze\_Latn, deu\_Latn, eng\_Latn, ind\_Latn \\
low & 210 & prs\_Arab, nqo\_Nkoo, emp\_Latn, pfl\_Latn, teo\_Latn, gpe\_Latn, izz\_Latn, shn\_Mymr, hak\_Latn, pls\_Latn, evn\_Cyrl, djk\_Latn, toj\_Latn, nog\_Cyrl, ctu\_Latn, tca\_Latn, jiv\_Latn, ach\_Latn, mrj\_Latn, ajp\_Arab, apc\_Arab, tab\_Cyrl, hvn\_Latn, tls\_Latn, bak\_Latn, ndc\_Latn, trv\_Latn, top\_Latn, kjh\_Cyrl, guh\_Latn, mni\_Mtei, csy\_Latn, noa\_Latn, dov\_Latn, bho\_Deva, kon\_Latn, hne\_Deva, kcg\_Latn, mni\_Beng, hus\_Latn, pau\_Latn, jbo\_Latn, dtp\_Latn, kmb\_Latn, hau\_Arab, pdc\_Latn, nch\_Latn, acf\_Latn, bim\_Latn, ixl\_Latn, dty\_Deva, kas\_Arab, lrc\_Arab, alz\_Latn, lez\_Cyrl, lld\_Latn, tdt\_Latn, acm\_Arab, bih\_Deva, mzh\_Latn, guw\_Latn, rop\_Latn, rwo\_Latn, ahk\_Latn, qub\_Latn, kri\_Latn, gub\_Latn, laj\_Latn, sxn\_Latn, luo\_Latn, tly\_Latn, pwn\_Latn, mag\_Deva, xav\_Latn, bum\_Latn, ubu\_Latn, roa\_Latn, mah\_Latn, tsg\_Latn, gcr\_Latn, arn\_Latn, csb\_Latn, guc\_Latn, bat\_Latn, knj\_Latn, cre\_Latn, bus\_Latn, anp\_Deva, aln\_Latn, nah\_Latn, zai\_Latn, kpv\_Cyrl, enq\_Latn, gvl\_Latn, wal\_Latn, fiu\_Latn, swh\_Latn, crh\_Latn, nia\_Latn, bqc\_Latn, map\_Latn, atj\_Latn, npi\_Deva, bru\_Latn, din\_Latn, pis\_Latn, gur\_Latn, cuk\_Latn, zne\_Latn, cdo\_Latn, lhu\_Latn, pcd\_Latn, mas\_Latn, bis\_Latn, ncj\_Latn, ibb\_Latn, tay\_Latn, bts\_Latn, tzj\_Latn, bzj\_Latn, cce\_Latn, jvn\_Latn, ndo\_Latn, rug\_Latn, koi\_Cyrl, mco\_Latn, fat\_Latn, olo\_Latn, inb\_Latn, mkn\_Latn, qvi\_Latn, mak\_Latn, ktu\_Latn, nrm\_Latn, kua\_Latn, san\_Latn, nbl\_Latn, kik\_Latn, dyu\_Latn, sgs\_Latn, msm\_Latn, mnw\_Latn, zha\_Latn, sja\_Latn, xal\_Cyrl, rmc\_Latn, ami\_Latn, sda\_Latn, tdx\_Latn, yap\_Latn, tzh\_Latn, sus\_Latn, ikk\_Latn, bas\_Latn, nde\_Latn, dsb\_Latn, seh\_Latn, knv\_Latn, amu\_Latn, dwr\_Latn, iku\_Cans, uig\_Latn, bxr\_Cyrl, tcy\_Knda, mau\_Latn, aoj\_Latn, gor\_Latn, cha\_Latn, fip\_Latn, chr\_Cher, mdf\_Cyrl, arb\_Arab, quw\_Latn, shp\_Latn, spp\_Latn, frp\_Latn, ape\_Latn, cbk\_Latn, mnw\_Mymr, mfe\_Latn, jam\_Latn, lad\_Latn, awa\_Deva, mad\_Latn, ote\_Latn, shi\_Latn, btx\_Latn, maz\_Latn, ppk\_Latn, smn\_Latn, twu\_Latn, blk\_Mymr, msi\_Latn, naq\_Latn, tly\_Arab, wuu\_Hani, mos\_Latn, cab\_Latn, zlm\_Latn, gag\_Latn, suz\_Deva, ksw\_Mymr, gug\_Latn, nij\_Latn, nov\_Latn, srm\_Latn, jac\_Latn, nyu\_Latn, yom\_Latn, gui\_Latn \\
medium & 68 & tha\_Thai, kat\_Latn, lim\_Latn, tgk\_Arab, che\_Cyrl, lav\_Latn, xho\_Latn, war\_Latn, nan\_Latn, grc\_Grek, orm\_Latn, zsm\_Latn, cnh\_Latn, yor\_Latn, arg\_Latn, tgk\_Latn, azj\_Latn, tel\_Latn, slk\_Latn, pap\_Latn, zho\_Hani, sme\_Latn, tgl\_Latn, uzn\_Cyrl, als\_Latn, san\_Deva, azb\_Arab, ory\_Orya, lmo\_Latn, bre\_Latn, mvf\_Mong, fao\_Latn, oci\_Latn, sah\_Cyrl, sco\_Latn, tuk\_Latn, aze\_Arab, hin\_Deva, haw\_Latn, glk\_Arab, oss\_Cyrl, lug\_Latn, tet\_Latn, tsn\_Latn, hrv\_Latn, gsw\_Latn, arz\_Arab, vec\_Latn, mon\_Latn, ilo\_Latn, ctd\_Latn, ben\_Beng, roh\_Latn, kal\_Latn, asm\_Beng, srp\_Latn, bod\_Tibt, hif\_Latn, rus\_Latn, nds\_Latn, lus\_Latn, ido\_Latn, lao\_Laoo, tir\_Ethi, chv\_Cyrl, wln\_Latn, kaa\_Latn, pnb\_Arab \\
medium-high & 79 & div\_Thaa, som\_Latn, jpn\_Japn, hat\_Latn, sna\_Latn, heb\_Hebr, bak\_Cyrl, nld\_Latn, tel\_Telu, kin\_Latn, msa\_Latn, gla\_Latn, bos\_Latn, dan\_Latn, smo\_Latn, ita\_Latn, mar\_Deva, pus\_Arab, srp\_Cyrl, spa\_Latn, lat\_Latn, hmn\_Latn, sin\_Sinh, zul\_Latn, bul\_Cyrl, amh\_Ethi, ron\_Latn, tam\_Taml, khm\_Khmr, nno\_Latn, cos\_Latn, fin\_Latn, ori\_Orya, uig\_Arab, hbs\_Cyrl, gle\_Latn, cym\_Latn, vie\_Latn, kor\_Hang, lit\_Latn, yid\_Hebr, ara\_Arab, sqi\_Latn, pol\_Latn, tur\_Latn, swa\_Latn, hau\_Latn, ceb\_Latn, eus\_Latn, kir\_Cyrl, mlg\_Latn, jav\_Latn, snd\_Arab, sot\_Latn, por\_Latn, uzb\_Cyrl, fas\_Arab, nor\_Latn, est\_Latn, hun\_Latn, ibo\_Latn, ltz\_Latn, swe\_Latn, tat\_Cyrl, ast\_Latn, mya\_Mymr, uzb\_Latn, sun\_Latn, ell\_Grek, ces\_Latn, mri\_Latn, ckb\_Arab, kur\_Latn, kaa\_Cyrl, nob\_Latn, ukr\_Cyrl, fry\_Latn, epo\_Latn, nya\_Latn \\
medium-low & 162 & aym\_Latn, rue\_Cyrl, rom\_Latn, dzo\_Tibt, poh\_Latn, sat\_Olck, ary\_Arab, fur\_Latn, mbt\_Latn, bpy\_Beng, iso\_Latn, pon\_Latn, glv\_Latn, new\_Deva, gym\_Latn, bgp\_Latn, kac\_Latn, abt\_Latn, quc\_Latn, otq\_Latn, sag\_Latn, cak\_Latn, avk\_Latn, pam\_Latn, meo\_Latn, tum\_Latn, bam\_Latn, kha\_Latn, syr\_Syrc, kom\_Cyrl, nhe\_Latn, bal\_Arab, srd\_Latn, krc\_Cyrl, lfn\_Latn, bar\_Latn, rcf\_Latn, nav\_Latn, nnb\_Latn, sdh\_Arab, aka\_Latn, bew\_Cyrl, bbc\_Latn, meu\_Latn, zza\_Latn, ext\_Latn, yue\_Hani, ekk\_Latn, xmf\_Geor, nap\_Latn, mzn\_Arab, pcm\_Latn, lij\_Latn, myv\_Cyrl, scn\_Latn, dag\_Latn, ban\_Latn, twi\_Latn, udm\_Cyrl, som\_Arab, nso\_Latn, pck\_Latn, crs\_Latn, acr\_Latn, tat\_Latn, afb\_Arab, uzs\_Arab, hil\_Latn, mgh\_Latn, tpi\_Latn, ady\_Cyrl, pag\_Latn, kiu\_Latn, ber\_Latn, iba\_Latn, ksh\_Latn, plt\_Latn, lin\_Latn, chk\_Latn, tzo\_Latn, tlh\_Latn, ile\_Latn, lub\_Latn, hui\_Latn, min\_Latn, bjn\_Latn, szl\_Latn, kbp\_Latn, inh\_Cyrl, que\_Latn, ven\_Latn, vls\_Latn, kbd\_Cyrl, run\_Latn, wol\_Latn, ace\_Latn, ada\_Latn, kek\_Latn, yua\_Latn, tbz\_Latn, gom\_Latn, ful\_Latn, mrj\_Cyrl, abk\_Cyrl, tuc\_Latn, stq\_Latn, mwl\_Latn, tvl\_Latn, quh\_Latn, gom\_Deva, mhr\_Cyrl, fij\_Latn, grn\_Latn, zap\_Latn, mam\_Latn, mps\_Latn, tiv\_Latn, ksd\_Latn, ton\_Latn, bik\_Latn, vol\_Latn, ava\_Cyrl, tso\_Latn, szy\_Latn, ngu\_Latn, hyw\_Armn, fon\_Latn, skr\_Arab, kos\_Latn, tyz\_Latn, kur\_Arab, srn\_Latn, tyv\_Cyrl, bci\_Latn, vep\_Latn, crh\_Cyrl, kpg\_Latn, hsb\_Latn, ssw\_Latn, zea\_Latn, ewe\_Latn, ium\_Latn, diq\_Latn, ltg\_Latn, nzi\_Latn, guj\_Deva, ina\_Latn, pms\_Latn, bua\_Cyrl, lvs\_Latn, eml\_Latn, hmo\_Latn, kum\_Cyrl, kab\_Latn, chm\_Cyrl, cor\_Latn, cfm\_Latn, alt\_Cyrl, bcl\_Latn, ang\_Latn, frr\_Latn, mai\_Deva \\
unseen & 393 & rap\_Latn, pmf\_Latn, lsi\_Latn, dje\_Latn, bkx\_Latn, ipk\_Latn, syw\_Deva, ann\_Latn, bag\_Latn, bat\_Cyrl, chu\_Cyrl, gwc\_Arab, adh\_Latn, szy\_Hani, shi\_Arab, njy\_Latn, pdu\_Latn, buo\_Latn, cuv\_Latn, udg\_Mlym, bax\_Latn, tio\_Latn, kjb\_Latn, taj\_Deva, lez\_Latn, olo\_Cyrl, rnl\_Latn, bri\_Latn, inh\_Latn, kas\_Cyrl, wni\_Latn, anp\_Latn, tsc\_Latn, mgg\_Latn, udi\_Cyrl, mdf\_Latn, agr\_Latn, xty\_Latn, llg\_Latn, nge\_Latn, gan\_Latn, tuv\_Latn, stk\_Latn, nut\_Latn, thy\_Thai, lgr\_Latn, hnj\_Latn, dar\_Cyrl, aia\_Latn, lwl\_Thai, tnl\_Latn, tvs\_Latn, jra\_Khmr, tay\_Hani, gal\_Latn, ybi\_Deva, snk\_Arab, gag\_Cyrl, tuk\_Cyrl, trv\_Hani, ydd\_Hebr, kea\_Latn, gbm\_Deva, kwi\_Latn, hro\_Latn, rki\_Latn, quy\_Latn, tdg\_Deva, zha\_Hani, pcg\_Mlym, tom\_Latn, nsn\_Latn, quf\_Latn, jmx\_Latn, kqr\_Latn, mrn\_Latn, bxa\_Latn, abc\_Latn, mve\_Arab, lfa\_Latn, qup\_Latn, yin\_Latn, roo\_Latn, mrw\_Latn, nxa\_Latn, yrk\_Cyrl, bem\_Latn, kvt\_Latn, csw\_Cans, bjr\_Latn, mgm\_Latn, ngn\_Latn, pib\_Latn, quz\_Latn, awb\_Latn, myk\_Latn, otq\_Arab, ino\_Latn, tkd\_Latn, bef\_Latn, bug\_Bugi, aeu\_Latn, nlv\_Latn, dty\_Latn, bkc\_Latn, mmu\_Latn, hak\_Hani, sea\_Latn, mlk\_Latn, cbr\_Latn, lmp\_Latn, tnn\_Latn, qvz\_Latn, pbt\_Arab, cjs\_Cyrl, mlw\_Latn, mnf\_Latn, bfm\_Latn, dig\_Latn, thk\_Latn, zxx\_Latn, lkb\_Latn, chr\_Latn, pnt\_Latn, vif\_Latn, fli\_Latn, got\_Latn, hbb\_Latn, tll\_Latn, bug\_Latn, kxp\_Arab, qaa\_Latn, krr\_Khmr, kjg\_Laoo, isu\_Latn, kmu\_Latn, gof\_Latn, sdk\_Latn, mne\_Latn, baw\_Latn, idt\_Latn, xkg\_Latn, mgo\_Latn, dtr\_Latn, kms\_Latn, ffm\_Latn, hna\_Latn, nxl\_Latn, bfd\_Latn, odk\_Arab, miq\_Latn, mhx\_Latn, kam\_Latn, yao\_Latn, pnt\_Grek, kby\_Latn, kpv\_Latn, kbx\_Latn, cim\_Latn, qvo\_Latn, pih\_Latn, nog\_Latn, nco\_Latn, rmy\_Cyrl, clo\_Latn, dmg\_Latn, aaa\_Latn, rel\_Latn, ben\_Latn, loh\_Latn, thl\_Deva, chd\_Latn, cni\_Latn, cjs\_Latn, lbe\_Latn, ybh\_Deva, zxx\_Zyyy, awa\_Latn, gou\_Latn, xmm\_Latn, nqo\_Latn, rut\_Cyrl, kbq\_Latn, tkr\_Latn, dwr\_Ethi, ckt\_Cyrl, ady\_Latn, yea\_Mlym, nhx\_Latn, niv\_Cyrl, bwt\_Latn, xmg\_Latn, chy\_Latn, mfj\_Latn, hre\_Latn, bbk\_Latn, shn\_Latn, lrc\_Latn, qvc\_Latn, muv\_Mlym, mdr\_Latn, luy\_Latn, lzh\_Hani, fuh\_Latn, mle\_Latn, brx\_Deva, pex\_Latn, kau\_Latn, yrk\_Latn, hin\_Latn, ekm\_Latn, msb\_Latn, unr\_Orya, cac\_Latn, chp\_Cans, ckt\_Latn, bss\_Latn, lts\_Latn, bbj\_Latn, ttt\_Cyrl, kwu\_Latn, smn\_Cyrl, kpy\_Cyrl, tod\_Latn, wbm\_Latn, tcy\_Latn, arc\_Syrc, nst\_Latn, tuz\_Latn, bob\_Latn, bfn\_Latn, pli\_Deva, snl\_Latn, kwd\_Latn, lgg\_Latn, nza\_Latn, wbr\_Deva, lan\_Latn, kmz\_Latn, bzi\_Thai, hao\_Latn, nla\_Latn, qxr\_Latn, ken\_Latn, tbj\_Latn, blk\_Latn, ybb\_Latn, nwe\_Latn, gan\_Hani, snk\_Latn, kak\_Latn, tpl\_Latn, hla\_Latn, tks\_Arab, pea\_Latn, bya\_Latn, enc\_Latn, jgo\_Latn, tnp\_Latn, aph\_Deva, bgf\_Latn, brv\_Laoo, nod\_Thai, niq\_Latn, nwi\_Latn, xmd\_Latn, gbj\_Orya, syr\_Latn, ify\_Latn, xal\_Latn, bra\_Deva, cgc\_Latn, bhs\_Latn, pwg\_Latn, ang\_Runr, oki\_Latn, qve\_Latn, qvm\_Latn, bkm\_Latn, bkh\_Latn, niv\_Latn, zuh\_Latn, mry\_Latn, fiu\_Cyrl, ssn\_Latn, rki\_Mymr, sox\_Latn, yav\_Latn, nyo\_Latn, dag\_Arab, qxh\_Latn, bze\_Latn, myx\_Latn, zaw\_Latn, ddg\_Latn, wnk\_Latn, bwx\_Latn, mqy\_Latn, lad\_Hebr, boz\_Latn, lue\_Latn, ded\_Latn, pli\_Latn, avk\_Cyrl, wms\_Latn, sgd\_Latn, azn\_Latn, ajz\_Latn, psp\_Latn, jra\_Latn, smt\_Latn, ags\_Latn, csw\_Latn, wtk\_Latn, emp\_Cyrl, koi\_Latn, tkr\_Cyrl, amp\_Latn, ymp\_Latn, mfh\_Latn, tdb\_Deva, omw\_Latn, khb\_Talu, doi\_Deva, gld\_Cyrl, ava\_Latn, chu\_Latn, dnw\_Latn, azo\_Latn, dug\_Latn, bce\_Latn, kmr\_Latn, kpy\_Armn, abq\_Cyrl, trp\_Latn, ewo\_Latn, the\_Deva, hig\_Latn, pkb\_Latn, mxu\_Latn, oji\_Latn, tnt\_Latn, mzm\_Latn, mns\_Cyrl, lbe\_Cyrl, qvh\_Latn, kmg\_Latn, sps\_Latn, brb\_Khmr, tah\_Latn, sxb\_Latn, mkz\_Latn, mgq\_Latn, got\_Goth, lns\_Latn, arc\_Latn, akb\_Latn, skr\_Latn, nsk\_Cans, sml\_Latn, pce\_Mymr, eee\_Thai, lhm\_Deva, yux\_Cyrl, bqm\_Latn, bcc\_Arab, nas\_Latn, agq\_Latn, xog\_Latn, tsb\_Latn, fub\_Latn, mqj\_Latn, nsk\_Latn, bxr\_Latn, dln\_Latn, ozm\_Latn, rmy\_Latn, cre\_Cans, kim\_Cyrl, cuh\_Latn, ngl\_Latn, yas\_Latn, bud\_Latn, miy\_Latn, ame\_Latn, pnz\_Latn, raj\_Deva, enb\_Latn, cmo\_Khmr, saq\_Latn, tpu\_Khmr, eve\_Cyrl, cdo\_Hani \\
\bottomrule
\end{tabular}
\end{table*}

\subsection{Token Counts}

\begin{figure*}[ht!]
    \centering
    \begin{subfigure}[b]{\textwidth}
        \centering
        \includegraphics[width=0.8\textwidth]{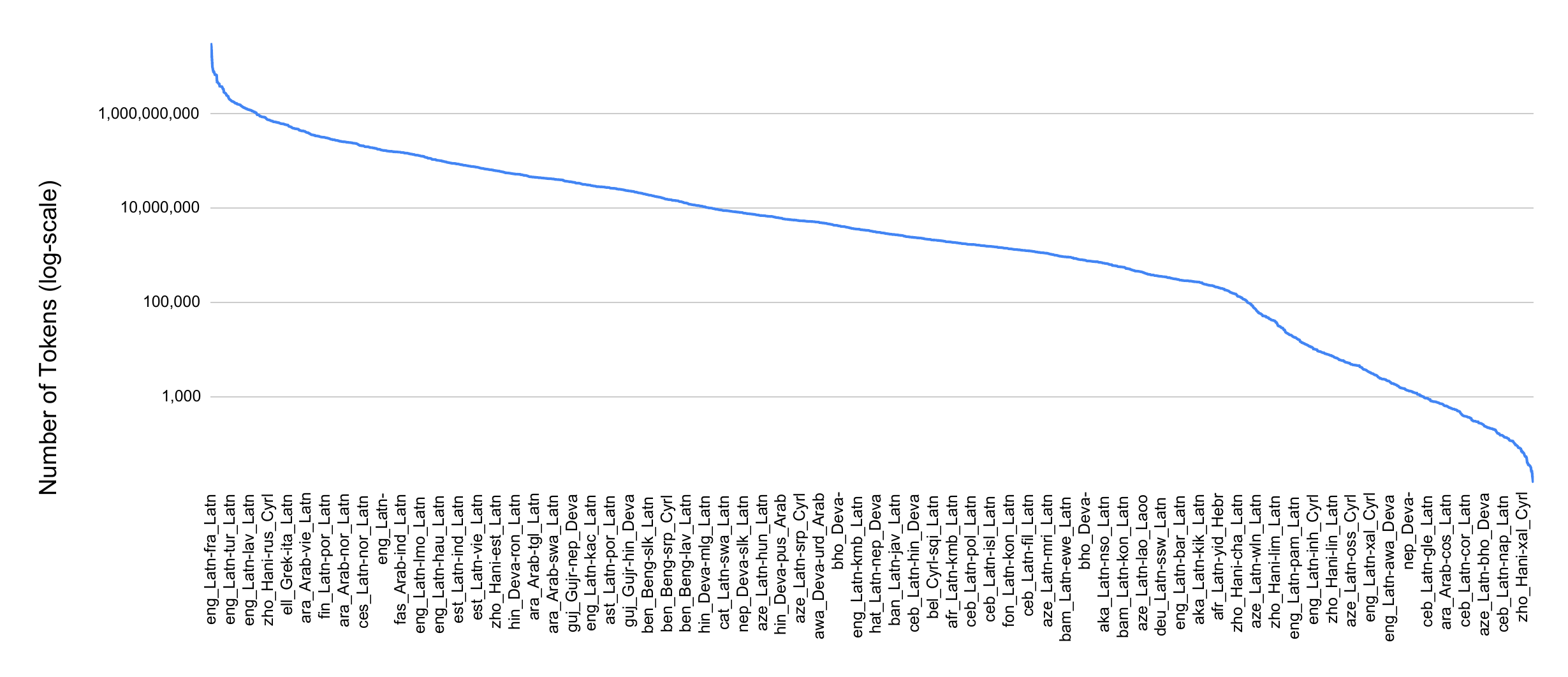}
        \caption{Number of tokens}
        \label{fig:bilingual_token_counts}
    \end{subfigure}
    \begin{subfigure}[b]{\textwidth}
        \centering
        \includegraphics[width=0.8\textwidth]{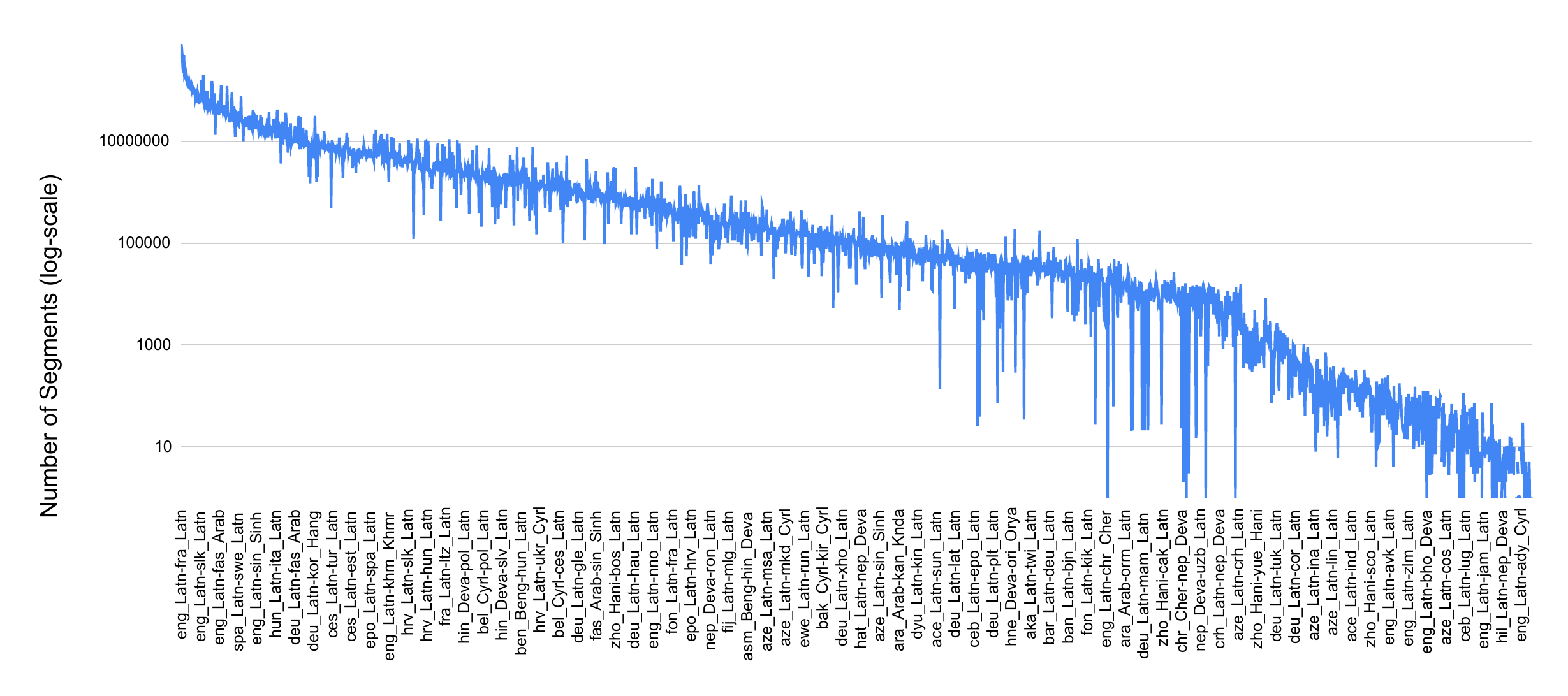}
        \caption{Number of segments}
        \label{fig:bilingual_segment_counts}
    \end{subfigure}  
    \caption{Numbers of segments and tokens across all language pairs in the \MALA bilingual translation corpus.}
    \label{fig:bilingual_counts}
\end{figure*}

\Cref{fig:bilingual_counts} shows the numbers of segments and tokens across all language pairs in the \MALA translation corpus. 

\section{Details of Data Mixes}
\label{sec:data_mix_detailed}

\Cref{tab:data_mix} provides how different data types and resource categories contribute to the overall dataset composition. 
Our data mixes aim to make a balanced distribution over high- and low-resources.
However, due to the nature of high-resource languages, they still contribute to a large portion.
For example, English research papers form a substantial portion.
Medium-high and medium-resource monolingual and bilingual texts contribute a lot.
Medium and low-resource languages consist of 33\% and 19\% of bilingual and monolingual mixes, respectively.

\subsection{Additional Code Data}
\label{sec:additional_code}

\paragraph{Code Data} We source code data from a de-duplicated version of The Stack \citep{kocetkov2023stack} and oversample files from the algorithmic coding \citep{DBLP:conf/nips/Puri0JZDZD0CDTB21} and data science domains~\footnote{\url{https://huggingface.co/datasets/AlgorithmicResearchGroup/arxiv_research_code}} owing to the benefits of training models on self-contained code \citep{fujii2025rewritingpretrainingdataboosts}. We discard all non-data programming languages that occur fewer than 50k times, with the exception of \texttt{llvm}, following prior work detailing its importance in multi-lingual and low-resource code generation \citep{szafraniec2022translation, paul2024ircoder}. We also discard samples that manifest code in rare but valid extensions. Finally, we source from data-heavy formats but follow precedent \citep{lozhkov2024starcoder} and subsample them more aggressively. The surviving data is filtered as follows:

\begin{itemize}
    \item For files forked more than 25 times, we retain them if the average line length is less than 120, the maximum line length is less than 300, and the alphanumeric fraction is more than 30\%.
    \item For files forked between 15 and 25 times, we retain them if the average line length is less than 90, the maximum line length is less than 150, and the alphanumeric fraction is more than 40\%.
    \item For files forked less than 15 times, we retain them if the average line length is less than 80, the maximum line length is less than 120, and the alphanumeric fraction is more than 45\%.
\end{itemize}

\paragraph{Code-Adjacent Procedural Data} We augment our data mix with procedural code-adjacent data ranging from instructive data such as StackOverflow QnAs, library documentation and tutorials~\footnote{\url{https://huggingface.co/datasets/BEE-spoke-data/code-tutorials-en}} to Jupyter notebooks, synthetic code textbooks \citep{DBLP:journals/corr/abs-2412-02595}, GitHub commits, and issue descriptions \citep{lozhkov2024starcoder}. We further mirror \citet{DBLP:conf/iclr/PaulYGKG25} in sourcing parallel math problem solutions to code data.

\begin{table*}[ht!]
\caption{Composition of training data mixtures showing original and final token counts by data type, with sampling rates and distribution percentages across bilingual and monolingual configurations.}
\label{tab:data_mix}
\setlength{\tabcolsep}{5pt}
\scriptsize
\centering
\begin{tabular}{l|l|l|ll|cc}
\toprule
\multirow{2}{*}{\textbf{Type}} & \multirow{2}{*}{\textbf{Category}} & \textbf{Sample} & \multicolumn{2}{c}{\textbf{Token Counts}}                                  & \multicolumn{2}{|c}{\textbf{Percentage of Mixes}}                                           \\
                               &                                    & \textbf{Rate}   & \multicolumn{1}{c}{\textbf{Original}} & \multicolumn{1}{c}{\textbf{Final}} & \multicolumn{1}{|c}{\textbf{Bilingual}} & \multicolumn{1}{c}{\textbf{Monolingual}} \\
\midrule
\multirow{7}{*}{instruction} & EN                        & 0.1    & 9,204,199,807   & 920,419,981    & 0.32\%        & 0.47\%         \\
                             & high                      & 0.2    & 39,403,448,029  & 7,880,689,606  & 2.72\%        & 4.01\%         \\
                             & medium-high               & 0.5    & 30,651,187,534  & 15,325,593,767 & 5.28\%        & 7.81\%         \\
                             & medium                    & 5.0    & 1,444,764,863   & 7,223,824,315  & 2.49\%        & 3.68\%         \\
                             & medium-low                & 20.0   & 47,691,495      & 953,829,900    & 0.33\%        & 0.49\%         \\
                             & low                       & 50.0   & 3,064,796       & 153,239,800    & 0.05\%        & 0.08\%         \\
                             & code/reasoning            & 1.0    & 612,208,775     & 612,208,775    & 0.21\%        & 0.31\%         \\
\hline
code                         & code                      & 1.0    & 43,478,432,765  & 43,478,432,765 & 14.99\%       & 22.15\%        \\
book                         & Gutenberg                 & 1.0    & 5,173,357,710   & 5,173,357,710  & 1.78\%        & 2.64\%         \\
\hline
\multirow{2}{*}{paper}       & EN                        & 1.0    & 38,256,934,181  & 38,256,934,181 & 13.19\%       & 19.49\%        \\
                             & ZH                        & 1.0    & 61,787,372      & 61,787,372     & 0.02\%        & 0.03\%         \\
\hline
\multirow{6}{*}{monolingual} & EN                        & 0.1    & 3,002,029,817   & 300,202,982    & 0.10\%        & 0.15\%         \\
                             & high                      & 0.5    & 40,411,201,964  & 20,205,600,982 & 6.97\%        & 10.29\%        \\
                             & medium-high               & 1.0    & 27,515,227,962  & 27,515,227,962 & 9.49\%        & 14.02\%        \\
                             & medium                    & 5.0    & 2,747,484,380   & 13,737,421,900 & 4.74\%        & 7.00\%         \\
                             & medium-low                & 20.0   & 481,935,633     & 9,638,712,660  & 3.32\%        & 4.91\%         \\
                             & low                       & 50.0   & 97,535,696      & 4,876,784,800  & 1.68\%        & 2.48\%         \\
\hline
\multirow{7}{*}{bilingual}   & very high                 & 0.1    & 85,001,097,362  & 4,250,054,868  & 1.47\%        & 0.00\%         \\
                             & high                      & 0.1    & 207,688,940,222 & 20,768,894,022 & 7.16\%        & 0.00\%         \\
                             & medium-high               & 0.2    & 46,777,497,823  & 9,355,499,565  & 3.23\%        & 0.00\%         \\
                             & medium                    & 0.5    & 64,375,100,302  & 32,187,550,151 & 11.10\%       & 0.00\%         \\
                             & medium-low                & 1.0    & 20,361,578,347  & 20,361,578,347 & 7.02\%        & 0.00\%         \\
                             & low                       & 2.0    & 2,503,240,829   & 5,006,481,658  & 1.73\%        & 0.00\%         \\
                             & very low                  & 10.0   & 175,309,923     & 1,753,099,230  & 0.60\%        & 0.00\%         \\
\bottomrule
\end{tabular}
\end{table*}

\section{Detailed Evaluation Setup}
\label{sec:detailed_setup}

\subsection{Tasks and Benchmarks}
\label{sec:benchmarks}

\Cref{tab:downstream_tasks} summarizes the tasks and benchmarks used for evaluation. 

\paragraph{Text Classification}
SIB-200~\cite{sib-200} and Taxi-1500~\cite{ma2023taxi1500} are two representative multilingual topic-classification benchmarks. SIB-200 is based on Flores-200, covering 205 languages and dialects. The topics are "science/technology", "travel", "politics", "sports", "health", "entertainment", and "geography".
Taxi-1500 is a sentence classification data set with 6 topics, i.e., recommendation, faith, description, sin, grace, and violence, collected from the Parallel Bible Corpus \citep{pbc}. It covers 1,502 typologically diverse languages spanning 112 language families. 

\paragraph{Commonsense Reasoning}

We evaluate \EMMA models' commonsense-reasoning skills with two multilingual benchmarks. XCOPA \citep{ponti-etal-2020-xcopa} targets causal inference across 11 languages, XStoryCloze \citep{lin2022few} gauges narrative completion in 11 languages.

\paragraph{Natural Language Inference}

We evaluate on XNLI \citep{conneau2018xnli}, where sentence pairs in different languages need to be classified as entailment, contradiction,
or neutral.

\paragraph{Machine Translation}
FLORES-200 \citep{costa2022no} is a multilingual benchmark designed to evaluate machine translation performance across 204 language pairs involving English, yielding 408 translation directions with an emphasis on low-resource settings. 

\paragraph{Text Summarization}
We use two multilingual news summarization datasets: MassiveSumm \citep{varab-schluter-2021-massivesumm} and XL-Sum \citep{hasan-etal-2021-xl}.
We subsample MassiveSumm into two sets, i.e., \textbf{MassiveSumm long}\footnote{\msummlong} designed for longer texts, aiming for a median of 5500 tokens within a range of 3500 to 7500 tokens, and \textbf{MassiveSumm short}\footnote{\msummshort} on shorter texts, targeting a median of 1200 tokens with a maximum of 1500 tokens. 
For practical reasons, we ensure a balanced number of documents across all languages, with a minimum of 100 and a maximum of 2500 documents per language in both MassiveSumm subsets. 
Additionally, we utilized the XL-Sum, a diverse dataset containing over one million professionally annotated article-summary pairs across 44 languages from the BBC.

\paragraph{Machine Reading Comprehension}
We use two datasets of machine reading comprehension: BELEBELE \citep{bandarkar2023belebele} and the multilingual ARC challenge. 
The BELEBELE dataset \citep{bandarkar2023belebele} is a parallel multilingual machine reading comprehension benchmark spanning 122 languages across resource levels. 
Each of its carefully validated multiple-choice questions (with four options per question) derives from FLORES-200 passages, testing nuanced comprehension while maintaining full parallelism for cross-lingual comparison. 
The AI2 Reasoning Challenge (ARC) dataset \citep{clark2018think} is a rigorously constructed benchmark for advanced question answering and reasoning evaluation. 
It comprises 7,787 genuine grade-school level science questions.
We use the multilingual ARC translated by \citet{lai2023okapi} using ChatGPT.

\paragraph{Math}

We use the Multilingual Grade School Math Benchmark (MGSM) \citep{shi2022language} that extends the GSM8K dataset \citep{cobbe2021training} to evaluate cross-lingual mathematical reasoning in large language models. Derived from GSM8K's collection of 8,500 linguistically diverse grade-school math word problems requiring multi-step reasoning, MGSM comprises 250 carefully selected problems manually translated into ten typologically diverse languages.

The evaluation benchmarks are abbreviated as follows:
\begin{itemize}[nolistsep,noitemsep]
    \item \texttt{XC}: XCOPA (Cross-lingual Choice of Plausible Alternatives)
    \item \texttt{XSC}: XStoryCloze (Cross-lingual Story Cloze Test)
    \item \texttt{BELE}: BELEBELE, Multilingual Reading Comprehension Benchmark
    \item \texttt{ARC}: Multilingual AI2 Reasoning Challenge
\end{itemize}

For the MGSM benchmark, we distinguish between:
\begin{itemize}[nolistsep,noitemsep]
    \item \texttt{Dir.}: Direct prompting (question-only)
    \item \texttt{CoT}: Chain-of-Thought prompting (with reasoning steps) \citep{wei2022chain}
\end{itemize}

\begin{table*}[t!]
\scriptsize
\centering
\caption{Evaluation statistics.
    Sample/Lang: average number of test samples per language; N Lang: number of languages covered.}
\label{tab:downstream_tasks}
\resizebox{\linewidth}{!}{%
\setlength{\tabcolsep}{3pt}
\begin{tabular}{llcrrc}
    \toprule
    \textbf{Tasks}  & \textbf{Dataset} & \textbf{Metric} & \textbf{Samples/Lang} & \textbf{N Lang} & \textbf{Domain}\\
    \midrule
     \multirow{2}{*}{Text Classification (\Cref{sec:detailed_classification})}& SIB200 \citep{sib-200} & Accuracy & 204 & 205 & Misc \\
     &Taxi1500 \citep{ma2023taxi1500} & Accuracy & 111 & 1507 & Bible \\
    \midrule 
    \multirow{2}{*}{Commonsense Reasoning (\Cref{sec:detailed_commonsense})}& XCOPA~\citep{ponti-etal-2020-xcopa} & Accuracy & 600  & 11 & Misc \\
    & XStoryCloze \citep{lin2022few} & Accuracy & 1870 & 11 & Misc \\
     \midrule
    \multirow{1}{*}{Natural Language Inference (\Cref{sec:detailed_nli})}& XNLI \citep{conneau2018xnli} & Accuracy & 2490 & 15 & Misc \\
     \midrule
    \multirow{1}{*}{Machine Translation (\Cref{sec:detailed_mt})}& FLORES-200 \citep{costa2022no} & BLEU, chrF++ & 1012 & 204 & Misc \\
     \midrule
    \multirow{3}{*}{Summarization (\Cref{sec:detailed_summarization})}& XL-Sum \citep{hasan-etal-2021-xl} & ROUGE-L, BERTScore & 2537 & 44 & News \\
            & MassiveSumm Long \citep{varab-schluter-2021-massivesumm} & ROUGE-L, BERTScore & 3908 & 55 & News\\
            & MassiveSumm Short \citep{varab-schluter-2021-massivesumm} & ROUGE-L, BERTScore & 5538 & 88 & News\\
    \midrule
    \multirow{2}{*}{Machine Comprehension (\Cref{sec:detailed_mrc})}& BELEBELE \citep{bandarkar2023belebele} & Accuracy & 900 & 122 & Misc \\
      & ARC multilingual \citep{lai2023okapi} & Accuracy & 1170 & 31 & Misc \\
    \midrule
    \multirow{2}{*}{Math (\Cref{sec:detailed_math})}& MGSM direct \citep{shi2022language} & Accuracy & 250 & 10 & Misc \\
	& MGSM CoT \citep{shi2022language} & Accuracy & 250 & 10 & Misc \\
    \bottomrule
    \end{tabular}
}
\end{table*}

\subsection{Evaluation Software}
\label{sec:evaluation_software}

We evaluate benchmarks using the \texttt{lm-evaluation-harness} framework~\citep{eval-harness} for supported tasks, and custom or open-source scripts for others. 
For text classification (e.g., SIB-200, Taxi-1500), we predict the highest-probability category using next-token probabilities via Transformers~\citep{wolf2019huggingface}. 
For translation and open-ended generation, we use vLLM~\citep{kwon2023efficient} for faster inference.

\subsection{Baselines}
\label{sec:baselines}

We compare our models with several groups of baseline approaches, covering both established and emerging LLMs.

\paragraph{Llama 2 series including CPT models:}
\begin{itemize}[nolistsep,noitemsep]
    \item Llama 2 base \& chat \citep{touvron2023llama2}: the second generation of Meta's foundational 7BB parameter models release in 2023;
    \item CodeLlama 2 \citep{roziere2023code}: continual pre-trained model specialized for code;
    \item LlaMAX Llama 2 base and Alpaca \citep{lu2024llamax}: continual pre-trained and instruction-tuned variants on Alpaca \citep{alpaca};
    \item MaLA-500 Llama 2 v1 \& v2 \citep{lin2024mala}: multilingual adaptations with continual pre-training and low-rank adaptation \citep{hu2022lora};
    \item YaYi Llama 2\furl{https://huggingface.co/wenge-research/yayi-7b-llama2}: a Chinese-optimized version;
    \item TowerBase and TowerInstruct Llama 2 \citep{alves2024tower}: domain-specific adaptations optimized mainly for machine translation;
    \item EMMA-500 Llama 2 \citep{ji2024emma}: a massively multilingual variant.
\end{itemize}

\paragraph{Multilingual LLMs including recent advances:}
\begin{itemize}[nolistsep,noitemsep]
    \item Occiglot Mistral v0.1 base and instruct \furl{https://huggingface.co/occiglot}: multilingual variants continue-trained and instruct-tuned on Mistral \citep{jiang2023mistral7b};
    \item BLOOM \& BLOOMZ \citep{scao2022bloom,muennighoff2022crosslingual}: early open multilingual and instruction-tuned models released in 2022;
    \item YaYi \furl{https://huggingface.co/wenge-research/yayi-7b}: a model with Chinese-focused multilingual capabilities;
    \item Aya 23 \citep{aryabumi2024aya23}: a multilingual instruction-tuned LLM supporting 23 languages;
    \item Aya Expanse \citep{dang2024aya}: a multilingual variant optimized with various post-training methods such as synthetic data augmentation, iterative preference training, and model merging;
    \item Gemma series: Google's multilingual models including generations 1\furl{https://huggingface.co/google/gemma-7b} \& 2 \citep{gemmateam2024gemma2improvingopen};
    \item Qwen series \citep{bai2023qwen}: Alibaba's multilingual offerings including generations 1.5\furl{https://huggingface.co/Qwen/Qwen1.5-7B}, 2 \citep{yang2024qwen2technicalreport}, \& 2.5 \citep{qwen2025qwen25technicalreport};
    \item Macro-LLM \citep{ming2024marcollmbridginglanguagesmassive}: a continue-pretrained LLM using Qwen2.
\end{itemize}

\paragraph{Llama 3 series including CPT models:}
\begin{itemize}[nolistsep,noitemsep]
    \item Llama 3 \citep{dubey2024llama3}: Meta's latest generation with a cutoff or March 2023;
    \item Llama 3.1 \citep{dubey2024llama3}: Refined version with improved capabilities, which extends the data cutoff to December 2023;
    \item LLaMAX Llama 3 base and Alpaca \citep{lu2024llamax}: continue-trained and instruction-tuned variants.
\end{itemize}

Our evaluation encompasses these representative models to ensure a comprehensive comparison. 
The selection covers diverse architectures and linguistic capabilities. 
We continue to expand our benchmarking coverage and welcome collaboration inquiries regarding additional model evaluations.

\section{Detailed Results}
\label{sec:detailed_results}

This section presents the detailed results of each evaluated task, including average scores categorized by language resource groups and per-language scores if the benchmark has a few languages. 
For massively multilingual benchmarks with more than a dozen of languages, we provide the results on our project website (\website).
We also provide all the model outputs of different generation tasks on the project website.

\subsection{Text Classification}
\label{sec:detailed_classification}

We evaluate with 3-shot prompting on the SIB-200 and Taxi-1500 text classification datasets.
The prompt template for SIB-200 is as follows:
\begin{lstlisting}
Topic Classification: science/technology, travel, politics, sports, health, entertainment, geography.
{examples}
The topic of the news "${text}" is
\end{lstlisting}

\noindent For Taxi-1500, the prompt template is as follows:
\begin{lstlisting}
Topic Classification: Recommendation, Faith, Description, Sin, Grace, Violence.
{examples}
The topic of the verse "${text}" is
\end{lstlisting}

As for SIB-200 task, Table~\ref{tab:sib200_taxi1500} shows that \texttt{\EMMA Llama 3.1 8B Bi} attains the highest accuracy among \EMMA model series (62.07\%), edging its backbone by +0.65\% and crossing the 60\% threshold together with \texttt{\EMMA Llama 3 8B Mono} (60.62\%).  Yet the other two variants fall significantly below their bases: \EMMA Llama 3 8B Bi drops to 39.40\%, while \EMMA Llama 3.1 8B Mono slips to 26.16\%.  The mixed picture---improvements for some settings, regressions for others---suggests that our current continual pre-training strategy for \EMMA models may not be consistently helpful for SIB-200 task, and calls for deeper analysis in terms of data mix and training strategy.

With its religious text domain, short inputs and 1500+ languages, Taxi-1500 is markedly harder. Here, the advantages of \EMMA are more pronounced. Where most 8 B models struggle to reach 20\% accuracy, all four \EMMA variants hover around or above that mark.  \EMMA Llama-3 8B-Bi achieves 25.13\%, a gain of 3.40\% over the base model, and \EMMA Llama-3.1 8B-Bi reaches 24.87\%.

\begin{table*}[ht]
\caption{3-shot results on SIB-200 and Taxi-1500 (Accuracy \%).}
\label{tab:sib200_taxi1500}
\setlength{\tabcolsep}{2pt}
\scriptsize
\centering
\begin{tabular}{l|rrrrrr|rrrrrr}
\toprule
\multicolumn{1}{l}{\multirow{2}{*}{\textbf{Model}}} & \multicolumn{6}{|c}{\textbf{SIB-200}}                                                                          & \multicolumn{6}{|c}{\textbf{Taxi-1500}}                                                                         \\ 
\cmidrule(lr){2-7}\cmidrule(lr){8-13}
                                                   & \textbf{Avg} & \textbf{High} & \textbf{Med-High} & \textbf{Medium} & \textbf{Med-Low} & \textbf{Low} 
                                                   & \textbf{Avg} & \textbf{High} & \textbf{Med-High} & \textbf{Medium} & \textbf{Med-Low} & \textbf{Low} \\
\midrule
                    Llama 2 7B & 22.41 & 25.98 &     22.17 &   22.43 &    23.36 & 22.74 & 17.54 & 19.57 &     17.78 &   17.89 &    18.17 & 18.04 \\
               Llama 2 7B Chat & 25.58 & 29.15 &     25.27 &   25.43 &    26.58 & 26.02 & 15.44 & 18.78 &     15.67 &   15.77 &    16.09 & 16.01 \\
                  CodeLlama 7B & 23.35 & 25.80 &     23.01 &   23.39 &    24.27 & 23.84 & 17.03 & 17.47 &     17.11 &   17.12 &    17.17 & 17.15 \\
             LLaMAX Llama 2 7B & 10.61 & 12.83 &     10.96 &   10.66 &    11.11 & 10.78 & 23.52 & 23.08 &     23.54 &   23.55 &    23.52 & 23.53 \\
      LLaMAX Llama 2 7B Alpaca & 27.89 & 33.36 &     27.69 &   27.88 &    29.74 & 28.66 & 15.09 & 18.87 &     15.31 &   15.43 &    15.70 & 15.57 \\
               MaLA-500 10B V1 & 23.25 & 23.00 &     23.44 &   23.03 &    23.30 & 23.13 & 25.27 & 23.53 &     25.08 &   25.00 &    24.83 & 24.76 \\
               MaLA-500 10B V2 & 19.30 & 17.83 &     18.57 &   18.42 &    19.51 & 19.04 & 23.39 & 21.27 &     23.35 &   23.07 &    22.94 & 22.72 \\
               Yayi Llama 2 7B & 24.57 & 27.99 &     24.08 &   24.64 &    25.72 & 25.03 & 17.73 & 18.62 &     17.84 &   17.90 &    18.01 & 17.98 \\
    Tower Base Llama 2 7B V0.1 & 19.34 & 21.30 &     18.90 &   19.07 &    19.86 & 19.53 & 17.73 & 18.74 &     17.76 &   17.87 &    18.04 & 18.06 \\
Tower Instruct Llama 2 7B V0.2 & 20.53 & 22.91 &     20.22 &   20.46 &    21.10 & 20.81 & 17.29 & 20.16 &     17.69 &   17.72 &    18.05 & 17.92 \\
           EMMA-500 Llama 2 7B & 31.27 & 33.02 &     30.34 &   30.86 &    31.93 & 31.97 & 19.82 & 24.32 &     21.07 &   21.32 &    21.68 & 21.73 \\
\hline
           Occiglot Mistral 7B & 32.69 & 37.77 &     32.91 &   32.87 &    34.07 & 33.27 & 22.26 & 24.74 &     22.53 &   22.61 &    22.69 & 22.62 \\
  Occiglot Mistral 7B Instruct & 34.31 & 37.79 &     34.01 &   34.20 &    35.37 & 34.87 & 18.76 & 24.59 &     19.41 &   19.42 &    19.68 & 19.59 \\
                     BLOOM 7B1 & 17.82 & 21.57 &     19.54 &   18.61 &    18.13 & 17.73 & 14.76 & 15.71 &     14.98 &   14.93 &    14.91 & 14.93 \\
                    BLOOMZ 7B1 & 29.73 & 29.77 &     29.95 &   30.13 &    29.89 & 29.99 & 16.96 & 16.89 &     16.99 &   16.98 &    17.00 & 17.00 \\
                       Yayi 7B & 35.76 & 39.51 &     36.50 &   36.62 &    36.40 & 36.40 & 16.12 & 16.56 &     16.17 &   16.24 &    16.26 & 16.19 \\
                     Aya 23 8B & 41.50 & 45.45 &     42.17 &   42.43 &    43.42 & 42.20 & 22.64 & 22.82 &     22.63 &   22.56 &    22.56 & 22.52 \\
                Aya Expanse 8B & 57.01 & 63.82 &     59.73 &   58.99 &    59.84 & 58.06 & 18.73 & 19.56 &     19.07 &   19.08 &    19.03 & 19.08 \\
                      Gemma 7B & 58.21 & 68.76 &     60.00 &   60.14 &    62.08 & 59.97 & 13.83 & 25.10 &     14.78 &   15.42 &    16.92 & 16.55 \\
                    Gemma 2 9B & 46.25 & 51.94 &     45.79 &   46.78 &    47.82 & 47.19 & 18.05 & 29.88 &     19.43 &   20.27 &    21.72 & 21.48 \\
                   Qwen 1.5 7B & 47.95 & 55.88 &     49.42 &   49.53 &    50.34 & 48.94 &  7.29 & 12.42 &      7.61 &    7.75 &     8.39 &  8.18 \\
                     Qwen 2 7B & 54.95 & 66.84 &     58.04 &   57.27 &    58.33 & 56.06 & 21.87 & 27.85 &     22.42 &   22.66 &    23.16 & 23.01 \\
                   Qwen 2.5 7B & 53.89 & 64.57 &     56.45 &   56.08 &    56.82 & 54.85 & 17.87 & 18.12 &     18.01 &   17.95 &    17.85 & 17.82 \\
              Macro-LLM GLO 7B & 64.15 & 72.73 &     65.98 &   65.73 &    66.96 & 65.07 & 21.99 & 21.31 &     21.84 &   21.92 &    21.93 & 21.97 \\
\hline
                    Llama 3 8B & 63.70 & 73.89 &     66.11 &   66.33 &    67.89 & 65.53 & 21.73 & 32.43 &     22.64 &   23.07 &    23.79 & 23.61 \\
                  Llama 3.1 8B & 61.42 & 70.83 &     63.42 &   63.67 &    65.33 & 63.17 & 20.20 & 28.22 &     21.12 &   21.59 &    22.22 & 22.02 \\
             LLaMAX Llama 3 8B & 48.60 & 51.41 &     48.19 &   48.94 &    50.16 & 49.70 & 23.01 & 22.32 &     22.84 &   22.85 &    22.82 & 22.91 \\
      LLaMAX Llama 3 8B Alpaca & 58.97 & 70.50 &     60.57 &   60.96 &    63.71 & 60.91 & 17.71 & 30.56 &     18.56 &   19.32 &    20.83 & 20.48 \\
\hline
      \EMMA Llama 3 8B Mono & 60.62 & 60.72 &     58.70 &   60.26 &    62.10 & 62.34 & 22.32 & 23.13 &     22.04 &   22.02 &    22.10 & 22.16 \\
        \EMMA Llama 3 8B Bi & 39.40 & 37.14 &     39.83 &   38.88 &    38.39 & 39.65 & 25.13 & 29.53 &     26.13 &   26.45 &    26.71 & 26.72 \\
    \EMMA Llama 3.1 8B Mono & 26.16 & 25.31 &     25.74 &   25.84 &    25.80 & 26.17 & 19.71 & 18.82 &     19.41 &   19.45 &    19.47 & 19.62 \\
      \EMMA Llama 3.1 8B Bi & 62.07 & 59.80 &     60.15 &   61.60 &    63.31 & 63.33 & 24.87 & 25.71 &     25.50 &   25.45 &    25.40 & 25.42 \\
\bottomrule
\end{tabular}	
\end{table*}

\subsection{Commonsense Reasoning}
\label{sec:detailed_commonsense}

In the commonsense reasoning evaluation, all evaluations are zero-shot; accuracy is the evaluation metric. Languages are bucketed into groups according to language availability and possible corpus size, following \citet{ji2024emma}.

Across XCOPA and XStoryCloze evaluation results (Table \ref{tab:commonsense_reasoning}), the four \EMMA variants built on Llama-3/3.1 8 B clearly lift commonsense-reasoning accuracy over their backbones. 
The best model, \texttt{\EMMA Llama 3.1 8B Bi}, averages 67.25\% on XCOPA and 68.47\% on XStoryCloze—roughly +5\% over vanilla Llama-3.1 and almost ahead of, all other similar size models. The performance gains are consistent across all different languages (\Cref{tab:xcopa_all,tab:xstorycloze_all}). 
Besides, improvements concentrate in medium- and low-resource languages: \EMMA lifts Llama-3/3.1 by roughly +7\% in medium-resource languages and flattens the usual low-resource performance drop on XStoryCloze ( 64\% vs. 53\% for the base), while high-resource accuracy remains competitive at around 70\%. 

One more interesting finding is that \EMMA monolingual variants are slightly behind bilingual variants by about 1\%, confirming that balanced bilingual continual training adds marginal yet consistent performance gains.

\begin{table*}[ht!]
\caption{0-shot results (Accuracy \%) on commonsense reasoning: XCOPA and XStoryCloze.}
\label{tab:commonsense_reasoning}
\setlength{\tabcolsep}{2pt}
\scriptsize
\centering
\begin{tabular}{l|rrrr|rrrr}
\toprule
\multirow{2}{*}{\textbf{Model}}            & \multicolumn{4}{c|}{\textbf{XCOPA}}      & \multicolumn{4}{c}{\textbf{XStoryCloze}} \\
\cmidrule(lr){2-5}\cmidrule(lr){6-9}
                                  & Avg   & High  & Medium & Low   & Avg    & High  & Medium & Low   \\
\midrule
Llama 2 7B                        & 56.67 & 62.10 & 53.90  & 51.60 & 57.55  & 63.38 & 54.45  & 49.21 \\
Llama 2 7B Chat                   & 55.85 & 61.25 & 53.13  & 50.60 & 58.41  & 64.80 & 54.77  & 49.74 \\
CodeLlama 2 7B                    & 54.69 & 58.70 & 52.53  & 51.60 & 55.68  & 60.68 & 52.33  & 49.90 \\
LLaMAX Llama 2 7B                 & 54.38 & 55.50 & 54.13  & 51.40 & 60.36  & 64.34 & 58.82  & 53.47 \\
LLaMAX Llama 2 7B Alpaca          & 56.60 & 59.80 & 55.17  & 52.40 & 63.83  & 69.08 & 62.01  & 54.33 \\
MaLA-500 Llama 2 10B v1           & 53.09 & 53.55 & 53.27  & 50.20 & 53.07  & 58.15 & 49.22  & 48.08 \\
MaLA-500 Llama 2 10B v2           & 53.09 & 53.55 & 53.27  & 50.20 & 53.07  & 58.15 & 49.22  & 48.08 \\
YaYi Llama 2 7B                   & 56.71 & 62.10 & 54.13  & 50.60 & 58.42  & 64.98 & 54.91  & 49.04 \\
TowerBase Llama 2 7B              & 56.33 & 62.50 & 52.90  & 52.20 & 57.78  & 64.35 & 53.67  & 49.57 \\
TowerInstruct Llama 2 7B          & 57.05 & 62.90 & 54.00  & 52.00 & 59.24  & 66.83 & 54.53  & 49.70 \\
EMMA-500 Llama 2 7B               & 63.11 & 66.60 & 62.57  & 52.40 & 66.38  & 68.92 & 65.73  & 61.32 \\
\hline
Occiglot Mistral 7B v0.1          & 56.67 & 62.80 & 53.37  & 52.00 & 58.10  & 65.18 & 53.28  & 50.03 \\
Occiglot Mistral 7B v0.1 Instruct & 56.55 & 62.85 & 52.97  & 52.80 & 59.39  & 66.94 & 54.33  & 50.63 \\
BLOOM 7B                          & 56.89 & 59.95 & 55.87  & 50.80 & 59.30  & 61.99 & 59.05  & 53.08 \\
BLOOMZ 7B                         & 54.87 & 56.35 & 54.60  & 50.60 & 57.12  & 61.14 & 55.82  & 49.67 \\
YaYi 7B                           & 56.64 & 59.55 & 55.50  & 51.80 & 60.67  & 64.90 & 59.40  & 52.65 \\
Aya 23 8B                         & 55.13 & 59.05 & 53.30  & 50.40 & 60.93  & 67.27 & 58.82  & 49.34 \\
Aya Expanse 8B                    & 56.38 & 61.50 & 53.77  & 51.60 & 64.80  & 73.10 & 61.93  & 49.80 \\
Gemma 7B                          & 63.64 & 70.35 & 61.43  & 50.00 & 65.01  & 69.46 & 64.49  & 54.93 \\
Gemma 2 9B                        & 66.33 & 73.40 & 64.27  & 50.40 & 67.67  & 72.47 & 66.69  & 57.64 \\
Qwen 1.5 7B                       & 59.44 & 66.85 & 55.90  & 51.00 & 59.85  & 66.62 & 56.04  & 50.56 \\
Qwen 2 7B                         & 60.31 & 68.65 & 56.40  & 50.40 & 61.46  & 69.45 & 56.97  & 50.50 \\
Qwen 2.5 7B                       & 61.71 & 72.30 & 56.63  & 49.80 & 62.06  & 69.41 & 58.36  & 51.09 \\
Marco-LLM GLO 7B                  & 62.45 & 72.20 & 57.77  & 51.60 & 63.87  & 71.26 & 61.23  & 50.66 \\
\hline
Llama 3 8B                        & 61.71 & 68.35 & 59.03  & 51.20 & 63.41  & 68.50 & 62.03  & 53.44 \\
Llama 3.1 8B                      & 61.71 & 69.30 & 58.80  & 48.80 & 63.58  & 68.66 & 62.09  & 53.87 \\
LLaMAX Llama 3 8B                 & 63.04 & 67.25 & 62.30  & 50.60 & 64.31  & 67.78 & 63.48  & 57.28 \\
LLaMAX Llama 3 8B Alpaca          & 64.36 & 69.55 & 62.97  & 52.00 & 68.27  & 72.92 & 67.27  & 58.64 \\
\hline
\EMMA Llama 3 8B Mono          & 66.20 & 69.25 & 66.20  & 54.00 & 67.36  & 69.25 & 66.83  & 63.70 \\
\EMMA Llama 3 8B Bi            & 66.82 & 70.80 & 66.67  & 51.80 & 68.35  & 70.22 & 67.79  & 64.82 \\
\EMMA Llama 3.1 8B Mono        & 65.38 & 68.85 & 65.13  & 53.00 & 67.64  & 69.64 & 67.19  & 63.57 \\
\EMMA Llama 3.1 8B Bi          & 67.25 & 71.75 & 66.83  & 51.80 & 68.47  & 70.34 & 68.13  & 64.49 \\
\bottomrule
\end{tabular}
\end{table*}

\begin{table*}[ht!]
\caption{0-shot results (ACC \%) on XCOPA in all languages}
\label{tab:xcopa_all}
\setlength{\tabcolsep}{2pt}
\resizebox{\linewidth}{!}{%
\centering
\begin{tabular}{lrrrrrrrrrrrrrrrrrrrrrrr}
\toprule
                            \textbf{Model} &    \textbf{Avg} &  \textbf{et-acc} &  \textbf{stderr} &  \textbf{ht-acc} &  \textbf{stderr} &  \textbf{id-acc} &  \textbf{stderr} &  \textbf{it-acc} &  \textbf{stderr} &  \textbf{qu-acc} &  \textbf{stderr} &  \textbf{sw-acc} &  \textbf{stderr} &  \textbf{ta-acc} &  \textbf{stderr} &  \textbf{th-acc} &  \textbf{stderr} &  \textbf{tr-acc} &  \textbf{stderr} &  \textbf{vi-acc} &  \textbf{stderr} &  \textbf{zh-acc} &  \textbf{stderr} \\
\midrule
                       Llama 2 7B & 56.67 &   48.60 &    2.24 &   50.60 &    2.24 &   62.40 &    2.17 &   65.80 &    2.12 &   51.60 &    2.24 &   52.20 &    2.24 &   53.40 &    2.23 &   56.20 &    2.22 &   54.80 &    2.23 &   62.80 &    2.16 &   65.00 &    2.14 \\
                  Llama 2 7B Chat & 55.85 &   47.80 &    2.24 &   50.80 &    2.24 &   62.40 &    2.17 &   67.20 &    2.10 &   50.60 &    2.24 &   52.20 &    2.24 &   50.60 &    2.24 &   55.00 &    2.23 &   55.20 &    2.23 &   61.20 &    2.18 &   61.40 &    2.18 \\
                   CodeLlama 2 7B & 54.69 &   46.80 &    2.23 &   51.80 &    2.24 &   57.40 &    2.21 &   63.00 &    2.16 &   51.60 &    2.24 &   48.80 &    2.24 &   55.00 &    2.23 &   55.40 &    2.23 &   53.80 &    2.23 &   55.80 &    2.22 &   62.20 &    2.17 \\
                LLaMAX Llama 2 7B & 54.38 &   49.20 &    2.24 &   52.60 &    2.24 &   53.80 &    2.23 &   52.60 &    2.24 &   51.40 &    2.24 &   54.00 &    2.23 &   58.00 &    2.21 &   57.20 &    2.21 &   53.00 &    2.23 &   53.00 &    2.23 &   63.40 &    2.16 \\
         LLaMAX Llama 2 7B Alpaca & 56.60 &   51.20 &    2.24 &   54.20 &    2.23 &   57.20 &    2.21 &   61.00 &    2.18 &   52.40 &    2.24 &   55.00 &    2.23 &   57.00 &    2.22 &   56.40 &    2.22 &   55.20 &    2.23 &   55.20 &    2.23 &   67.80 &    2.09 \\
          MaLA-500 Llama 2 10B v1 & 53.09 &   48.60 &    2.24 &   53.40 &    2.23 &   53.00 &    2.23 &   59.40 &    2.20 &   50.20 &    2.24 &   52.80 &    2.23 &   57.60 &    2.21 &   54.20 &    2.23 &   51.60 &    2.24 &   52.40 &    2.24 &   50.80 &    2.24 \\
          MaLA-500 Llama 2 10B v2 & 53.09 &   48.60 &    2.24 &   53.40 &    2.23 &   53.00 &    2.23 &   59.40 &    2.20 &   50.20 &    2.24 &   52.80 &    2.23 &   57.60 &    2.21 &   54.20 &    2.23 &   51.60 &    2.24 &   52.40 &    2.24 &   50.80 &    2.24 \\
                  YaYi Llama 2 7B & 56.71 &   48.80 &    2.24 &   50.80 &    2.24 &   62.60 &    2.17 &   67.00 &    2.10 &   50.60 &    2.24 &   53.20 &    2.23 &   55.20 &    2.23 &   54.20 &    2.23 &   55.40 &    2.23 &   63.20 &    2.16 &   62.80 &    2.16 \\
             TowerBase Llama 2 7B & 56.33 &   46.00 &    2.23 &   50.20 &    2.24 &   60.20 &    2.19 &   70.80 &    2.04 &   52.20 &    2.24 &   50.60 &    2.24 &   54.40 &    2.23 &   56.00 &    2.22 &   53.80 &    2.23 &   59.20 &    2.20 &   66.20 &    2.12 \\
         TowerInstruct Llama 2 7B & 57.05 &   48.80 &    2.24 &   51.60 &    2.24 &   62.00 &    2.17 &   71.00 &    2.03 &   52.00 &    2.24 &   51.00 &    2.24 &   54.20 &    2.23 &   56.40 &    2.22 &   54.60 &    2.23 &   58.60 &    2.20 &   67.40 &    2.10 \\
              EMMA-500 Llama 2 7B & 63.11 &   61.40 &    2.18 &   58.00 &    2.21 &   74.20 &    1.96 &   69.40 &    2.06 &   52.40 &    2.24 &   66.20 &    2.12 &   60.00 &    2.19 &   55.60 &    2.22 &   62.00 &    2.17 &   70.20 &    2.05 &   64.80 &    2.14 \\
\hline
         Occiglot Mistral 7B v0.1 & 56.67 &   47.20 &    2.23 &   51.40 &    2.24 &   57.00 &    2.22 &   74.60 &    1.95 &   52.00 &    2.24 &   51.60 &    2.24 &   57.20 &    2.21 &   55.80 &    2.22 &   54.40 &    2.23 &   55.20 &    2.23 &   67.00 &    2.10 \\
Occiglot Mistral 7B v0.1 Instruct & 56.55 &   46.80 &    2.23 &   51.00 &    2.24 &   58.40 &    2.21 &   73.80 &    1.97 &   52.80 &    2.23 &   50.00 &    2.24 &   56.60 &    2.22 &   55.00 &    2.23 &   56.20 &    2.22 &   55.80 &    2.22 &   65.60 &    2.13 \\
                         BLOOM 7B & 56.89 &   48.20 &    2.24 &   50.80 &    2.24 &   69.80 &    2.06 &   52.80 &    2.23 &   50.80 &    2.24 &   51.80 &    2.24 &   59.20 &    2.20 &   55.40 &    2.23 &   51.00 &    2.24 &   70.80 &    2.04 &   65.20 &    2.13 \\
                        BLOOMZ 7B & 54.87 &   49.20 &    2.24 &   54.00 &    2.23 &   60.60 &    2.19 &   51.40 &    2.24 &   50.60 &    2.24 &   53.40 &    2.23 &   57.40 &    2.21 &   53.00 &    2.23 &   52.20 &    2.24 &   59.80 &    2.19 &   62.00 &    2.17 \\
                          YaYi 7B & 56.64 &   50.40 &    2.24 &   53.00 &    2.23 &   63.40 &    2.16 &   51.80 &    2.24 &   51.80 &    2.24 &   55.40 &    2.23 &   56.20 &    2.22 &   54.60 &    2.23 &   52.00 &    2.24 &   66.40 &    2.11 &   68.00 &    2.09 \\
                        Aya 23 8B & 55.13 &   50.20 &    2.24 &   52.20 &    2.24 &   57.00 &    2.22 &   60.40 &    2.19 &   50.40 &    2.24 &   52.20 &    2.24 &   55.60 &    2.22 &   52.60 &    2.24 &   59.60 &    2.20 &   58.80 &    2.20 &   57.40 &    2.21 \\
                   Aya Expanse 8B & 56.38 &   50.80 &    2.24 &   51.60 &    2.24 &   61.20 &    2.18 &   63.40 &    2.16 &   51.60 &    2.24 &   53.60 &    2.23 &   55.20 &    2.23 &   50.20 &    2.24 &   57.80 &    2.21 &   60.20 &    2.19 &   64.60 &    2.14 \\
\hline
                         Gemma 7B & 63.64 &   59.20 &    2.20 &   54.80 &    2.23 &   72.00 &    2.01 &   72.80 &    1.99 &   50.00 &    2.24 &   60.60 &    2.19 &   61.60 &    2.18 &   60.40 &    2.19 &   65.60 &    2.13 &   74.20 &    1.96 &   68.80 &    2.07 \\
                       Gemma 2 9B & 66.33 &   63.80 &    2.15 &   52.80 &    2.23 &   77.80 &    1.86 &   75.80 &    1.92 &   50.40 &    2.24 &   63.80 &    2.15 &   63.60 &    2.15 &   63.80 &    2.15 &   67.40 &    2.10 &   76.60 &    1.90 &   73.80 &    1.97 \\
                      Qwen 1.5 7B & 59.44 &   52.00 &    2.24 &   53.00 &    2.23 &   64.40 &    2.14 &   65.20 &    2.13 &   51.00 &    2.24 &   52.60 &    2.24 &   55.80 &    2.22 &   57.60 &    2.21 &   58.80 &    2.20 &   69.20 &    2.07 &   74.20 &    1.96 \\
                        Qwen 2 7B & 60.31 &   50.80 &    2.24 &   50.80 &    2.24 &   70.60 &    2.04 &   71.40 &    2.02 &   50.40 &    2.24 &   52.40 &    2.24 &   52.80 &    2.23 &   61.00 &    2.18 &   57.20 &    2.21 &   69.40 &    2.06 &   76.60 &    1.90 \\
                      Qwen 2.5 7B & 61.71 &   49.40 &    2.24 &   52.80 &    2.23 &   72.60 &    2.00 &   74.60 &    1.95 &   49.80 &    2.24 &   52.20 &    2.24 &   54.60 &    2.23 &   58.20 &    2.21 &   60.00 &    2.19 &   75.20 &    1.93 &   79.40 &    1.81 \\
                 Marco-LLM GLO 7B & 62.45 &   49.60 &    2.24 &   52.40 &    2.24 &   74.20 &    1.96 &   73.40 &    1.98 &   51.60 &    2.24 &   54.40 &    2.23 &   57.20 &    2.21 &   58.80 &    2.20 &   66.40 &    2.11 &   73.80 &    1.97 &   75.20 &    1.93 \\
\hline
                       Llama 3 8B & 61.71 &   53.40 &    2.23 &   52.80 &    2.23 &   71.40 &    2.02 &   71.60 &    2.02 &   51.20 &    2.24 &   58.00 &    2.21 &   60.00 &    2.19 &   58.60 &    2.20 &   62.40 &    2.17 &   71.60 &    2.02 &   67.80 &    2.09 \\
                     Llama 3.1 8B & 61.71 &   53.00 &    2.23 &   53.80 &    2.23 &   71.60 &    2.02 &   72.60 &    2.00 &   48.80 &    2.24 &   55.40 &    2.23 &   61.20 &    2.18 &   57.80 &    2.21 &   61.80 &    2.18 &   72.40 &    2.00 &   70.40 &    2.04 \\
                LLaMAX Llama 3 8B & 63.04 &   61.80 &    2.18 &   56.40 &    2.22 &   73.00 &    1.99 &   68.40 &    2.08 &   50.60 &    2.24 &   62.60 &    2.17 &   60.80 &    2.19 &   59.20 &    2.20 &   61.40 &    2.18 &   70.40 &    2.04 &   68.80 &    2.07 \\
         LLaMAX Llama 3 8B Alpaca & 64.36 &   63.00 &    2.16 &   54.00 &    2.23 &   73.00 &    1.99 &   71.80 &    2.01 &   52.00 &    2.24 &   63.40 &    2.16 &   63.20 &    2.16 &   61.20 &    2.18 &   64.00 &    2.15 &   71.80 &    2.01 &   70.60 &    2.04 \\
\hline
         \EMMA Llama 3 8B Mono & 66.20 &   70.80 &    2.04 &   61.80 &    2.18 &   75.00 &    1.94 &   73.40 &    1.98 &   54.00 &    2.23 &   67.60 &    2.10 &   61.80 &    2.18 &   60.20 &    2.19 &   64.00 &    2.15 &   73.00 &    1.99 &   66.60 &    2.11 \\
           \EMMA Llama 3 8B Bi & 66.82 &   71.40 &    2.02 &   60.80 &    2.19 &   78.00 &    1.85 &   73.40 &    1.98 &   51.80 &    2.24 &   68.20 &    2.08 &   63.40 &    2.16 &   58.20 &    2.21 &   66.60 &    2.11 &   73.20 &    1.98 &   70.00 &    2.05 \\
       \EMMA Llama 3.1 8B Mono & 65.38 &   65.20 &    2.13 &   59.00 &    2.20 &   76.00 &    1.91 &   73.00 &    1.99 &   53.00 &    2.23 &   68.40 &    2.08 &   62.40 &    2.17 &   59.80 &    2.19 &   62.80 &    2.16 &   74.60 &    1.95 &   65.00 &    2.14 \\
         \EMMA Llama 3.1 8B Bi & 67.25 &   70.40 &    2.04 &   61.00 &    2.18 &   77.00 &    1.88 &   74.60 &    1.95 &   51.80 &    2.24 &   69.60 &    2.06 &   63.20 &    2.16 &   59.80 &    2.19 &   67.20 &    2.10 &   74.60 &    1.95 &   70.60 &    2.04 \\
\bottomrule
\end{tabular}
}
\end{table*}

\begin{table*}[ht!]
\setlength{\tabcolsep}{3pt}
\centering
\caption{0-shot results (Accuracy \%) on XStoryCloze in all languages}
\label{tab:xstorycloze_all}
\setlength{\tabcolsep}{2pt}
\resizebox{\linewidth}{!}{%
\begin{tabular}{lrrrrrrrrrrrrrrrrrrrrrrr}
\toprule
\textbf{Model} & \textbf{Avg} & \textbf{ar-acc} & \textbf{stderr} & \textbf{en-acc} & \textbf{stderr} & \textbf{es-acc} & \textbf{stderr} & \textbf{eu-acc} & \textbf{stderr} & \textbf{hi-acc} & \textbf{stderr} & \textbf{id-acc} & \textbf{stderr} & \textbf{my-acc} & \textbf{stderr} & \textbf{ru-acc} & \textbf{stderr} & \textbf{sw-acc} & \textbf{stderr} & \textbf{te-acc} & \textbf{stderr} & \textbf{zh-acc} & \textbf{stderr} \\
\midrule
                       Llama 2 7B & 57.55 &   49.90 &    1.29 &   77.04 &    1.08 &   67.37 &    1.21 &   50.36 &    1.29 &   53.74 &    1.28 &   59.23 &    1.26 &   48.05 &    1.29 &   63.00 &    1.24 &   50.50 &    1.29 &   54.33 &    1.28 &   59.56 &    1.26 \\
                  Llama 2 7B Chat & 58.41 &   50.50 &    1.29 &   78.69 &    1.05 &   67.11 &    1.21 &   50.83 &    1.29 &   54.07 &    1.28 &   59.63 &    1.26 &   48.64 &    1.29 &   65.52 &    1.22 &   52.02 &    1.29 &   53.34 &    1.28 &   62.21 &    1.25 \\
                   CodeLlama 2 7B & 55.68 &   50.10 &    1.29 &   71.48 &    1.16 &   63.40 &    1.24 &   50.43 &    1.29 &   49.70 &    1.29 &   55.86 &    1.28 &   49.37 &    1.29 &   59.23 &    1.26 &   50.03 &    1.29 &   53.74 &    1.28 &   59.17 &    1.26 \\
                LLaMAX Llama 2 7B & 60.36 &   58.90 &    1.27 &   75.51 &    1.11 &   65.25 &    1.23 &   54.47 &    1.28 &   58.17 &    1.27 &   60.62 &    1.26 &   52.48 &    1.29 &   61.22 &    1.25 &   57.18 &    1.27 &   59.30 &    1.26 &   60.82 &    1.26 \\
         LLaMAX Llama 2 7B Alpaca & 63.83 &   60.36 &    1.26 &   81.47 &    1.00 &   70.68 &    1.17 &   54.86 &    1.28 &   62.14 &    1.25 &   66.45 &    1.22 &   53.81 &    1.28 &   67.44 &    1.21 &   60.16 &    1.26 &   59.30 &    1.26 &   65.45 &    1.22 \\
          MaLA-500 Llama 2 10B v1 & 53.07 &   48.18 &    1.29 &   73.53 &    1.14 &   62.41 &    1.25 &   49.90 &    1.29 &   47.65 &    1.29 &   47.92 &    1.29 &   46.26 &    1.28 &   54.93 &    1.28 &   48.71 &    1.29 &   52.61 &    1.28 &   51.69 &    1.29 \\
          MaLA-500 Llama 2 10B v2 & 53.07 &   48.18 &    1.29 &   73.53 &    1.14 &   62.41 &    1.25 &   49.90 &    1.29 &   47.65 &    1.29 &   47.92 &    1.29 &   46.26 &    1.28 &   54.93 &    1.28 &   48.71 &    1.29 &   52.61 &    1.28 &   51.69 &    1.29 \\
                  YaYi Llama 2 7B & 58.42 &   49.97 &    1.29 &   79.09 &    1.05 &   68.70 &    1.19 &   50.63 &    1.29 &   54.27 &    1.28 &   61.42 &    1.25 &   47.45 &    1.29 &   64.79 &    1.23 &   50.03 &    1.29 &   53.94 &    1.28 &   62.34 &    1.25 \\
             TowerBase Llama 2 7B & 57.78 &   49.17 &    1.29 &   77.23 &    1.08 &   69.82 &    1.18 &   50.76 &    1.29 &   52.88 &    1.28 &   58.31 &    1.27 &   48.38 &    1.29 &   67.04 &    1.21 &   50.36 &    1.29 &   53.14 &    1.28 &   58.50 &    1.27 \\
         TowerInstruct Llama 2 7B & 59.24 &   49.31 &    1.29 &   80.87 &    1.01 &   71.61 &    1.16 &   50.69 &    1.29 &   52.95 &    1.28 &   59.56 &    1.26 &   48.71 &    1.29 &   69.36 &    1.19 &   51.49 &    1.29 &   54.14 &    1.28 &   63.00 &    1.24 \\
              EMMA-500 Llama 2 7B & 66.38 &   66.25 &    1.22 &   76.44 &    1.09 &   70.02 &    1.18 &   64.73 &    1.23 &   64.92 &    1.23 &   68.63 &    1.19 &   57.91 &    1.27 &   68.50 &    1.20 &   64.73 &    1.23 &   64.66 &    1.23 &   63.40 &    1.24 \\
\hline
         Occiglot Mistral 7B v0.1 & 58.10 &   51.29 &    1.29 &   77.37 &    1.08 &   73.40 &    1.14 &   52.08 &    1.29 &   51.49 &    1.29 &   58.64 &    1.27 &   47.98 &    1.29 &   62.94 &    1.24 &   49.83 &    1.29 &   53.14 &    1.28 &   60.89 &    1.26 \\
Occiglot Mistral 7B v0.1 Instruct & 59.39 &   52.68 &    1.28 &   79.42 &    1.04 &   74.19 &    1.13 &   53.01 &    1.28 &   52.75 &    1.28 &   60.36 &    1.26 &   48.25 &    1.29 &   65.06 &    1.23 &   50.43 &    1.29 &   53.81 &    1.28 &   63.34 &    1.24 \\
                         BLOOM 7B & 59.30 &   58.57 &    1.27 &   70.55 &    1.17 &   66.18 &    1.22 &   57.25 &    1.27 &   60.42 &    1.26 &   64.53 &    1.23 &   48.91 &    1.29 &   52.75 &    1.28 &   53.94 &    1.28 &   57.31 &    1.27 &   61.88 &    1.25 \\
                        BLOOMZ 7B & 57.12 &   56.52 &    1.28 &   73.00 &    1.14 &   64.59 &    1.23 &   51.09 &    1.29 &   57.64 &    1.27 &   55.33 &    1.28 &   48.25 &    1.29 &   52.15 &    1.29 &   52.15 &    1.29 &   58.17 &    1.27 &   59.43 &    1.26 \\
                          YaYi 7B & 60.67 &   61.81 &    1.25 &   74.32 &    1.12 &   69.42 &    1.19 &   56.06 &    1.28 &   63.67 &    1.24 &   62.41 &    1.25 &   49.24 &    1.29 &   52.15 &    1.29 &   53.61 &    1.28 &   57.91 &    1.27 &   66.78 &    1.21 \\
                        Aya 23 8B & 60.93 &   62.61 &    1.25 &   74.59 &    1.12 &   67.31 &    1.21 &   50.96 &    1.29 &   64.26 &    1.23 &   66.64 &    1.21 &   47.72 &    1.29 &   68.70 &    1.19 &   50.36 &    1.29 &   54.00 &    1.28 &   63.14 &    1.24 \\
                   Aya Expanse 8B & 64.80 &   69.42 &    1.19 &   80.41 &    1.02 &   74.26 &    1.13 &   51.36 &    1.29 &   68.50 &    1.20 &   72.47 &    1.15 &   48.25 &    1.29 &   73.86 &    1.13 &   51.62 &    1.29 &   55.13 &    1.28 &   67.57 &    1.20 \\
                         Gemma 7B & 65.01 &   60.42 &    1.26 &   80.15 &    1.03 &   70.62 &    1.17 &   57.58 &    1.27 &   64.92 &    1.23 &   67.64 &    1.20 &   52.28 &    1.29 &   70.55 &    1.17 &   62.14 &    1.25 &   63.27 &    1.24 &   65.59 &    1.22 \\
                       Gemma 2 9B & 67.67 &   65.25 &    1.23 &   80.15 &    1.03 &   74.26 &    1.13 &   60.16 &    1.26 &   66.91 &    1.21 &   71.34 &    1.16 &   55.13 &    1.28 &   73.73 &    1.13 &   63.73 &    1.24 &   64.79 &    1.23 &   68.96 &    1.19 \\
                      Qwen 1.5 7B & 59.85 &   55.39 &    1.28 &   78.16 &    1.06 &   68.30 &    1.20 &   51.89 &    1.29 &   55.66 &    1.28 &   62.87 &    1.24 &   49.24 &    1.29 &   63.27 &    1.24 &   51.69 &    1.29 &   53.94 &    1.28 &   67.97 &    1.20 \\
                        Qwen 2 7B & 61.46 &   59.96 &    1.26 &   78.95 &    1.05 &   69.76 &    1.18 &   52.02 &    1.29 &   57.97 &    1.27 &   64.39 &    1.23 &   48.97 &    1.29 &   69.56 &    1.18 &   51.42 &    1.29 &   54.07 &    1.28 &   69.03 &    1.19 \\
                      Qwen 2.5 7B & 62.06 &   62.41 &    1.25 &   77.56 &    1.07 &   69.16 &    1.19 &   51.82 &    1.29 &   58.70 &    1.27 &   67.04 &    1.21 &   50.36 &    1.29 &   69.69 &    1.18 &   51.82 &    1.29 &   55.86 &    1.28 &   68.23 &    1.20 \\
                 Marco-LLM GLO 7B & 63.87 &   64.73 &    1.23 &   79.75 &    1.03 &   72.14 &    1.15 &   52.28 &    1.29 &   61.02 &    1.26 &   68.70 &    1.19 &   49.04 &    1.29 &   70.42 &    1.17 &   53.67 &    1.28 &   61.55 &    1.25 &   69.29 &    1.19 \\
\hline
                       Llama 3 8B & 63.41 &   58.64 &    1.27 &   78.69 &    1.05 &   70.62 &    1.17 &   55.79 &    1.28 &   62.81 &    1.24 &   65.92 &    1.22 &   51.09 &    1.29 &   68.76 &    1.19 &   56.39 &    1.28 &   63.00 &    1.24 &   65.78 &    1.22 \\
                     Llama 3.1 8B & 63.58 &   59.10 &    1.27 &   78.16 &    1.06 &   70.81 &    1.17 &   55.33 &    1.28 &   63.27 &    1.24 &   68.03 &    1.20 &   52.42 &    1.29 &   68.63 &    1.19 &   55.92 &    1.28 &   61.15 &    1.25 &   66.58 &    1.21 \\
                LLaMAX Llama 3 8B & 64.31 &   59.96 &    1.26 &   76.97 &    1.08 &   69.09 &    1.19 &   59.23 &    1.26 &   62.87 &    1.24 &   66.78 &    1.21 &   55.33 &    1.28 &   68.30 &    1.20 &   60.69 &    1.26 &   63.60 &    1.24 &   64.59 &    1.23 \\
         LLaMAX Llama 3 8B Alpaca & 68.27 &   64.73 &    1.23 &   83.19 &    0.96 &   73.00 &    1.14 &   59.96 &    1.26 &   67.50 &    1.21 &   72.40 &    1.15 &   57.31 &    1.27 &   72.80 &    1.15 &   64.00 &    1.24 &   65.19 &    1.23 &   70.88 &    1.17 \\
\hline
         \EMMA Llama 3 8B Mono & 67.36 &   64.06 &    1.23 &   76.17 &    1.10 &   69.42 &    1.19 &   66.71 &    1.21 &   66.05 &    1.22 &   68.63 &    1.19 &   60.69 &    1.26 &   69.49 &    1.18 &   65.59 &    1.22 &   67.04 &    1.21 &   67.11 &    1.21 \\
           \EMMA Llama 3 8B Bi & 68.35 &   66.38 &    1.22 &   75.78 &    1.10 &   71.28 &    1.16 &   66.64 &    1.21 &   66.91 &    1.21 &   69.56 &    1.18 &   63.00 &    1.24 &   70.95 &    1.17 &   67.37 &    1.21 &   67.31 &    1.21 &   66.71 &    1.21 \\
       \EMMA Llama 3.1 8B Mono & 67.64 &   64.20 &    1.23 &   75.71 &    1.10 &   70.62 &    1.17 &   66.18 &    1.22 &   67.17 &    1.21 &   69.62 &    1.18 &   60.95 &    1.26 &   70.62 &    1.17 &   66.05 &    1.22 &   65.92 &    1.22 &   67.04 &    1.21 \\
         \EMMA Llama 3.1 8B Bi & 68.47 &   66.18 &    1.22 &   75.12 &    1.11 &   71.34 &    1.16 &   66.45 &    1.22 &   67.77 &    1.20 &   70.68 &    1.17 &   62.54 &    1.25 &   71.41 &    1.16 &   67.50 &    1.21 &   66.58 &    1.21 &   67.64 &    1.20 \\
\bottomrule
\end{tabular}
}
\end{table*}

\subsection{Natural Language Inference}
\label{sec:detailed_nli}

According to the XNLI evaluation results shown in Table \ref{tab:XNLI}, the bilingual \EMMA 8B variant outperform its backbone on average (45.15\% vs 44.97\%) and lift low-resource accuracy by roughly +4\%, closing most of the gap to the stronger Gemma 2 9 B while matching Aya Expanse 8 B. The monolingual Llama-3 variant keeps pace (44.15\%), but the same Llama backbone models with monolingual continual training leads to a performance drop (39.98\% in \EMMA Llama 3.1 8B Mono), leaving the bilingual Llama-3.1 model (44.67\%) as the only Llama-3.1-based \EMMA choice that preserves gains in low-resource languages without a drop in high-resource languages. Overall, \EMMA offers modest NLI benefits concentrated in low-resource languages.

\begin{table}[ht!]
\caption{0-shot results on XNLI (Accuracy \%).}
\label{tab:XNLI}
\setlength{\tabcolsep}{2pt}
\scriptsize
\centering
\begin{tabular}{lrrrr}
\toprule
                            Model &   Avg &  High &  Medium &   Low \\
\midrule
                       Llama 2 7B & 40.19 & 45.26 &   37.72 & 34.97 \\
                  Llama 2 7B Chat & 38.58 & 42.77 &   36.75 & 33.87 \\
                   CodeLlama 2 7B & 40.19 & 46.27 &   37.29 & 33.86 \\
                LLaMAX Llama 2 7B & 44.27 & 46.53 &   42.64 & 43.03 \\
         LLaMAX Llama 2 7B Alpaca & 45.09 & 48.47 &   42.80 & 42.89 \\
          MaLA-500 Llama 2 10B v1 & 38.11 & 42.10 &   35.85 & 34.65 \\
          MaLA-500 Llama 2 10B v2 & 38.11 & 42.10 &   35.85 & 34.65 \\
                  YaYi Llama 2 7B & 41.28 & 47.32 &   38.41 & 34.94 \\
             TowerBase Llama 2 7B & 39.84 & 46.08 &   36.33 & 34.39 \\
         TowerInstruct Llama 2 7B & 40.36 & 47.07 &   36.92 & 33.79 \\
              EMMA-500 Llama 2 7B & 45.14 & 46.09 &   44.40 & 44.71 \\
\hline
         Occiglot Mistral 7B v0.1 & 42.35 & 49.90 &   38.39 & 35.19 \\
Occiglot Mistral 7B v0.1 Instruct & 40.81 & 47.58 &   37.18 & 34.52 \\
                         BLOOM 7B & 41.60 & 45.13 &   39.69 & 38.38 \\
                        BLOOMZ 7B & 37.13 & 40.02 &   35.56 & 34.51 \\
                          YaYi 7B & 39.87 & 43.85 &   38.24 & 35.15 \\
                        Aya 23 8B & 43.12 & 48.51 &   41.95 & 34.67 \\
                   Aya Expanse 8B & 45.56 & 50.48 &   44.24 & 38.38 \\
                         Gemma 7B & 42.58 & 46.44 &   41.00 & 38.01 \\
                       Gemma 2 9B & 46.74 & 48.50 &   45.11 & 46.49 \\
                      Qwen 1.5 7B & 39.47 & 40.95 &   38.80 & 37.83 \\
                        Qwen 2 7B & 42.77 & 47.31 &   41.35 & 36.53 \\
                      Qwen 2.5 7B & 43.31 & 47.80 &   41.85 & 37.24 \\
                 Marco-LLM GLO 7B & 43.99 & 48.74 &   41.44 & 39.57 \\
\hline
                       Llama 3 8B & 44.97 & 48.82 &   43.84 & 39.56 \\
                     Llama 3.1 8B & 45.62 & 49.61 &   44.04 & 40.83 \\
                LLaMAX Llama 3 8B & 44.13 & 46.30 &   42.83 & 42.41 \\
         LLaMAX Llama 3 8B Alpaca & 45.08 & 48.18 &   42.82 & 43.40 \\
\hline
         \EMMA Llama 3 8B Mono & 44.15 & 47.15 &   41.73 & 42.99 \\
           \EMMA Llama 3 8B Bi & 45.15 & 47.20 &   43.64 & 44.07 \\
       \EMMA Llama 3.1 8B Mono & 39.98 & 42.34 &   39.00 & 37.20 \\
         \EMMA Llama 3.1 8B Bi & 44.67 & 46.79 &   42.88 & 44.00 \\
\bottomrule
\end{tabular}
\end{table}

\begin{table*}[ht!]
\caption{0-shot results (Accuracy \%) on XNLI in all languages.}
\label{tab:xnli_all}
\setlength{\tabcolsep}{2pt}
\resizebox{\linewidth}{!}{%
\begin{tabular}{lrrrrrrrrrrrrrrrrrrrrrrrrrrrrrrr}
\toprule
\textbf{Model} & \textbf{Avg} & \textbf{ar-acc} & \textbf{stderr} & \textbf{bg-acc} & \textbf{stderr} & \textbf{de-acc} & \textbf{stderr} & \textbf{el-acc} & \textbf{stderr} & \textbf{en-acc} & \textbf{stderr} & \textbf{es-acc} & \textbf{stderr} & \textbf{fr-acc} & \textbf{stderr} & \textbf{hi-acc} & \textbf{stderr} & \textbf{ru-acc} & \textbf{stderr} & \textbf{sw-acc} & \textbf{stderr} & \textbf{th-acc} & \textbf{stderr} & \textbf{tr-acc} & \textbf{stderr} & \textbf{ur-acc} & \textbf{stderr} & \textbf{vi-acc} & \textbf{stderr} & \textbf{zh-acc} & \textbf{stderr} \\
\midrule
                       Llama 2 7B & 40.19 &   35.42 &    0.96 &   42.65 &    0.99 &   47.11 &    1.00 &   36.67 &    0.97 &   55.30 &    1.00 &   40.52 &    0.98 &   50.08 &    1.00 &   37.71 &    0.97 &   42.37 &    0.99 &   34.94 &    0.96 &   36.35 &    0.96 &   37.27 &    0.97 &   33.61 &    0.95 &   36.63 &    0.97 &   36.18 &    0.96 \\
                  Llama 2 7B Chat & 38.58 &   34.42 &    0.95 &   37.07 &    0.97 &   43.09 &    0.99 &   38.15 &    0.97 &   50.24 &    1.00 &   39.44 &    0.98 &   44.82 &    1.00 &   35.78 &    0.96 &   42.09 &    0.99 &   34.22 &    0.95 &   33.49 &    0.95 &   36.95 &    0.97 &   33.90 &    0.95 &   38.11 &    0.97 &   36.95 &    0.97 \\
                   CodeLlama 2 7B & 40.19 &   33.41 &    0.95 &   37.75 &    0.97 &   47.23 &    1.00 &   37.63 &    0.97 &   54.78 &    1.00 &   44.38 &    1.00 &   49.20 &    1.00 &   35.94 &    0.96 &   46.06 &    1.00 &   33.29 &    0.94 &   35.02 &    0.96 &   38.59 &    0.98 &   33.25 &    0.94 &   40.40 &    0.98 &   35.94 &    0.96 \\
                LLaMAX Llama 2 7B & 44.27 &   33.78 &    0.95 &   46.83 &    1.00 &   48.96 &    1.00 &   42.57 &    0.99 &   54.90 &    1.00 &   47.59 &    1.00 &   47.79 &    1.00 &   45.50 &    1.00 &   45.54 &    1.00 &   43.05 &    0.99 &   41.85 &    0.99 &   43.29 &    0.99 &   44.18 &    1.00 &   43.86 &    0.99 &   34.38 &    0.95 \\
         LLaMAX Llama 2 7B Alpaca & 45.09 &   34.42 &    0.95 &   46.39 &    1.00 &   49.76 &    1.00 &   43.41 &    0.99 &   58.11 &    0.99 &   48.96 &    1.00 &   51.97 &    1.00 &   45.62 &    1.00 &   46.27 &    1.00 &   43.57 &    0.99 &   40.80 &    0.99 &   43.86 &    0.99 &   44.30 &    1.00 &   43.09 &    0.99 &   35.78 &    0.96 \\
          MaLA-500 Llama 2 10B v1 & 38.11 &   35.94 &    0.96 &   41.20 &    0.99 &   47.51 &    1.00 &   34.46 &    0.95 &   56.18 &    0.99 &   34.10 &    0.95 &   47.59 &    1.00 &   33.65 &    0.95 &   33.94 &    0.95 &   35.22 &    0.96 &   33.69 &    0.95 &   33.82 &    0.95 &   35.02 &    0.96 &   36.02 &    0.96 &   33.25 &    0.94 \\
          MaLA-500 Llama 2 10B v2 & 38.11 &   35.94 &    0.96 &   41.20 &    0.99 &   47.51 &    1.00 &   34.46 &    0.95 &   56.18 &    0.99 &   34.10 &    0.95 &   47.59 &    1.00 &   33.65 &    0.95 &   33.94 &    0.95 &   35.22 &    0.96 &   33.69 &    0.95 &   33.82 &    0.95 &   35.02 &    0.96 &   36.02 &    0.96 &   33.25 &    0.94 \\
                  YaYi Llama 2 7B & 41.28 &   34.14 &    0.95 &   42.61 &    0.99 &   48.84 &    1.00 &   37.35 &    0.97 &   56.47 &    0.99 &   45.78 &    1.00 &   51.04 &    1.00 &   39.48 &    0.98 &   46.27 &    1.00 &   35.70 &    0.96 &   35.62 &    0.96 &   39.36 &    0.98 &   33.49 &    0.95 &   37.51 &    0.97 &   35.54 &    0.96 \\
             TowerBase Llama 2 7B & 39.84 &   33.90 &    0.95 &   41.37 &    0.99 &   47.87 &    1.00 &   35.26 &    0.96 &   56.35 &    0.99 &   41.69 &    0.99 &   49.44 &    1.00 &   34.54 &    0.95 &   45.94 &    1.00 &   35.02 &    0.96 &   34.78 &    0.95 &   35.74 &    0.96 &   33.37 &    0.95 &   37.19 &    0.97 &   35.22 &    0.96 \\
         TowerInstruct Llama 2 7B & 40.36 &   33.65 &    0.95 &   42.93 &    0.99 &   48.84 &    1.00 &   34.98 &    0.96 &   56.95 &    0.99 &   46.51 &    1.00 &   46.43 &    1.00 &   34.74 &    0.95 &   46.27 &    1.00 &   33.94 &    0.95 &   33.90 &    0.95 &   37.87 &    0.97 &   33.53 &    0.95 &   37.35 &    0.97 &   37.47 &    0.97 \\
              EMMA-500 Llama 2 7B & 45.14 &   34.78 &    0.95 &   46.27 &    1.00 &   47.07 &    1.00 &   45.86 &    1.00 &   53.78 &    1.00 &   47.07 &    1.00 &   46.87 &    1.00 &   47.59 &    1.00 &   46.55 &    1.00 &   46.18 &    1.00 &   41.97 &    0.99 &   44.86 &    1.00 &   45.98 &    1.00 &   47.03 &    1.00 &   35.22 &    0.96 \\
\hline
         Occiglot Mistral 7B v0.1 & 42.35 &   33.86 &    0.95 &   41.37 &    0.99 &   51.77 &    1.00 &   37.71 &    0.97 &   55.86 &    1.00 &   51.65 &    1.00 &   51.93 &    1.00 &   35.74 &    0.96 &   47.63 &    1.00 &   34.70 &    0.95 &   37.39 &    0.97 &   43.01 &    0.99 &   33.49 &    0.95 &   38.63 &    0.98 &   40.56 &    0.98 \\
Occiglot Mistral 7B v0.1 Instruct & 40.81 &   34.38 &    0.95 &   38.84 &    0.98 &   50.84 &    1.00 &   40.00 &    0.98 &   55.66 &    1.00 &   48.63 &    1.00 &   51.69 &    1.00 &   34.30 &    0.95 &   40.04 &    0.98 &   33.17 &    0.94 &   36.35 &    0.96 &   37.99 &    0.97 &   34.06 &    0.95 &   37.59 &    0.97 &   38.63 &    0.98 \\
                         BLOOM 7B & 41.60 &   33.85 &    0.67 &   39.92 &    0.69 &   39.78 &    0.69 &   35.37 &    0.68 &   53.97 &    0.70 &   48.82 &    0.71 &   49.80 &    0.71 &   46.51 &    0.70 &   43.03 &    0.70 &   37.88 &    0.69 &   35.05 &    0.67 &   35.09 &    0.67 &   42.20 &    0.70 &   47.39 &    0.71 &   35.35 &    0.68 \\
                        BLOOMZ 7B & 37.13 &   32.69 &    0.94 &   34.02 &    0.95 &   41.69 &    0.99 &   35.82 &    0.96 &   46.87 &    1.00 &   36.02 &    0.96 &   43.05 &    0.99 &   40.40 &    0.98 &   37.47 &    0.97 &   33.61 &    0.95 &   33.09 &    0.94 &   33.73 &    0.95 &   36.83 &    0.97 &   36.67 &    0.97 &   35.02 &    0.96 \\
                          YaYi 7B & 39.87 &   39.80 &    0.98 &   35.78 &    0.96 &   42.61 &    0.99 &   36.47 &    0.96 &   50.52 &    1.00 &   47.71 &    1.00 &   48.19 &    1.00 &   40.04 &    0.98 &   39.12 &    0.98 &   34.06 &    0.95 &   34.34 &    0.95 &   33.17 &    0.94 &   37.07 &    0.97 &   44.18 &    1.00 &   34.94 &    0.96 \\
                        Aya 23 8B & 43.12 &   33.90 &    0.95 &   39.36 &    0.98 &   49.36 &    1.00 &   41.12 &    0.99 &   51.57 &    1.00 &   50.36 &    1.00 &   51.16 &    1.00 &   46.67 &    1.00 &   48.67 &    1.00 &   34.50 &    0.95 &   35.90 &    0.96 &   48.39 &    1.00 &   33.61 &    0.95 &   42.25 &    0.99 &   39.96 &    0.98 \\
                   Aya Expanse 8B & 45.56 &   34.10 &    0.95 &   41.89 &    0.99 &   51.04 &    1.00 &   42.17 &    0.99 &   53.86 &    1.00 &   47.35 &    1.00 &   53.65 &    1.00 &   48.31 &    1.00 &   51.00 &    1.00 &   37.03 &    0.97 &   41.04 &    0.99 &   48.84 &    1.00 &   37.07 &    0.97 &   50.12 &    1.00 &   45.98 &    1.00 \\
                         Gemma 7B & 42.58 &   33.49 &    0.95 &   43.49 &    0.99 &   48.63 &    1.00 &   38.11 &    0.97 &   52.05 &    1.00 &   44.14 &    1.00 &   49.76 &    1.00 &   44.34 &    1.00 &   47.39 &    1.00 &   40.64 &    0.98 &   37.95 &    0.97 &   43.25 &    0.99 &   35.42 &    0.96 &   43.29 &    0.99 &   36.67 &    0.97 \\
                       Gemma 2 9B & 46.74 &   34.18 &    0.95 &   49.52 &    1.00 &   51.37 &    1.00 &   43.25 &    0.99 &   53.45 &    1.00 &   51.41 &    1.00 &   52.29 &    1.00 &   47.31 &    1.00 &   49.56 &    1.00 &   45.58 &    1.00 &   49.88 &    1.00 &   50.72 &    1.00 &   44.02 &    1.00 &   45.70 &    1.00 &   32.93 &    0.94 \\
                      Qwen 1.5 7B & 39.47 &   34.34 &    0.95 &   40.92 &    0.99 &   42.77 &    0.99 &   36.18 &    0.96 &   49.08 &    1.00 &   37.87 &    0.97 &   43.13 &    0.99 &   38.15 &    0.97 &   38.88 &    0.98 &   35.42 &    0.96 &   44.22 &    1.00 &   38.84 &    0.98 &   33.86 &    0.95 &   44.38 &    1.00 &   33.98 &    0.95 \\
                        Qwen 2 7B & 42.77 &   33.69 &    0.95 &   45.46 &    1.00 &   48.19 &    1.00 &   36.83 &    0.97 &   54.26 &    1.00 &   47.23 &    1.00 &   51.41 &    1.00 &   44.98 &    1.00 &   47.39 &    1.00 &   37.31 &    0.97 &   38.27 &    0.97 &   43.53 &    0.99 &   34.02 &    0.95 &   43.57 &    0.99 &   35.38 &    0.96 \\
                      Qwen 2.5 7B & 43.31 &   34.38 &    0.95 &   42.85 &    0.99 &   47.31 &    1.00 &   41.45 &    0.99 &   53.65 &    1.00 &   48.92 &    1.00 &   51.57 &    1.00 &   42.49 &    0.99 &   47.71 &    1.00 &   34.22 &    0.95 &   43.33 &    0.99 &   42.57 &    0.99 &   34.18 &    0.95 &   47.39 &    1.00 &   37.63 &    0.97 \\
                 Marco-LLM GLO 7B & 43.99 &   33.45 &    0.95 &   45.50 &    1.00 &   50.40 &    1.00 &   37.35 &    0.97 &   53.82 &    1.00 &   50.52 &    1.00 &   51.57 &    1.00 &   42.13 &    0.99 &   46.63 &    1.00 &   34.70 &    0.95 &   44.62 &    1.00 &   44.82 &    1.00 &   39.40 &    0.98 &   45.38 &    1.00 &   39.52 &    0.98 \\
\hline
                       Llama 3 8B & 44.97 &   33.65 &    0.95 &   45.34 &    1.00 &   50.48 &    1.00 &   39.28 &    0.98 &   55.02 &    1.00 &   49.52 &    1.00 &   50.56 &    1.00 &   47.55 &    1.00 &   49.16 &    1.00 &   38.92 &    0.98 &   46.27 &    1.00 &   48.23 &    1.00 &   33.49 &    0.95 &   49.00 &    1.00 &   38.15 &    0.97 \\
                     Llama 3.1 8B & 45.62 &   33.86 &    0.95 &   45.58 &    1.00 &   51.45 &    1.00 &   38.96 &    0.98 &   55.22 &    1.00 &   50.28 &    1.00 &   51.77 &    1.00 &   49.40 &    1.00 &   49.16 &    1.00 &   39.36 &    0.98 &   48.07 &    1.00 &   49.32 &    1.00 &   35.06 &    0.96 &   47.11 &    1.00 &   39.76 &    0.98 \\
                LLaMAX Llama 3 8B & 44.13 &   34.26 &    0.95 &   43.82 &    0.99 &   51.29 &    1.00 &   43.21 &    0.99 &   54.70 &    1.00 &   43.65 &    0.99 &   43.78 &    0.99 &   44.14 &    1.00 &   46.27 &    1.00 &   42.45 &    0.99 &   45.58 &    1.00 &   46.55 &    1.00 &   39.20 &    0.98 &   45.02 &    1.00 &   38.11 &    0.97 \\
         LLaMAX Llama 3 8B Alpaca & 45.08 &   35.42 &    0.96 &   42.57 &    0.99 &   51.45 &    1.00 &   42.61 &    0.99 &   55.06 &    1.00 &   46.91 &    1.00 &   50.28 &    1.00 &   45.10 &    1.00 &   48.80 &    1.00 &   43.73 &    0.99 &   47.23 &    1.00 &   44.78 &    1.00 &   39.24 &    0.98 &   46.43 &    1.00 &   36.59 &    0.97 \\
\hline
         \EMMA Llama 3 8B Mono & 44.15 &   33.01 &    0.94 &   46.06 &    1.00 &   48.59 &    1.00 &   37.91 &    0.97 &   56.39 &    0.99 &   47.43 &    1.00 &   47.95 &    1.00 &   45.78 &    1.00 &   47.79 &    1.00 &   46.14 &    1.00 &   41.00 &    0.99 &   44.50 &    1.00 &   41.81 &    0.99 &   43.09 &    0.99 &   34.74 &    0.95 \\
           \EMMA Llama 3 8B Bi & 45.15 &   33.94 &    0.95 &   46.71 &    1.00 &   50.08 &    1.00 &   44.26 &    1.00 &   53.69 &    1.00 &   48.51 &    1.00 &   48.88 &    1.00 &   48.67 &    1.00 &   48.15 &    1.00 &   45.10 &    1.00 &   45.82 &    1.00 &   44.70 &    1.00 &   41.29 &    0.99 &   43.57 &    0.99 &   33.90 &    0.95 \\
       \EMMA Llama 3.1 8B Mono & 39.98 &   34.66 &    0.95 &   41.33 &    0.99 &   42.61 &    0.99 &   35.62 &    0.96 &   47.99 &    1.00 &   42.33 &    0.99 &   42.41 &    0.99 &   41.16 &    0.99 &   44.58 &    1.00 &   38.47 &    0.98 &   37.19 &    0.97 &   40.56 &    0.98 &   35.94 &    0.96 &   40.68 &    0.98 &   34.10 &    0.95 \\
         \EMMA Llama 3.1 8B Bi & 44.67 &   34.22 &    0.95 &   44.58 &    1.00 &   49.84 &    1.00 &   45.14 &    1.00 &   51.85 &    1.00 &   47.83 &    1.00 &   46.71 &    1.00 &   47.91 &    1.00 &   48.96 &    1.00 &   46.55 &    1.00 &   41.73 &    0.99 &   44.06 &    1.00 &   43.73 &    0.99 &   41.41 &    0.99 &   35.54 &    0.96 \\
\bottomrule
\end{tabular}
}
\end{table*}

\subsection{Machine Translation}
\label{sec:detailed_mt}

We evaluate all models for machine translation under a 3-shot prompting setup. Specifically, we use the prompt below.
\begin{lstlisting}
Translate the following sentence from {src_lang} to {tgt_lang}
[{src_lang}]: {src_sent}
[{tgt_lang}]:
\end{lstlisting}

Translation quality is assessed using BLEU~\citep{papineni-etal-2002-bleu} and chrF++ \citep{popovic-2015-chrf}, implemented via \texttt{sacrebleu}\citep{post-2018-call}. BLEU is computed using the \texttt{flores200} tokenizer, which allows consistent segmentation even for languages lacking explicit word boundaries; chrF++ is calculated with a word n-gram order of 2. For transparency and reproducibility, we report full metric signatures. \footnote{BLEU: nrefs:1|case:mixed|eff:no|tok:flores200|smooth:exp |version:2.4.2; chrF++: nrefs:1|case:mixed|eff:yes|nc:6|nw:2| space:no|version:2.4.2} %

On translations from all other languages to English (Table \ref{tab:flores200_xeng}), \EMMA variants based on both Llama-3 8B and Llama-3.1 8B show clear, systematic gains under monolingual and bilingual continual training. Specifically, our continual training strategy boosts the BLEU/chrF++ scores of Llama-3 8B from 23.78/43.72 to 30.38/50.86 (\EMMA Llama 3 8B Mono, +6.6 BLEU) and further to 34.4/54.33 (\EMMA Llama 3 8B Bi, +10.6 BLEU); the same schedule lifts Llama-3.1 8B from 24.19/44.1 to 27.57/48.12 (\EMMA Llama 3.1 8B Mono, +3.4 BLEU) and 33.6/53.5 ((\EMMA Llama 3 8B Bi, +9.5 BLEU).  

Gains are even larger when English is the source language (Table \ref{tab:flores200_engx}): \EMMA Llama 3 Bi more than doubles base-model BLEU (9.93 $\rightarrow$ 24.02) and ChrF++ (24.08 $\rightarrow$ 42.15), while \EMMA Llama 3.1 Bi shows a comparable jump (10.11 $\rightarrow$ 23.86 BLEU).  Improvements are greatest in the medium-low and low-resource languages, confirming that language-balanced bilingual continual training is especially effective for non-English translations in low-resource settings.

\begin{table*}[ht!]
\caption{3-shot results (BLEU/chrF++) on FLORES-200, from all other languages to English (X-Eng)}
\label{tab:flores200_xeng}
\setlength{\tabcolsep}{2pt}
\scriptsize
\centering
\begin{tabular}{lllllll}
\toprule
\textbf{Model}                 & \textbf{Avg} & \textbf{High} & \textbf{Medium-high} & \textbf{Medium} & \textbf{Medium-low} & \textbf{Low} \\
\midrule
                    Llama 2 7B & 12.93/30.32 & 19.98/38.72 & 12.90/30.87 & 13.13/31.00 & 15.02/33.00 & 13.69/31.26 \\
               Llama 2 7B Chat & 12.28/31.72 & 19.12/39.31 & 11.96/31.89 & 12.24/32.08 & 14.32/34.15 & 13.05/32.61 \\
                  CodeLlama 7B & 10.82/28.57 & 17.77/37.45 & 10.72/29.09 & 10.93/29.19 & 12.77/31.13 & 11.52/29.47 \\
             LLaMAX Llama 2 7B &  1.99/13.66 &  3.63/21.51 &  2.26/15.24 &  2.16/14.65 &  2.48/16.02 &  2.08/14.08 \\
      LLaMAX Llama 2 7B Alpaca & 22.29/42.27 & 34.98/56.64 & 24.68/45.21 & 24.04/44.54 & 26.58/47.33 & 23.85/44.08 \\
               MaLA-500 10B V1 &  2.29/13.60 &  5.01/16.22 &  2.34/13.32 &  2.57/13.81 &  2.85/14.18 &  2.55/13.89 \\
               MaLA-500 10B V2 &  2.87/15.44 &  5.84/18.73 &  2.85/15.33 &  3.25/15.87 &  3.51/16.08 &  3.17/15.76 \\
               Yayi Llama 2 7B & 12.98/31.38 & 19.77/39.36 & 12.93/31.97 & 13.13/32.03 & 15.02/33.96 & 13.76/32.30 \\
    Tower Base Llama 2 7B V0.1 & 13.74/31.47 & 21.68/40.54 & 13.81/32.18 & 13.97/32.28 & 16.05/34.31 & 14.50/32.44 \\
Tower Instruct Llama 2 7B V0.2 &  4.81/25.43 &  9.75/33.70 &  5.29/25.73 &  5.18/25.97 &  5.85/27.70 &  5.24/26.32 \\
           EMMA-500 Llama 2 7B & 25.37/45.78 & 34.84/56.32 & 26.72/47.21 & 26.93/47.47 & 28.89/49.76 & 27.34/48.00 \\
\hline
           Occiglot Mistral 7B & 13.12/31.13 & 19.94/38.88 & 12.78/30.98 & 13.31/31.64 & 15.17/33.58 & 14.00/32.13 \\
  Occiglot Mistral 7B Instruct & 11.61/31.65 & 16.92/39.16 & 11.27/31.61 & 11.80/32.21 & 13.20/33.98 & 12.31/32.61 \\
                     BLOOM 7B1 &  9.57/27.84 & 15.83/36.50 & 12.72/32.18 & 11.77/31.02 & 10.88/29.80 &  9.96/28.51 \\
                    BLOOMZ 7B1 & 20.22/34.74 & 31.74/46.29 & 26.57/41.03 & 24.76/39.25 & 22.39/37.03 & 21.06/35.63 \\
                       Yayi 7B &  4.82/21.36 &  5.04/23.09 &  6.25/24.58 &  5.69/23.82 &  5.20/22.08 &  4.72/21.25 \\
                     Aya 23 8B & 13.87/32.36 & 19.49/39.55 & 15.32/34.13 & 14.87/33.96 & 16.27/35.46 & 14.48/33.31 \\
                Aya Expanse 8B & 13.12/36.86 & 17.35/44.58 & 14.77/39.30 & 14.54/39.15 & 14.81/39.83 & 13.61/37.85 \\
\hline
                      Gemma 7B & 23.79/43.68 & 36.37/57.74 & 26.72/47.06 & 26.34/46.77 & 27.78/48.45 & 25.47/45.65 \\
                    Gemma 2 9B & 23.15/38.87 & 34.42/51.93 & 23.37/38.63 & 24.85/40.38 & 27.22/43.58 & 25.19/41.36 \\
                   Qwen 1.5 7B & 15.58/35.87 & 24.44/46.60 & 16.96/38.19 & 16.97/38.06 & 18.17/39.11 & 16.42/37.01 \\
                     Qwen 2 7B & 17.39/37.61 & 28.37/50.71 & 19.49/40.81 & 19.53/40.66 & 20.66/41.61 & 18.40/38.90 \\
                   Qwen 2.5 7B & 18.95/38.89 & 29.32/51.08 & 21.04/41.75 & 21.21/41.80 & 22.11/42.65 & 19.97/40.11 \\
              Macro-LLM GLO 7B & 25.17/44.57 & 35.96/56.86 & 28.34/48.18 & 28.15/47.90 & 28.54/48.47 & 26.38/45.89 \\
\hline
                    Llama 3 8B & 23.78/43.72 & 35.67/57.20 & 26.55/47.03 & 26.51/46.96 & 28.08/48.72 & 25.49/45.72 \\
                  Llama 3.1 8B & 24.19/44.10 & 36.13/57.57 & 26.76/47.27 & 26.73/47.25 & 28.40/49.05 & 25.91/46.13 \\
             LLaMAX Llama 3 8B &   0.48/4.66 &   0.77/6.36 &   0.55/5.11 &   0.54/5.08 &   0.57/5.19 &   0.52/4.87 \\
      LLaMAX Llama 3 8B Alpaca & 25.10/45.45 & 37.58/58.67 & 27.58/48.13 & 27.49/48.12 & 29.79/50.60 & 26.91/47.43 \\
\hline
      \EMMA Llama 3 8B Mono & 30.38/50.86 & 39.52/60.25 & 31.90/52.47 & 32.44/53.10 & 34.13/54.81 & 32.55/53.16 \\
        \EMMA Llama 3 8B Bi & 34.40/54.33 & 43.48/63.45 & 36.45/56.47 & 36.98/57.00 & 38.45/58.44 & 36.72/56.69 \\
    \EMMA Llama 3.1 8B Mono & 27.57/48.12 & 37.34/58.82 & 29.18/49.85 & 29.50/50.33 & 31.20/52.20 & 29.51/50.34 \\
      \EMMA Llama 3.1 8B Bi & 33.64/53.49 & 42.78/62.87 & 35.87/55.72 & 36.20/56.16 & 37.61/57.62 & 35.91/55.84 \\
\bottomrule
\end{tabular}
\end{table*}

\begin{table*}[ht!]
\caption{3-shot results (BLEU/chrF++) on FLORES-200, from English to all other languages (Eng-X)}
\label{tab:flores200_engx}
\setlength{\tabcolsep}{2pt}
\scriptsize
\centering
\begin{tabular}{lllllll}
\toprule
\textbf{Model}                 & \textbf{Avg} & \textbf{High} & \textbf{Medium-high} & \textbf{Medium} & \textbf{Medium-low} & \textbf{Low} \\
\midrule
                    Llama 2 7B &  4.62/15.13 & 10.72/26.07 &  4.99/16.25 &  4.14/15.68 &  6.03/18.21 &  4.93/16.23 \\
               Llama 2 7B Chat &  4.95/16.95 & 10.81/26.27 &  5.19/17.51 &  4.56/17.18 &  6.27/19.62 &  5.30/18.12 \\
                  CodeLlama 7B &  4.27/14.94 & 10.03/25.25 &  4.61/15.56 &  3.92/15.40 &  5.58/17.75 &  4.57/15.98 \\
             LLaMAX Llama 2 7B &   0.80/7.42 &  2.08/13.88 &   1.14/8.98 &   0.92/8.14 &   1.04/8.88 &   0.86/7.92 \\
      LLaMAX Llama 2 7B Alpaca & 12.51/28.35 & 26.39/45.94 & 16.08/32.23 & 14.07/30.87 & 16.00/34.14 & 13.36/30.45 \\
               MaLA-500 10B V1 &   0.60/6.08 &  1.62/10.25 &   0.69/6.15 &   0.57/6.26 &   0.80/7.13 &   0.65/6.64 \\
               MaLA-500 10B V2 &   0.54/6.38 &  1.37/10.37 &   0.59/6.45 &   0.46/6.42 &   0.67/7.31 &   0.55/6.93 \\
               Yayi Llama 2 7B &  4.41/14.87 & 10.77/25.96 &  4.89/15.86 &  4.07/15.38 &  5.90/18.04 &  4.81/16.04 \\
    Tower Base Llama 2 7B V0.1 &  4.83/16.03 & 11.56/25.70 &  5.35/16.43 &  4.37/16.06 &  6.03/18.47 &  5.10/17.15 \\
Tower Instruct Llama 2 7B V0.2 &  3.23/15.64 &  7.66/24.72 &  3.72/16.21 &  3.21/16.09 &  4.00/17.89 &  3.54/16.84 \\
           EMMA-500 Llama 2 7B & 15.58/33.25 & 28.66/46.92 & 18.00/34.66 & 16.56/34.25 & 18.85/37.66 & 16.87/35.88 \\
\hline
           Occiglot Mistral 7B &  4.32/16.10 & 10.97/25.72 &  5.13/16.65 &  4.36/16.57 &  5.51/18.54 &  4.69/17.23 \\
  Occiglot Mistral 7B Instruct &  3.99/15.80 &  9.80/24.99 &  4.69/16.44 &  4.01/16.39 &  5.09/18.34 &  4.31/16.92 \\
                     BLOOM 7B1 &  2.81/11.80 &  6.98/20.40 &  4.30/14.82 &  3.51/13.81 &  3.27/13.71 &  2.67/12.32 \\
                    BLOOMZ 7B1 &  7.44/16.10 & 22.08/32.72 & 12.35/22.06 & 10.55/20.25 &  9.21/18.84 &  7.48/16.83 \\
                       Yayi 7B &  4.37/13.50 & 12.07/27.36 &  6.87/18.72 &  5.64/16.88 &  5.19/15.91 &  4.23/14.20 \\
                     Aya 23 8B &  6.46/16.15 & 10.76/23.27 &  8.16/18.42 &  6.53/17.12 &  7.75/18.86 &  6.41/16.89 \\
                Aya Expanse 8B &  6.88/23.89 & 10.76/32.48 &  8.24/26.48 &  7.21/25.53 &  7.79/26.89 &  6.78/25.13 \\
                      Gemma 7B &  9.05/23.05 & 18.18/37.69 & 10.87/26.37 & 10.04/25.66 & 11.49/27.58 &  9.56/24.75 \\
                    Gemma 2 9B & 12.09/26.48 & 25.23/44.22 & 14.93/30.26 & 13.67/29.59 & 15.29/32.05 & 12.71/28.50 \\
                   Qwen 1.5 7B &  5.87/17.77 & 13.12/29.98 &  7.14/19.71 &  6.01/19.13 &  7.13/21.03 &  5.93/18.87 \\
                     Qwen 2 7B &  5.56/17.17 & 12.08/28.96 &  6.53/19.22 &  5.77/18.80 &  6.72/20.30 &  5.57/18.17 \\
                   Qwen 2.5 7B &  5.72/17.49 & 12.15/28.73 &  6.81/19.33 &  5.92/18.94 &  6.92/20.61 &  5.71/18.46 \\
              Macro-LLM GLO 7B &  9.27/23.34 & 17.74/36.96 & 11.10/26.36 & 10.30/25.76 & 11.19/27.21 &  9.51/24.79 \\
\hline
                    Llama 3 8B &  9.93/24.08 & 20.79/40.03 & 11.76/26.96 & 10.97/26.59 & 12.58/28.89 & 10.51/25.84 \\
                  Llama 3.1 8B & 10.11/24.69 & 21.29/40.67 & 12.00/27.60 & 11.10/27.20 & 12.76/29.55 & 10.66/26.46 \\
             LLaMAX Llama 3 8B &   0.45/4.65 &   1.11/7.77 &   0.64/5.45 &   0.56/5.19 &   0.57/5.28 &   0.49/4.93 \\
      LLaMAX Llama 3 8B Alpaca & 11.64/26.86 & 25.20/45.23 & 14.71/30.78 & 13.25/29.63 & 14.99/32.76 & 12.44/28.94 \\
\hline
      \EMMA Llama 3 8B Mono & 20.33/38.34 & 33.89/51.45 & 22.10/39.37 & 21.39/39.56 & 24.19/42.93 & 22.03/41.29 \\
        \EMMA Llama 3 8B Bi & 24.02/42.15 & 38.25/55.19 & 25.73/43.49 & 25.32/43.74 & 28.32/47.00 & 25.99/45.22 \\
    \EMMA Llama 3.1 8B Mono & 19.44/37.41 & 32.50/50.37 & 21.25/38.54 & 20.42/38.61 & 23.17/41.97 & 21.06/40.32 \\
      \EMMA Llama 3.1 8B Bi & 23.86/42.07 & 37.81/54.81 & 25.76/43.65 & 25.26/43.74 & 28.15/46.84 & 25.80/45.07 \\
\bottomrule
\end{tabular}
\end{table*}

\subsection{Text Summarization}
\label{sec:detailed_summarization}
We conduct zero-shot text summarization evaluations on three benchmark datasets: MassiveSumm long, MassiveSumm short, and XL-Sum.
The prompt template for the three benchmarks is as follows:
\begin{lstlisting}
Document: {text}
Based on the previous text, provide a brief single summary:
\end{lstlisting}
In order to try to get LLM to answer in the corresponding language of the document, we use Google Translate to translate the prompt into the corresponding language.

We evaluate the generated summaries using two widely recognized metrics: ROUGE-L~\citep{lin-2004-rouge} and BERTScore~\citep{zhang2019bertscore}. ROUGE-L captures lexical overlap by measuring the longest common subsequence (LCS) between the reference and generated summaries. Precision and recall are calculated as the ratio of the LCS to the generated and reference summary lengths, respectively, with an F-score used for the final metric. BERTScore, on the other hand, measures semantic similarity by comparing contextual embeddings from pre-trained language models. Specifically, we employ the \texttt{bert-base-multilingual-cased}\footnote{\url{https://huggingface.co/google-bert/bert-base-multilingual-cased}} model, ensuring compatibility across multiple languages.

Our \EMMA models demonstrate consistent improvements over their Llama 3 and Llama 3.1 base models in zero-shot summarization across the MassiveSumm (long/short) and XL-Sum benchmarks (Tables \ref{tab:massivesumm_long}, \ref{tab:massivesumm_short}, \ref{tab:xl-sum}). 
Notably, enhancements in semantic quality, measured by BERTScore, are frequently observed, especially on the MassiveSumm benchmarks.
Nonetheless, Llama 3/3.1 base models do not obtain impressive summarization performance on these two subsets of MassiveSumm. 
On the XL-Sum benchmark, our \EMMA Llama 3/3.1 CPT models achieve better average performance than recent advances such as Aya 23 and Gemma 2.

\begin{table*}[ht!]
\caption{Zero-short performance of MassiveSumm long set (ROUGE-L/BERTScore). Our \EMMA models demonstrate consistent improvements over their Llama 3 and Llama 3.1 base models.}
\label{tab:massivesumm_long}
\setlength{\tabcolsep}{2pt}
\scriptsize
\centering
\begin{tabular}{lllllll}
\toprule
\textbf{Model}                 & \textbf{Avg} & \textbf{High} & \textbf{Medium-high} & \textbf{Medium} & \textbf{Medium-low} & \textbf{Low} \\
\midrule
                    Llama 2 7B & 4.74/63.89 & 3.88/61.92 &  3.88/61.92 & 3.40/62.75 & 4.78/63.95 & 4.78/63.95 \\
               Llama 2 7B Chat & 4.73/63.52 & 3.88/61.57 &  3.88/61.57 & 3.42/62.34 & 4.76/63.62 & 4.76/63.62 \\
                  CodeLlama 7B & 5.63/64.51 & 4.45/60.73 &  4.45/60.73 & 4.11/63.72 & 5.58/64.54 & 5.58/64.54 \\
             LLaMAX Llama 2 7B & 4.56/62.69 & 3.78/61.26 &  3.78/61.26 & 3.29/61.00 & 4.60/62.73 & 4.60/62.73 \\
      LLaMAX Llama 2 7B Alpaca & 4.61/62.76 & 3.76/61.42 &  3.76/61.42 & 3.28/61.22 & 4.65/62.81 & 4.65/62.81 \\
               MaLA-500 10B V1 & 4.39/64.50 & 3.59/61.82 &  3.59/61.82 & 3.11/63.73 & 4.41/64.58 & 4.41/64.58 \\
               MaLA-500 10B V2 & 4.37/64.66 & 3.62/62.01 &  3.62/62.01 & 3.11/63.82 & 4.40/64.73 & 4.40/64.73 \\
               Yayi Llama 2 7B & 4.98/64.17 & 4.05/62.17 &  4.05/62.17 & 3.69/63.15 & 5.02/64.21 & 5.02/64.21 \\
    Tower Base Llama 2 7B V0.1 & 4.81/64.51 & 3.88/62.39 &  3.88/62.39 & 3.43/63.49 & 4.85/64.59 & 4.85/64.59 \\
Tower Instruct Llama 2 7B V0.2 & 4.82/64.61 & 3.84/62.52 &  3.84/62.52 & 3.40/63.49 & 4.85/64.67 & 4.85/64.67 \\
           EMMA-500 Llama 2 7B & 4.79/63.80 & 3.81/61.99 &  3.81/61.99 & 3.43/62.66 & 4.82/63.88 & 4.82/63.88 \\
\hline
           Occiglot Mistral 7B & 5.14/63.95 & 4.37/61.66 &  4.37/61.66 & 3.93/64.35 & 5.15/64.05 & 5.15/64.05 \\
  Occiglot Mistral 7B Instruct & 5.16/63.50 & 4.57/61.79 &  4.57/61.79 & 4.19/63.86 & 5.19/63.53 & 5.19/63.53 \\
                     BLOOM 7B1 & 4.88/64.36 & 4.00/62.34 &  4.00/62.34 & 3.76/64.65 & 4.92/64.45 & 4.92/64.45 \\
                    BLOOMZ 7B1 & 2.91/57.20 & 2.12/58.71 &  2.12/58.71 & 1.89/57.54 & 2.94/57.17 & 2.94/57.17 \\
                       Yayi 7B & 4.95/64.24 & 4.26/61.87 &  4.26/61.87 & 3.81/64.25 & 4.99/64.32 & 4.99/64.32 \\
                     Aya 23 8B & 6.33/65.94 & 5.54/62.67 &  5.54/62.67 & 4.69/65.29 & 6.34/66.01 & 6.34/66.01 \\
                Aya Expanse 8B & 7.44/67.66 & 6.29/64.39 &  6.29/64.39 & 5.65/66.65 & 7.48/67.76 & 7.48/67.76 \\
                      Gemma 7B & 6.18/62.14 & 5.25/60.19 &  5.25/60.19 & 4.80/62.14 & 6.23/62.19 & 6.23/62.19 \\
                    Gemma 2 9B & 5.86/59.70 & 4.96/57.51 &  4.96/57.51 & 4.81/59.83 & 5.89/59.74 & 5.89/59.74 \\
                   Qwen 1.5 7B & 6.09/59.19 & 5.42/58.28 &  5.42/58.28 & 5.07/60.73 & 6.09/59.14 & 6.09/59.14 \\
                     Qwen 2 7B & 6.65/56.31 & 6.47/54.91 &  6.47/54.91 & 5.43/54.74 & 6.64/56.16 & 6.64/56.16 \\
                   Qwen 2.5 7B & 7.44/61.62 & 6.88/58.85 &  6.88/58.85 & 6.19/59.98 & 7.50/61.59 & 7.50/61.59 \\
              Macro-LLM GLO 7B & 6.57/57.15 & 5.84/53.04 &  5.84/53.04 & 5.26/55.92 & 6.60/57.11 & 6.60/57.11 \\
\hline
                    Llama 3 8B & 5.10/55.58 & 4.15/54.48 &  4.15/54.48 & 4.04/57.63 & 5.13/55.65 & 5.13/55.65 \\
                  Llama 3.1 8B & 5.41/56.09 & 4.36/55.24 &  4.36/55.24 & 4.34/57.85 & 5.45/56.19 & 5.45/56.19 \\
             LLaMAX Llama 3 8B & 6.00/65.90 & 4.74/62.84 &  4.74/62.84 & 4.57/65.35 & 6.03/66.00 & 6.03/66.00 \\
      LLaMAX Llama 3 8B Alpaca & 7.62/67.64 & 6.13/64.77 &  6.13/64.77 & 5.92/66.58 & 7.69/67.77 & 7.69/67.77 \\
\hline
      \EMMA Llama 3 8B Mono & 5.38/61.06 & 4.61/57.84 &  4.61/57.84 & 4.42/60.75 & 5.41/61.08 & 5.41/61.08 \\
        \EMMA Llama 3 8B Bi & 5.57/59.23 & 4.85/57.46 &  4.85/57.46 & 4.67/60.13 & 5.62/59.30 & 5.62/59.30 \\
    \EMMA Llama 3.1 8B Mono & 5.44/61.89 & 4.70/59.04 &  4.70/59.04 & 4.49/61.99 & 5.48/61.98 & 5.48/61.98 \\
      \EMMA Llama 3.1 8B Bi & 4.78/58.47 & 3.70/53.79 &  3.70/53.79 & 3.90/58.36 & 4.81/58.48 & 4.81/58.48 \\
\bottomrule
\end{tabular}
\end{table*}

\begin{table*}[ht!]
\caption{Zero-short performance of MassiveSumm short set (ROUGE-L/BERTScore). Our \EMMA models demonstrate consistent improvements over their Llama 3 and Llama 3.1 base models.}
\label{tab:massivesumm_short}
\setlength{\tabcolsep}{2pt}
\scriptsize
\centering
\begin{tabular}{lllllll}
\toprule
\textbf{Model}                 & \textbf{Avg} & \textbf{High} & \textbf{Medium-high} & \textbf{Medium} & \textbf{Medium-low} & \textbf{Low} \\
\midrule
                    Llama 2 7B &  7.85/65.35 &  6.64/62.55 &  6.23/62.78 & 5.92/64.92 &  7.79/65.43 &  7.88/65.33 \\
               Llama 2 7B Chat &  9.76/67.01 &  8.88/64.52 &  8.29/64.64 & 7.59/66.03 &  9.83/67.08 &  9.87/66.98 \\
                  CodeLlama 7B &  7.59/64.83 &  6.19/61.81 &  5.88/62.10 & 5.54/64.49 &  7.52/65.03 &  7.60/64.93 \\
             LLaMAX Llama 2 7B &  5.22/63.06 &  4.47/59.76 &  4.10/59.92 & 3.46/62.60 &  5.16/63.15 &  5.26/63.10 \\
      LLaMAX Llama 2 7B Alpaca & 10.71/67.92 &  9.52/64.83 &  8.93/65.09 & 8.12/66.93 & 10.73/67.99 & 10.81/67.90 \\
               MaLA-500 10B V1 &  4.97/63.51 &  4.34/61.56 &  4.00/61.65 & 3.53/63.32 &  5.04/63.76 &  5.13/63.74 \\
               MaLA-500 10B V2 &  5.02/63.75 &  4.34/61.61 &  4.00/61.69 & 3.55/63.46 &  5.09/63.97 &  5.18/63.95 \\
               Yayi Llama 2 7B &  7.80/65.24 &  6.91/63.23 &  6.46/63.33 & 5.96/64.80 &  7.84/65.34 &  7.87/65.21 \\
    Tower Base Llama 2 7B V0.1 &  8.11/65.53 &  6.70/62.59 &  6.26/62.78 & 5.94/64.96 &  8.04/65.58 &  8.12/65.48 \\
Tower Instruct Llama 2 7B V0.2 & 10.14/67.76 &  8.88/65.06 &  8.24/65.12 & 7.44/66.81 & 10.15/67.86 & 10.18/67.71 \\
           EMMA-500 Llama 2 7B &  8.32/65.14 &  6.85/62.52 &  6.41/62.56 & 6.13/64.26 &  8.30/65.25 &  8.43/65.16 \\
\hline
           Occiglot Mistral 7B &  8.16/63.65 &  6.39/61.67 &  6.05/61.74 & 5.79/64.17 &  8.15/64.00 &  8.26/63.96 \\
  Occiglot Mistral 7B Instruct &  7.82/63.79 &  6.72/62.31 &  6.37/62.13 & 6.05/64.25 &  7.72/63.74 &  7.84/63.71 \\
                     BLOOM 7B1 &  6.79/62.30 &  5.71/58.74 &  5.34/58.80 & 4.97/61.23 &  6.73/62.31 &  6.82/62.26 \\
                    BLOOMZ 7B1 &  3.28/29.75 &  3.72/39.04 &  3.41/36.51 & 2.78/32.31 &  3.16/29.30 &  3.20/29.11 \\
                       Yayi 7B &  8.28/65.44 &  7.73/65.10 &  7.25/64.79 & 6.62/65.09 &  8.30/65.66 &  8.35/65.42 \\
                     Aya 23 8B &  8.43/65.85 &  7.49/63.54 &  6.99/63.73 & 6.14/65.59 &  8.47/66.04 &  8.47/65.85 \\
                Aya Expanse 8B &  9.24/67.68 &  8.49/65.77 &  7.96/65.95 & 7.38/67.20 &  9.32/67.88 &  9.39/67.72 \\
                      Gemma 7B &  8.35/62.25 &  7.56/60.19 &  7.09/60.45 & 6.39/62.35 &  8.35/62.28 &  8.47/62.29 \\
                    Gemma 2 9B &  7.86/58.11 &  6.36/55.20 &  5.86/53.92 & 5.95/57.87 &  7.85/57.99 &  7.97/58.11 \\
                   Qwen 1.5 7B &  8.49/62.70 &  7.88/61.33 &  7.40/61.00 & 6.87/63.25 &  8.44/62.60 &  8.50/62.57 \\
                     Qwen 2 7B &  8.63/56.14 &  7.65/53.08 &  7.17/53.36 & 6.33/55.72 &  8.49/55.57 &  8.54/55.69 \\
                   Qwen 2.5 7B &  9.04/58.91 &  9.49/60.70 &  8.85/60.34 & 7.64/60.70 &  9.03/58.90 &  9.04/58.56 \\
              Macro-LLM GLO 7B &  8.10/57.45 &  8.09/56.96 &  7.51/57.15 & 6.87/60.07 &  8.06/57.25 &  8.10/57.29 \\
\hline
                    Llama 3 8B &  6.44/50.98 &  4.78/48.31 &  4.53/48.77 & 4.68/52.54 &  6.29/50.70 &  6.45/50.91 \\
                  Llama 3.1 8B &  6.77/54.96 &  5.31/52.26 &  5.02/53.04 & 4.91/55.77 &  6.67/54.79 &  6.83/54.97 \\
             LLaMAX Llama 3 8B &  8.77/66.46 &  7.05/63.80 &  6.60/63.77 & 6.27/65.97 &  8.65/66.58 &  8.76/66.47 \\
      LLaMAX Llama 3 8B Alpaca & 12.44/68.95 & 10.27/67.22 &  9.54/67.08 & 9.42/68.35 & 12.44/69.09 & 12.48/68.94 \\
\hline
      \EMMA Llama 3 8B Mono &  7.18/63.46 &  6.39/61.99 &  5.97/62.06 & 5.68/64.19 &  7.19/63.69 &  7.24/63.50 \\
        \EMMA Llama 3 8B Bi &  7.74/63.18 &  6.49/60.71 &  6.01/60.71 & 5.72/63.07 &  7.76/63.28 &  7.82/63.26 \\
    \EMMA Llama 3.1 8B Mono &  7.21/63.36 &  6.68/63.09 &  6.15/61.81 & 5.67/62.64 &  7.25/63.45 &  7.34/63.46 \\
      \EMMA Llama 3.1 8B Bi &  6.67/61.64 &  5.61/58.37 &  5.21/58.16 & 5.01/61.53 &  6.67/61.59 &  6.75/61.64 \\
\bottomrule
\end{tabular}
\end{table*}

\begin{table*}[ht!]
\caption{Zero-short performance of XL-Sum (ROUGE-L/BERTScore). Our \EMMA models demonstrate consistent improvements over their Llama 3 and Llama 3.1 base models.}
\label{tab:xl-sum}
\setlength{\tabcolsep}{2pt}
\scriptsize
\centering
\begin{tabular}{lllllll}
\toprule
                         Model &         Avg &        High & Medium-high &      Medium &  Medium-low &         Low \\
\midrule
                    Llama 2 7B &  7.11/66.52 &  9.75/64.83 &  9.52/65.35 &  8.34/65.08 &  7.62/65.58 &  7.73/65.39 \\
               Llama 2 7B Chat &  8.84/68.44 & 12.91/67.19 & 12.32/67.84 & 10.47/67.45 &  9.54/67.69 &  9.67/67.51 \\
                  CodeLlama 7B &  7.15/65.74 &  9.89/62.99 &  9.66/64.20 &  8.38/63.72 &  7.71/64.74 &  7.83/64.68 \\
             LLaMAX Llama 2 7B &  5.29/64.59 &  6.86/61.74 &  6.66/62.29 &  5.99/61.96 &  5.71/63.24 &  5.79/63.17 \\
      LLaMAX Llama 2 7B Alpaca & 10.11/69.24 & 14.44/67.88 & 14.31/68.35 & 12.42/68.26 & 11.12/68.38 & 11.19/68.19 \\
               MaLA-500 10B V1 &  5.45/63.96 &  7.98/60.89 &  7.40/61.16 &  6.38/61.14 &  5.88/62.48 &  5.94/62.55 \\
               MaLA-500 10B V2 &  5.44/64.28 &  7.80/61.21 &  7.28/61.62 &  6.34/61.51 &  5.86/62.71 &  5.91/62.79 \\
               Yayi Llama 2 7B &  7.98/67.21 & 11.64/66.06 & 10.91/66.54 &  9.41/66.17 &  8.59/66.63 &  8.71/66.32 \\
    Tower Base Llama 2 7B V0.1 &  7.65/67.09 & 10.64/65.26 & 10.45/65.62 &  8.97/65.39 &  8.21/66.12 &  8.30/65.97 \\
Tower Instruct Llama 2 7B V0.2 &  8.89/68.46 & 12.81/67.15 & 12.00/67.56 & 10.30/67.06 &  9.53/67.72 &  9.65/67.52 \\
           EMMA-500 Llama 2 7B &  8.58/67.20 & 11.61/66.17 & 11.75/66.93 & 10.29/67.92 &  9.41/67.80 &  9.52/67.68 \\
\hline
           Occiglot Mistral 7B &  7.33/66.20 & 10.38/63.81 & 10.17/64.44 &  8.86/64.39 &  7.93/65.00 &  8.02/64.88 \\
  Occiglot Mistral 7B Instruct &  8.31/66.96 & 12.53/65.19 & 11.44/65.98 & 10.08/65.16 &  9.04/66.17 &  9.11/66.04 \\
                     BLOOM 7B1 &  6.99/64.78 &  8.83/61.91 &  8.62/63.21 &  7.64/62.69 &  7.43/63.72 &  7.55/63.43 \\
                    BLOOMZ 7B1 & 11.15/69.82 & 20.33/67.38 & 19.39/68.58 & 15.83/68.77 & 12.52/69.37 & 12.50/69.21 \\
                       Yayi 7B & 12.06/69.74 & 19.99/67.84 & 19.25/68.82 & 16.07/68.85 & 13.42/69.14 & 13.37/68.97 \\
                     Aya 23 8B &  8.68/66.79 & 12.91/66.50 & 11.90/66.45 &  9.92/67.34 &  9.25/67.11 &  9.43/66.91 \\
                Aya Expanse 8B & 10.51/68.73 & 14.88/67.38 & 14.09/67.96 & 12.58/69.15 & 11.24/69.07 & 11.32/68.88 \\
                      Gemma 7B &  6.70/64.52 &  9.16/62.37 &  9.16/62.76 &  8.17/64.49 &  7.33/64.93 &  7.30/64.67 \\
                    Gemma 2 9B &  7.38/65.45 &  9.82/61.04 &  9.83/62.66 &  8.91/62.78 &  8.01/64.14 &  8.07/64.11 \\
                   Qwen 1.5 7B &  9.58/69.13 & 14.33/67.56 & 13.58/68.14 & 12.04/68.02 & 10.42/68.34 & 10.44/68.16 \\
                     Qwen 2 7B & 10.18/69.35 & 15.67/68.37 & 14.66/68.66 & 12.70/68.42 & 11.12/68.58 & 11.10/68.39 \\
                   Qwen 2.5 7B & 10.41/69.69 & 15.68/69.57 & 14.96/69.69 & 12.95/70.62 & 11.38/70.19 & 11.31/69.93 \\
              Macro-LLM GLO 7B & 11.46/70.41 & 17.22/70.74 & 16.65/70.83 & 14.31/71.32 & 12.65/71.03 & 12.62/70.76 \\
\hline
                    Llama 3 8B &  8.47/67.08 & 10.96/64.16 & 11.24/65.16 & 10.12/65.04 &  9.29/65.82 &  9.31/65.79 \\
                  Llama 3.1 8B &  8.57/66.97 & 10.97/63.60 & 11.19/64.55 & 10.14/64.81 &  9.45/65.66 &  9.43/65.66 \\
             LLaMAX Llama 3 8B &  8.28/66.72 & 10.85/64.80 & 10.97/65.33 &  9.71/66.92 &  9.04/67.25 &  9.07/66.99 \\
      LLaMAX Llama 3 8B Alpaca & 11.39/69.99 & 15.45/69.66 & 15.82/70.20 & 13.36/70.75 & 12.44/70.53 & 12.42/70.34 \\
\hline
      \EMMA Llama 3 8B Mono &  9.11/66.12 & 12.96/64.51 & 13.25/65.23 & 11.30/66.81 & 10.09/66.83 &  9.96/66.50 \\
        \EMMA Llama 3 8B Bi &  9.10/66.72 & 13.30/65.28 & 13.11/65.91 & 11.17/67.34 & 10.10/67.46 &  9.94/67.12 \\
    \EMMA Llama 3.1 8B Mono &  9.66/67.21 & 12.88/65.35 & 13.18/66.30 & 11.53/67.78 & 10.63/67.93 & 10.58/67.66 \\
      \EMMA Llama 3.1 8B Bi &  8.56/65.90 & 12.19/64.10 & 11.75/64.40 & 10.51/66.43 &  9.51/66.53 &  9.47/66.31 \\
\bottomrule
\end{tabular}
\end{table*}

\subsection{Machine Reading Comprehension}
\label{sec:detailed_mrc}

We first evaluate on the BELEBELE dataset \citep{bandarkar2023belebele}, a comprehensive multilingual benchmark for machine reading comprehension, spanning 122 languages across both high- and low-resource categories. 
Notably, this benchmark demonstrates significant difficulty, with even the English subset posing substantial challenges to state-of-the-art language models like Llama 3. 
As shown in \Cref{tab:MRC}, which presents zero-shot performance across language resource groups, our continual pre-training approach on Llama 3 and 3.1, as well as LLaMAX CPT models, yields some degraded improvements over the base Llama model.
However, when measured by the number of languages with improved performance, our \EMMA CPT is better than the base models. 
Contemporary models such as Aya Expanse, Llama 3.1, and Qwen 2 demonstrate more competent performance, suggesting advances in multilingual understanding and reasoning capacities. 
On the harder ARC-Multilingual set, the bilingual Llama-3 variant essentially ties the base model ($\approx$ 35\%), while the others dip by up to one point. 
These results suggest that EMMA-style bilingual continual pre-training alone is insufficient for cross-language passage‐question reasoning at the 8 B scale, the fundamental limitations of Llama models, and the average-based evaluation practice.

\begin{table*}[ht!]
\caption{0-shot results (Accuracy \%) on BELEBELE, and 5-shot results (Accuracy \%) on the multilingual ARC.}
\label{tab:MRC}
\setlength{\tabcolsep}{2pt}
\scriptsize
\centering
\begin{tabular}{lrrrrrr|rrrr}
\toprule
\multirow{2}{*}{\textbf{Model}}   & \multicolumn{6}{c|}{\textbf{BELEBELE}}                                                                & \multicolumn{4}{c}{\textbf{ARC Multilingual}}                 \\ \cline{2-11} 
                                  & \textbf{Avg} & \textbf{High} & \textbf{Med-high} & \textbf{Medium} & \textbf{Med-low} & \textbf{Low} & \textbf{Avg} & \textbf{High} & \textbf{Medium} & \textbf{Low} \\
\midrule
Llama 2 7B                        & 26.27        & 26.76         & 26.35             & 26.07           & 26.41            & 26.27        & 27.56        & 33.12         & 27.31           & 21.02        \\
Llama 2 7B Chat                   & 29.05        & 31.84         & 29.97             & 28.95           & 29.47            & 29.09        & 28.02        & 33.69         & 27.79           & 21.29        \\
CodeLlama 2 7B                    & 27.38        & 27.37         & 27.33             & 27.30           & 27.30            & 27.38        & 25.23        & 28.86         & 24.64           & 21.65        \\
LLaMAX Llama 2 7B                 & 23.09        & 23.23         & 23.15             & 23.07           & 23.10            & 23.08        & 26.09        & 30.00         & 25.92           & 21.48        \\
LLaMAX Llama 2 7B Alpaca          & 24.48        & 25.46         & 24.82             & 24.41           & 24.60            & 24.49        & 31.06        & 36.89         & 31.85           & 22.49        \\
MaLA-500 Llama 2 10B v1           & 22.96        & 23.02         & 22.98             & 22.97           & 22.98            & 22.97        & 21.16        & 21.92         & 20.48           & 21.32        \\
MaLA-500 Llama 2 10B v2           & 22.96        & 23.02         & 22.98             & 22.97           & 22.98            & 22.97        & 21.16        & 21.92         & 20.48           & 21.32        \\
YaYi Llama 2 7B                   & 28.32        & 29.64         & 28.67             & 28.11           & 28.37            & 28.26        & 28.40        & 34.30         & 28.35           & 21.11        \\
TowerBase Llama 2 7B              & 26.36        & 27.43         & 26.85             & 26.29           & 26.48            & 26.34        & 27.94        & 35.32         & 26.82           & 20.51        \\
TowerInstruct Llama 2 7B          & 27.93        & 29.88         & 28.51             & 27.57           & 28.19            & 27.92        & 30.10        & 38.88         & 28.85           & 21.16        \\
EMMA-500 Llama 2 7B               & 26.75        & 28.32         & 28.18             & 27.58           & 27.14            & 26.94        & 29.53        & 34.10         & 29.82           & 23.34        \\
\hline
Occiglot Mistral 7B v0.1          & 30.16        & 32.25         & 30.94             & 30.02           & 30.40            & 30.15        & 29.77        & 38.39         & 28.51           & 21.03        \\
Occiglot Mistral 7B v0.1 Instruct & 32.05        & 34.14         & 32.62             & 31.74           & 32.40            & 32.08        & 30.88        & 40.29         & 29.65           & 21.13        \\
BLOOM 7B                          & 24.11        & 24.25         & 24.52             & 24.12           & 24.11            & 24.08        & 23.65        & 26.27         & 22.72           & 21.89        \\
BLOOMZ 7B                         & 39.32        & 45.43         & 43.67             & 41.51           & 40.08            & 39.51        & 23.95        & 26.94         & 22.74           & 22.18        \\
YaYi 7B                           & 37.97        & 44.37         & 42.71             & 40.49           & 38.72            & 38.09        & 24.44        & 27.96         & 23.29           & 21.91        \\
Aya 23 8B                         & 40.08        & 43.85         & 41.71             & 39.22           & 40.93            & 39.81        & 31.08        & 40.05         & 30.02           & 21.61        \\
Aya Expanse 8B                    & 46.98        & 52.22         & 48.99             & 46.57           & 48.36            & 46.93        & 36.56        & 47.87         & 36.15           & 23.09        \\
Gemma 7B                          & 43.37        & 52.63         & 47.83             & 44.82           & 45.43            & 43.94        & 38.68        & 46.46         & 40.47           & 26.06        \\
Gemma 2 9B                        & 54.49        & 64.10         & 58.90             & 55.83           & 56.85            & 55.05        & 44.15        & 54.59         & 46.18           & 27.82        \\
Qwen 1.5 7B                       & 41.83        & 48.86         & 44.79             & 42.18           & 43.00            & 41.78        & 28.93        & 35.55         & 28.14           & 21.92        \\
Qwen 2 7B                         & 49.31        & 57.62         & 52.20             & 50.04           & 51.16            & 49.48        & 33.82        & 43.88         & 32.64           & 23.17        \\
Qwen 2.5 7B                       & 54.11        & 63.47         & 57.91             & 55.47           & 56.24            & 54.30        & 35.30        & 46.43         & 34.31           & 23.00        \\
Marco-LLM GLO 7B                  & 53.95        & 63.54         & 58.67             & 55.50           & 56.20            & 54.31        & 36.34        & 46.72         & 35.88           & 24.11        \\
\hline
Llama 3 8B                        & 40.73        & 46.07         & 42.92             & 41.03           & 41.87            & 40.88        & 34.80        & 42.43         & 35.53           & 24.06        \\
Llama 3.1 8B                      & 45.19        & 52.50         & 48.01             & 45.65           & 46.74            & 45.34        & 34.93        & 42.43         & 35.89           & 24.00        \\
LLaMAX Llama 3 8B                 & 36.96        & 40.06         & 37.92             & 36.78           & 37.68            & 37.09        & 33.54        & 39.81         & 34.84           & 23.59        \\
LLaMAX Llama 3 8B Alpaca          & 39.41        & 44.06         & 41.10             & 39.52           & 40.53            & 39.61        & 34.53        & 41.34         & 35.56           & 24.34        \\
\hline
\EMMA Llama 3 8B Mono          & 39.73        & 43.65         & 41.40             & 39.96           & 40.76            & 40.01        & 33.22        & 38.32         & 34.56           & 24.67        \\
\EMMA Llama 3 8B Bi            & 39.84        & 43.51         & 41.09             & 40.05           & 40.74            & 39.99        & 34.84        & 40.37         & 36.45           & 25.31        \\
\EMMA Llama 3.1 8B Mono        & 38.86        & 41.68         & 39.34             & 38.49           & 39.62            & 38.94        & 34.00        & 39.38         & 35.39           & 25.01        \\
\EMMA Llama 3.1 8B Bi          & 37.00        & 40.59         & 38.36             & 37.21           & 37.94            & 37.25        & 34.59        & 39.85         & 36.21           & 25.38        \\ \hline
\end{tabular}
\end{table*}

\subsection{Math}
\label{sec:detailed_math}

We use the Multilingual Grade School Math Benchmark (MGSM) \citep{shi2022language} to evaluate the mathematical reasoning in LLMs. 
As shown in \Cref{tab:mgsm}, we evaluate model performance using 3-shot prompting with flexible answer matching. 
Our assessment employs two distinct prompting strategies: (1) Direct: direct question prompting, and (2) Chain-of-Thought (CoT): prompting accompanied by step-by-step reasoning examples \citep{wei2022chain}. 
The results show that our CPT models, as well as LLaMAX CPT models, obtain decreased average performance on direct prompting. Our \texttt{\EMMA Llama 3.1 8B Mono} model slightly improves the base model.
Notably, our \EMMA Llama 3/3.1 CPT models significantly improve the performance on low-resource languages for both direct and CoT prompting.

\Cref{tab:mgsm_direct_all,tab:mgsm_cot_all} present the per-language performance across all the languages by direct and CoT prompting. 

\begin{table*}[ht!]
\caption{3-shot results (Accuracy \%) on MGSM obtained by direct and CoT prompting. Our \EMMA CPT models significantly improve the performance on low-resource languages.}
\label{tab:mgsm}
\setlength{\tabcolsep}{2pt}
\scriptsize
\centering
\begin{tabular}{l|rrrr|rrrr}
\toprule
\multirow{2}{*}{\textbf{Model}}   & \multicolumn{4}{c|}{\textbf{Direct}}                           & \multicolumn{4}{c}{\textbf{CoT}}                              \\
                                  & \textbf{Avg} & \textbf{High} & \textbf{Medium} & \textbf{Low} & \textbf{Avg} & \textbf{High} & \textbf{Medium} & \textbf{Low} \\
\midrule
Llama 2 7B                        & 6.69         & 8.07          & 2.13            & 1.20         & 6.36         & 7.60          & 2.13            & 0.80         \\
Llama 2 7B Chat                   & 10.22        & 13.73         & 2.13            & 0.80         & 10.91        & 13.53         & 2.80            & 1.60         \\
CodeLlama 2 7B                    & 5.93         & 7.07          & 2.93            & 1.20         & 6.64         & 8.73          & 2.67            & 2.00         \\
LLaMAX Llama 2 7B                 & 3.35         & 4.00          & 2.00            & 0.80         & 3.62         & 4.33          & 2.27            & 2.40         \\
LLaMAX Llama 2 7B Alpaca          & 5.05         & 5.20          & 4.00            & 1.60         & 6.35         & 8.07          & 4.13            & 0.80         \\
MaLA-500 Llama 2 10B v1           & 0.91         & 1.33          & 0.27            & 0.00         & 0.73         & 1.27          & 0.00            & 0.00         \\
MaLA-500 Llama 2 10B v2           & 0.91         & 1.33          & 0.27            & 0.00         & 0.73         & 1.27          & 0.00            & 0.00         \\
YaYi Llama 2 7B                   & 7.09         & 9.47          & 1.47            & 1.20         & 7.22         & 8.73          & 2.27            & 1.20         \\
TowerBase Llama 2 7B              & 6.15         & 8.33          & 1.73            & 0.80         & 6.16         & 8.60          & 2.40            & 0.80         \\
TowerInstruct Llama 2 7B          & 7.24         & 9.53          & 1.73            & 2.00         & 8.24         & 10.47         & 1.87            & 1.20         \\
EMMA-500 Llama 2 7B               & 17.02        & 19.20         & 11.87           & 2.40         & 18.09        & 20.00         & 13.20           & 2.80         \\
\hline
Occiglot Mistral 7B v0.1          & 13.31        & 16.87         & 4.53            & 1.60         & 14.07        & 18.80         & 3.60            & 1.20         \\
Occiglot Mistral 7B v0.1 Instruct & 22.76        & 29.80         & 7.47            & 2.80         & 22.16        & 30.40         & 7.87            & 2.80         \\
BLOOM 7B                          & 2.87         & 2.60          & 2.80            & 3.60         & 2.29         & 2.20          & 1.47            & 2.00         \\
BLOOMZ 7B                         & 2.55         & 2.67          & 2.40            & 1.20         & 2.15         & 1.67          & 3.07            & 2.00         \\
YaYi 7B                           & 2.76         & 2.93          & 1.47            & 2.40         & 3.02         & 2.93          & 2.40            & 2.00         \\
Aya 23 8B                         & 22.29        & 30.67         & 3.47            & 0.80         & 24.71        & 35.07         & 5.47            & 2.40         \\
Aya Expanse 8B                    & 43.02        & 55.47         & 18.93           & 5.20         & 41.45        & 54.67         & 18.53           & 5.20         \\
Gemma 7B                          & 38.22        & 36.60         & 38.27           & 27.20        & 35.78        & 34.67         & 37.07           & 26.80        \\
Gemma 2 9B                        & 32.95        & 28.00         & 35.73           & 30.80        & 44.69        & 36.07         & 52.00           & 47.20        \\
Qwen 1.5 7B                       & 31.56        & 40.00         & 16.00           & 4.00         & 30.36        & 40.60         & 14.80           & 2.40         \\
Qwen 2 7B                         & 48.95        & 54.40         & 38.80           & 14.80        & 51.47        & 58.93         & 39.07           & 14.40        \\
Qwen 2.5 7B                       & 53.78        & 65.33         & 36.13           & 8.00         & 55.60        & 68.87         & 36.40           & 8.80         \\
Marco-LLM GLO 7B                  & 51.85        & 63.80         & 32.80           & 11.60        & 52.02        & 62.93         & 34.93           & 11.60        \\
\hline
Llama 3 8B                        & 27.45        & 27.87         & 26.13           & 5.60         & 28.13        & 28.53         & 26.67           & 5.20         \\
Llama 3.1 8B                      & 28.36        & 29.00         & 26.13           & 4.40         & 27.31        & 27.27         & 25.47           & 8.40         \\
LLaMAX Llama 3 8B                 & 20.80        & 22.73         & 20.40           & 6.80         & 19.96        & 20.93         & 22.40           & 3.60         \\
LLaMAX Llama 3 8B Alpaca          & 14.18        & 14.87         & 11.87           & 6.00         & 17.16        & 20.87         & 12.27           & 4.00         \\
\hline
\EMMA Llama 3 8B Mono          & 23.53        & 22.33         & 25.60           & 12.00        & 25.33        & 24.67         & 27.73           & 13.60        \\
\EMMA Llama 3 8B Bi            & 23.49        & 23.00         & 24.67           & 10.80        & 26.29        & 24.87         & 30.67           & 11.60        \\
\EMMA Llama 3.1 8B Mono        & 24.95        & 23.67         & 27.20           & 11.20        & 27.35        & 26.60         & 30.27           & 10.80        \\
\EMMA Llama 3.1 8B Bi          & 23.85        & 23.67         & 25.33           & 11.20        & 25.76        & 26.27         & 27.87           & 12.80        \\ \hline
\end{tabular}
\end{table*}

\begin{table*}[ht!]
\caption{3-shot results (Accuracy \%) on MGSM in all languages by direct prompting and flexible matching.}
\label{tab:mgsm_direct_all}
\setlength{\tabcolsep}{2pt}
\resizebox{\linewidth}{!}{%
\begin{tabular}{lrrrrrrrrrrrrrrrrrrrrrrr}
\toprule
\textbf{Model} & \textbf{Avg} & \textbf{bn} & \textbf{bn-stderr} & \textbf{de} & \textbf{de-stderr} & \textbf{en} & \textbf{en-stderr} & \textbf{es} & \textbf{es-stderr} & \textbf{fr} & \textbf{fr-stderr} & \textbf{ja} & \textbf{ja-stderr} & \textbf{ru} & \textbf{ru-stderr} & \textbf{sw} & \textbf{sw-stderr} & \textbf{te} & \textbf{te-stderr} & \textbf{th} & \textbf{th-stderr} & \textbf{zh} & \textbf{zh-stderr} \\
\midrule
                       Llama 2 7B &          6.69 &  2.80 &       1.05 &  8.00 &       1.72 & 17.60 &       2.41 & 11.20 &       2.00 & 12.00 &       2.06 &  2.40 &       0.97 &  8.00 &       1.72 &  2.80 &       1.05 &  1.20 &       0.69 &  0.80 &       0.56 &  6.80 &       1.60 \\
                  Llama 2 7B Chat &         10.22 &  2.80 &       1.05 & 16.80 &       2.37 & 22.80 &       2.66 & 19.60 &       2.52 & 19.20 &       2.50 &  2.40 &       0.97 & 14.40 &       2.22 &  0.80 &       0.56 &  0.80 &       0.56 &  2.80 &       1.05 & 10.00 &       1.90 \\
                   CodeLlama 2 7B &          5.93 &  1.60 &       0.80 &  8.80 &       1.80 & 12.80 &       2.12 &  8.80 &       1.80 & 10.40 &       1.93 &  5.60 &       1.46 &  6.00 &       1.51 &  2.00 &       0.89 &  1.20 &       0.69 &  5.20 &       1.41 &  2.80 &       1.05 \\
                LLaMAX Llama 2 7B &          3.35 &  2.80 &       1.05 &  3.60 &       1.18 &  6.00 &       1.51 &  2.00 &       0.89 &  7.20 &       1.64 &  3.20 &       1.12 &  2.40 &       0.97 &  1.60 &       0.80 &  0.80 &       0.56 &  1.60 &       0.80 &  5.60 &       1.46 \\
         LLaMAX Llama 2 7B Alpaca &          5.05 &  4.00 &       1.24 &  3.60 &       1.18 & 10.80 &       1.97 &  6.00 &       1.51 &  6.40 &       1.55 &  3.20 &       1.12 &  4.40 &       1.30 &  4.80 &       1.35 &  1.60 &       0.80 &  3.20 &       1.12 &  7.60 &       1.68 \\
          MaLA-500 Llama 2 10B v1 &          0.91 &  0.00 &       0.00 &  0.00 &       0.00 &  1.20 &       0.69 &  2.00 &       0.89 &  2.40 &       0.97 &  1.20 &       0.69 &  2.00 &       0.89 &  0.40 &       0.40 &  0.00 &       0.00 &  0.40 &       0.40 &  0.40 &       0.40 \\
          MaLA-500 Llama 2 10B v2 &          0.91 &  0.00 &       0.00 &  0.00 &       0.00 &  1.20 &       0.69 &  2.00 &       0.89 &  2.40 &       0.97 &  1.20 &       0.69 &  2.00 &       0.89 &  0.40 &       0.40 &  0.00 &       0.00 &  0.40 &       0.40 &  0.40 &       0.40 \\
                  YaYi Llama 2 7B &          7.09 &  3.20 &       1.12 &  8.40 &       1.76 & 15.60 &       2.30 & 16.00 &       2.32 & 10.40 &       1.93 &  5.20 &       1.41 &  5.60 &       1.46 &  0.80 &       0.56 &  1.20 &       0.69 &  0.40 &       0.40 & 11.20 &       2.00 \\
             TowerBase Llama 2 7B &          6.15 &  2.40 &       0.97 &  8.40 &       1.76 & 11.60 &       2.03 &  9.20 &       1.83 &  8.80 &       1.80 &  4.80 &       1.35 & 10.00 &       1.90 &  1.20 &       0.69 &  0.80 &       0.56 &  1.60 &       0.80 &  8.80 &       1.80 \\
         TowerInstruct Llama 2 7B &          7.24 &  1.60 &       0.80 & 10.00 &       1.90 & 15.20 &       2.28 & 15.60 &       2.30 & 12.80 &       2.12 &  1.60 &       0.80 & 10.00 &       1.90 &  1.60 &       0.80 &  2.00 &       0.89 &  2.00 &       0.89 &  7.20 &       1.64 \\
              EMMA-500 Llama 2 7B &         17.02 &  8.80 &       1.80 & 23.20 &       2.68 & 34.00 &       3.00 & 28.00 &       2.85 & 25.60 &       2.77 &  9.20 &       1.83 & 22.80 &       2.66 & 16.80 &       2.37 &  2.40 &       0.97 & 10.00 &       1.90 &  6.40 &       1.55 \\
\hline
         Occiglot Mistral 7B v0.1 &         13.31 &  3.20 &       1.12 & 21.20 &       2.59 & 30.00 &       2.90 & 27.20 &       2.82 & 21.60 &       2.61 &  6.40 &       1.55 & 15.20 &       2.28 &  2.40 &       0.97 &  1.60 &       0.80 &  8.00 &       1.72 &  9.60 &       1.87 \\
Occiglot Mistral 7B v0.1 Instruct &         22.76 &  4.80 &       1.35 & 34.00 &       3.00 & 46.40 &       3.16 & 40.00 &       3.10 & 31.60 &       2.95 & 18.40 &       2.46 & 23.60 &       2.69 &  6.40 &       1.55 &  2.80 &       1.05 & 11.20 &       2.00 & 31.20 &       2.94 \\
                         BLOOM 7B &          2.87 &  2.40 &       0.97 &  1.60 &       0.80 &  4.00 &       1.24 &  3.60 &       1.18 &  1.20 &       0.69 &  2.00 &       0.89 &  3.60 &       1.18 &  4.00 &       1.24 &  3.60 &       1.18 &  2.00 &       0.89 &  3.60 &       1.18 \\
                        BLOOMZ 7B &          2.55 &  3.20 &       1.12 &  1.60 &       0.80 &  3.60 &       1.18 &  2.40 &       0.97 &  3.20 &       1.12 &  2.80 &       1.05 &  3.60 &       1.18 &  2.80 &       1.05 &  1.20 &       0.69 &  1.20 &       0.69 &  2.40 &       0.97 \\
                          YaYi 7B &          2.76 &  2.40 &       0.97 &  2.00 &       0.89 &  6.00 &       1.51 &  2.40 &       0.97 &  5.60 &       1.46 &  1.60 &       0.80 &  1.20 &       0.69 &  1.20 &       0.69 &  2.40 &       0.97 &  0.80 &       0.56 &  4.80 &       1.35 \\
                        Aya 23 8B &         22.29 &  3.20 &       1.12 & 37.20 &       3.06 & 50.00 &       3.17 & 41.20 &       3.12 & 39.20 &       3.09 & 27.60 &       2.83 & 34.00 &       3.00 &  2.80 &       1.05 &  0.80 &       0.56 &  4.40 &       1.30 &  4.80 &       1.35 \\
                   Aya Expanse 8B &         43.02 & 22.40 &       2.64 & 69.20 &       2.93 & 78.40 &       2.61 & 72.00 &       2.85 & 63.60 &       3.05 & 50.80 &       3.17 & 69.60 &       2.92 & 12.80 &       2.12 &  5.20 &       1.41 & 21.60 &       2.61 &  7.60 &       1.68 \\
                         Gemma 7B &         38.22 & 34.40 &       3.01 & 44.80 &       3.15 & 58.80 &       3.12 & 48.00 &       3.17 & 39.60 &       3.10 & 16.80 &       2.37 & 41.20 &       3.12 & 37.60 &       3.07 & 27.20 &       2.82 & 42.80 &       3.14 & 29.20 &       2.88 \\
                       Gemma 2 9B &         32.95 & 29.60 &       2.89 & 40.40 &       3.11 & 56.40 &       3.14 & 50.80 &       3.17 & 36.00 &       3.04 &  1.20 &       0.69 & 35.20 &       3.03 & 37.60 &       3.07 & 30.80 &       2.93 & 40.00 &       3.10 &  4.40 &       1.30 \\
                      Qwen 1.5 7B &         31.56 & 12.40 &       2.09 & 44.00 &       3.15 & 55.20 &       3.15 & 48.40 &       3.17 & 45.20 &       3.15 & 16.80 &       2.37 & 43.20 &       3.14 &  7.20 &       1.64 &  4.00 &       1.24 & 28.40 &       2.86 & 42.40 &       3.13 \\
                        Qwen 2 7B &         48.95 & 44.40 &       3.15 & 65.60 &       3.01 & 80.80 &       2.50 & 76.00 &       2.71 & 69.60 &       2.92 &  1.60 &       0.80 & 67.20 &       2.98 & 16.00 &       2.32 & 14.80 &       2.25 & 56.00 &       3.15 & 46.40 &       3.16 \\
                      Qwen 2.5 7B &         53.78 & 26.80 &       2.81 & 63.20 &       3.06 & 83.20 &       2.37 & 74.00 &       2.78 & 66.40 &       2.99 & 52.40 &       3.16 & 72.80 &       2.82 & 20.00 &       2.53 &  8.00 &       1.72 & 61.60 &       3.08 & 63.20 &       3.06 \\
                 Marco-LLM GLO 7B &         51.85 & 31.60 &       2.95 & 64.40 &       3.03 & 77.60 &       2.64 & 71.20 &       2.87 & 63.20 &       3.06 & 50.40 &       3.17 & 63.60 &       3.05 & 18.00 &       2.43 & 11.60 &       2.03 & 48.80 &       3.17 & 70.00 &       2.90 \\
\hline
                       Llama 3 8B &         27.45 & 17.60 &       2.41 & 39.60 &       3.10 & 50.80 &       3.17 & 47.20 &       3.16 & 36.40 &       3.05 &  3.60 &       1.18 & 37.60 &       3.07 & 24.00 &       2.71 &  5.60 &       1.46 & 36.80 &       3.06 &  2.80 &       1.05 \\
                     Llama 3.1 8B &         28.36 & 20.00 &       2.53 & 41.20 &       3.12 & 55.20 &       3.15 & 48.40 &       3.17 & 38.40 &       3.08 &  3.60 &       1.18 & 40.00 &       3.10 & 20.80 &       2.57 &  4.40 &       1.30 & 37.60 &       3.07 &  2.40 &       0.97 \\
                LLaMAX Llama 3 8B &         20.80 & 13.20 &       2.15 & 20.40 &       2.55 & 24.40 &       2.72 & 26.00 &       2.78 & 22.00 &       2.63 & 20.80 &       2.57 & 22.80 &       2.66 & 21.20 &       2.59 &  6.80 &       1.60 & 26.80 &       2.81 & 24.40 &       2.72 \\
         LLaMAX Llama 3 8B Alpaca &         14.18 & 12.00 &       2.06 & 17.60 &       2.41 & 25.20 &       2.75 & 17.20 &       2.39 & 17.20 &       2.39 & 11.20 &       2.00 & 15.60 &       2.30 & 10.80 &       1.97 &  6.00 &       1.51 & 12.80 &       2.12 & 10.40 &       1.93 \\
\hline
         \EMMA Llama 3 8B Mono &         23.53 & 20.00 &       2.53 & 30.00 &       2.90 & 36.00 &       3.04 & 34.80 &       3.02 & 30.00 &       2.90 &  3.20 &       1.12 & 33.20 &       2.98 & 33.20 &       2.98 & 12.00 &       2.06 & 23.60 &       2.69 &  2.80 &       1.05 \\
           \EMMA Llama 3 8B Bi &         23.49 & 21.20 &       2.59 & 30.80 &       2.93 & 35.60 &       3.03 & 36.80 &       3.06 & 34.00 &       3.00 &  2.80 &       1.05 & 30.40 &       2.92 & 30.00 &       2.90 & 10.80 &       1.97 & 22.80 &       2.66 &  3.20 &       1.12 \\
       \EMMA Llama 3.1 8B Mono &         24.95 & 21.20 &       2.59 & 34.80 &       3.02 & 39.60 &       3.10 & 35.60 &       3.03 & 33.20 &       2.98 &  1.20 &       0.69 & 34.80 &       3.02 & 36.80 &       3.06 & 11.20 &       2.00 & 23.60 &       2.69 &  2.40 &       0.97 \\
         \EMMA Llama 3.1 8B Bi &         23.85 & 20.80 &       2.57 & 38.00 &       3.08 & 33.20 &       2.98 & 39.60 &       3.10 & 32.40 &       2.97 &  2.00 &       0.89 & 28.40 &       2.86 & 27.20 &       2.82 & 11.20 &       2.00 & 28.00 &       2.85 &  1.60 &       0.80 \\
\bottomrule
\end{tabular}
}
\end{table*}

\begin{table*}[ht!]
\caption{3-shot results (Accuracy \%) on MGSM in all languages by CoT prompting and flexible matching.}
\label{tab:mgsm_cot_all}
\setlength{\tabcolsep}{2pt}
\resizebox{\linewidth}{!}{%
\begin{tabular}{lrrrrrrrrrrrrrrrrrrrrrrr}
\toprule
\textbf{Model} & \textbf{Avg} & \textbf{bn} & \textbf{bn-stderr} & \textbf{de} & \textbf{de-stderr} & \textbf{en} & \textbf{en-stderr} & \textbf{es} & \textbf{es-stderr} & \textbf{fr} & \textbf{fr-stderr} & \textbf{ja} & \textbf{ja-stderr} & \textbf{ru} & \textbf{ru-stderr} & \textbf{sw} & \textbf{sw-stderr} & \textbf{te} & \textbf{te-stderr} & \textbf{th} & \textbf{th-stderr} & \textbf{zh} & \textbf{zh-stderr} \\
\midrule
                       Llama 2 7B &                6.36 &  2.00 &       0.89 &  7.60 &       1.68 & 16.00 &       2.32 & 12.40 &       2.09 &  9.20 &       1.83 &  3.60 &       1.18 &  8.00 &       1.72 &  2.00 &       0.89 &  0.80 &       0.56 &  1.60 &       0.80 &  8.00 &       1.72 \\
                  Llama 2 7B Chat &               10.91 &  2.40 &       0.97 & 17.60 &       2.41 & 27.20 &       2.82 & 19.20 &       2.50 & 18.80 &       2.48 &  4.00 &       1.24 & 15.20 &       2.28 &  2.00 &       0.89 &  0.80 &       0.56 &  2.80 &       1.05 & 11.60 &       2.03 \\
                   CodeLlama 2 7B &                6.64 &  1.60 &       0.80 &  9.20 &       1.83 & 13.20 &       2.15 & 10.80 &       1.97 & 10.80 &       1.97 &  4.40 &       1.30 &  7.60 &       1.68 &  1.20 &       0.69 &  1.60 &       0.80 &  6.00 &       1.51 &  4.00 &       1.24 \\
                LLaMAX Llama 2 7B &                3.62 &  3.60 &       1.18 &  3.60 &       1.18 &  7.20 &       1.64 &  3.60 &       1.18 &  5.20 &       1.41 &  2.80 &       1.05 &  2.80 &       1.05 &  0.80 &       0.56 &  0.80 &       0.56 &  2.00 &       0.89 &  4.80 &       1.35 \\
         LLaMAX Llama 2 7B Alpaca &                6.35 &  3.20 &       1.12 &  4.80 &       1.35 & 15.20 &       2.28 &  9.60 &       1.87 &  5.20 &       1.41 &  4.40 &       1.30 &  3.60 &       1.18 &  4.80 &       1.35 &  0.40 &       0.40 &  4.80 &       1.35 &  6.80 &       1.60 \\
          MaLA-500 Llama 2 10B v1 &                0.73 &  0.00 &       0.00 &  0.40 &       0.40 &  0.40 &       0.40 &  0.40 &       0.40 &  1.60 &       0.80 &  1.60 &       0.80 &  2.40 &       0.97 &  0.40 &       0.40 &  0.00 &       0.00 &  0.00 &       0.00 &  0.80 &       0.56 \\
          MaLA-500 Llama 2 10B v2 &                0.73 &  0.00 &       0.00 &  0.40 &       0.40 &  0.40 &       0.40 &  0.40 &       0.40 &  1.60 &       0.80 &  1.60 &       0.80 &  2.40 &       0.97 &  0.40 &       0.40 &  0.00 &       0.00 &  0.00 &       0.00 &  0.80 &       0.56 \\
                  YaYi Llama 2 7B &                7.22 &  2.80 &       1.05 &  8.40 &       1.76 & 16.80 &       2.37 & 12.40 &       2.09 & 10.40 &       1.93 &  4.80 &       1.35 &  7.20 &       1.64 &  2.00 &       0.89 &  1.20 &       0.69 &  3.20 &       1.12 & 12.40 &       2.09 \\
             TowerBase Llama 2 7B &                6.16 &  3.60 &       1.18 &  8.00 &       1.72 & 11.20 &       2.00 &  8.40 &       1.76 &  7.60 &       1.68 &  3.60 &       1.18 &  8.40 &       1.76 &  1.20 &       0.69 &  0.80 &       0.56 &  2.80 &       1.05 &  9.20 &       1.83 \\
         TowerInstruct Llama 2 7B &                8.24 &  1.20 &       0.69 & 12.80 &       2.12 & 18.80 &       2.48 & 15.20 &       2.28 & 12.00 &       2.06 &  2.40 &       0.97 & 13.60 &       2.17 &  1.60 &       0.80 &  1.20 &       0.69 &  3.20 &       1.12 & 10.80 &       1.97 \\
              EMMA-500 Llama 2 7B &               18.09 &  8.40 &       1.76 & 22.40 &       2.64 & 37.60 &       3.07 & 25.60 &       2.77 & 21.60 &       2.61 &  8.00 &       1.72 & 22.80 &       2.66 & 21.20 &       2.59 &  2.40 &       0.97 & 11.60 &       2.03 & 16.40 &       2.35 \\
\hline
         Occiglot Mistral 7B v0.1 &               14.07 &  2.40 &       0.97 & 21.60 &       2.61 & 33.20 &       2.98 & 26.80 &       2.81 & 22.40 &       2.64 &  6.40 &       1.55 & 17.20 &       2.39 &  3.20 &       1.12 &  1.20 &       0.69 &  6.00 &       1.51 & 11.20 &       2.00 \\
Occiglot Mistral 7B v0.1 Instruct &               22.16 &  4.00 &       1.24 & 33.20 &       2.98 & 43.20 &       3.14 & 42.00 &       3.13 & 31.60 &       2.95 & 16.80 &       2.37 & 24.40 &       2.72 &  5.60 &       1.46 &  1.60 &       0.80 &  8.40 &       1.76 & 24.80 &       2.74 \\
                         BLOOM 7B &                2.29 &  2.00 &       0.89 &  1.60 &       0.80 &  4.80 &       1.35 &  2.40 &       0.97 &  2.40 &       0.97 &  0.40 &       0.40 &  4.40 &       1.30 &  1.60 &       0.80 &  2.00 &       0.89 &  2.00 &       0.89 &  2.40 &       0.97 \\
                        BLOOMZ 7B &                2.15 &  2.00 &       0.89 &  2.00 &       0.89 &  2.40 &       0.97 &  3.20 &       1.12 &  3.20 &       1.12 &  2.00 &       0.89 &  1.60 &       0.80 &  2.40 &       0.97 &  2.00 &       0.89 &  1.60 &       0.80 &  1.20 &       0.69 \\
                          YaYi 7B &                3.02 &  4.80 &       1.35 &  3.20 &       1.12 &  4.80 &       1.35 &  3.60 &       1.18 &  4.00 &       1.24 &  2.00 &       0.89 &  1.20 &       0.69 &  1.60 &       0.80 &  2.80 &       1.05 &  0.80 &       0.56 &  6.00 &       1.51 \\
                        Aya 23 8B &               24.71 &  6.00 &       1.51 & 40.80 &       3.11 & 48.80 &       3.17 & 41.20 &       3.12 & 40.00 &       3.10 & 28.40 &       2.86 & 37.20 &       3.06 &  5.60 &       1.46 &  2.40 &       0.97 &  8.40 &       1.76 &  6.80 &       1.60 \\
                   Aya Expanse 8B &               41.45 & 21.20 &       2.59 & 68.80 &       2.94 & 78.80 &       2.59 & 76.80 &       2.68 & 65.20 &       3.02 & 29.60 &       2.89 & 67.60 &       2.97 & 11.20 &       2.00 &  2.40 &       0.97 & 20.00 &       2.53 &  2.80 &       1.05 \\
                         Gemma 7B &               35.78 & 36.40 &       3.05 & 40.80 &       3.11 & 60.80 &       3.09 & 45.20 &       3.15 & 42.00 &       3.13 &  3.60 &       1.18 & 39.20 &       3.09 & 38.80 &       3.09 & 24.80 &       2.74 & 44.00 &       3.15 &  4.80 &       1.35 \\
                       Gemma 2 9B &               44.69 & 48.80 &       3.17 & 53.60 &       3.16 & 72.00 &       2.85 & 60.40 &       3.10 & 50.00 &       3.17 &  2.00 &       0.89 & 50.00 &       3.17 & 53.60 &       3.16 & 44.40 &       3.15 & 55.20 &       3.15 &  1.60 &       0.80 \\
                      Qwen 1.5 7B &               30.36 & 12.80 &       2.12 & 43.20 &       3.14 & 58.40 &       3.12 & 50.40 &       3.17 & 42.00 &       3.13 & 11.20 &       2.00 & 36.80 &       3.06 &  8.80 &       1.80 &  4.00 &       1.24 & 25.60 &       2.77 & 26.00 &       2.78 \\
                        Qwen 2 7B &               51.47 & 45.20 &       3.15 & 66.00 &       3.00 & 82.00 &       2.43 & 75.20 &       2.74 & 71.60 &       2.86 &  2.80 &       1.05 & 68.40 &       2.95 & 17.20 &       2.39 & 17.60 &       2.41 & 58.40 &       3.12 & 60.80 &       3.09 \\
                      Qwen 2.5 7B &               55.60 & 25.60 &       2.77 & 69.60 &       2.92 & 81.60 &       2.46 & 73.60 &       2.79 & 68.00 &       2.96 & 60.00 &       3.10 & 74.80 &       2.75 & 17.20 &       2.39 &  9.20 &       1.83 & 56.80 &       3.14 & 72.80 &       2.82 \\
                 Marco-LLM GLO 7B &               52.02 & 32.00 &       2.96 & 65.20 &       3.02 & 78.80 &       2.59 & 70.00 &       2.90 & 64.00 &       3.04 & 48.80 &       3.17 & 64.40 &       3.03 & 18.00 &       2.43 &  9.60 &       1.87 & 56.00 &       3.15 & 63.60 &       3.05 \\
\hline
                       Llama 3 8B &               28.13 & 19.20 &       2.50 & 38.40 &       3.08 & 54.00 &       3.16 & 46.00 &       3.16 & 36.00 &       3.04 &  2.00 &       0.89 & 38.00 &       3.08 & 25.20 &       2.75 &  6.40 &       1.55 & 39.20 &       3.09 &  4.00 &       1.24 \\
                     Llama 3.1 8B &               27.31 & 19.20 &       2.50 & 42.00 &       3.13 & 52.80 &       3.16 & 42.80 &       3.14 & 34.80 &       3.02 &  2.80 &       1.05 & 37.20 &       3.06 & 25.20 &       2.75 &  5.60 &       1.46 & 35.60 &       3.03 &  1.60 &       0.80 \\
                LLaMAX Llama 3 8B &               19.96 & 12.40 &       2.09 & 20.40 &       2.55 & 24.80 &       2.74 & 25.60 &       2.77 & 18.40 &       2.46 & 18.80 &       2.48 & 20.40 &       2.55 & 23.20 &       2.68 &  3.60 &       1.18 & 30.80 &       2.93 & 19.60 &       2.52 \\
         LLaMAX Llama 3 8B Alpaca &               17.16 & 14.80 &       2.25 & 18.00 &       2.43 & 26.00 &       2.78 & 22.40 &       2.64 & 24.00 &       2.71 & 13.60 &       2.17 & 22.40 &       2.64 & 12.40 &       2.09 &  7.20 &       1.64 & 11.60 &       2.03 & 13.20 &       2.15 \\
\hline
         \EMMA Llama 3 8B Mono &               25.33 & 23.20 &       2.68 & 34.00 &       3.00 & 37.60 &       3.07 & 40.40 &       3.11 & 32.80 &       2.98 &  1.60 &       0.80 & 33.60 &       2.99 & 32.40 &       2.97 & 12.80 &       2.12 & 24.00 &       2.71 &  2.80 &       1.05 \\
           \EMMA Llama 3 8B Bi &               26.29 & 22.00 &       2.63 & 34.40 &       3.01 & 40.40 &       3.11 & 42.80 &       3.14 & 37.20 &       3.06 &  0.80 &       0.56 & 31.60 &       2.95 & 33.60 &       2.99 & 12.00 &       2.06 & 29.20 &       2.88 &  1.20 &       0.69 \\
       \EMMA Llama 3.1 8B Mono &               27.35 & 22.00 &       2.63 & 38.00 &       3.08 & 40.00 &       3.10 & 41.60 &       3.12 & 36.40 &       3.05 &  3.20 &       1.12 & 37.20 &       3.06 & 42.40 &       3.13 & 10.40 &       1.93 & 26.40 &       2.79 &  2.00 &       0.89 \\
         \EMMA Llama 3.1 8B Bi &               25.76 & 24.40 &       2.72 & 38.00 &       3.08 & 35.20 &       3.03 & 38.80 &       3.09 & 35.20 &       3.03 &  1.60 &       0.80 & 33.20 &       2.98 & 35.20 &       3.03 &  8.80 &       1.80 & 25.20 &       2.75 &  1.20 &       0.69 \\
\bottomrule
\end{tabular}
}
\end{table*}

\section{Ethics Considerations}
\label{sec:ethics}

\paragraph{Data Collection and Use} This work involves the compilation and use of large-scale monolingual and bilingual textual corpora covering over 500 languages. All datasets used for continual pre-training are obtained from publicly available sources or established linguistic resources that permit academic research use. Given the scale and diversity of the data, we acknowledge the potential presence of sensitive or harmful content. Future work could include filtering mechanisms or annotation efforts to improve the quality and safety of multilingual data.

\paragraph{Representation of Low-Resource Languages}
Our research aims to expand LLM coverage to underrepresented languages. However, the quantity and quality of data for low-resource languages remain uneven. This may inadvertently lead to biased model behaviors or inadequate performance for some linguistic groups. 

\paragraph{Community Involvement}
This work does not directly involve community collaboration. However, we recognize the importance of inclusive research practices and welcome future partnerships with linguists, native speakers, and regional institutions to improve the quality and cultural relevance of multilingual language technologies.

\end{appendices}

\end{document}